\documentclass[10pt,journal,compsoc]{IEEEtran}
\usepackage{graphicx}
\usepackage{subfloat}

\usepackage{float}
\usepackage{amsmath}
\usepackage{CJK}
\usepackage{amssymb}
\usepackage{indentfirst}
\usepackage{mathtools}
\usepackage{subfigure}
\usepackage{stfloats}
\usepackage{mathrsfs}
\usepackage{bm}
\usepackage{blindtext}
\usepackage{algorithm}
\usepackage{algpseudocode}
\usepackage{amsmath}

\usepackage{amsmath,amssymb}
\DeclareSymbolFont{EulerExtension}{U}{euex}{m}{n}
\DeclareMathSymbol{\euintop}{\mathop} {EulerExtension}{"52}
\DeclareMathSymbol{\euointop}{\mathop} {EulerExtension}{"48}

\usepackage[Symbol]{upgreek}
\usepackage{booktabs}
\usepackage{multirow}

\usepackage{array}
\newcommand{\PreserveBackslash}[1]{\let\temp=\\#1\let\\=\temp}

\ifCLASSINFOpdf

\else

\fi

\begin{document}
%

\title{Non-Cooperative Game Theory Based Rate\\Adaptation for Dynamic Video Streaming over\\HTTP}
%
%
%

\author{Hui~Yuan,~\IEEEmembership{Senior Member,~IEEE,}
        Huayong~Fu,
        Ju~Liu,~\IEEEmembership{Senior Member,~IEEE,}\\
        Junhui~Hou,~\IEEEmembership{Member,~IEEE,}
        and~Sam~Kwong,~\IEEEmembership{Fellow,~IEEE}

\thanks{This work was supported in part by the National Natural Science Foundation of China under Grants 61571274,61672443;
in part by the Shandong Natural Science Funds for Distinguished
Young Scholar under Grant JQ201614; in part by Hong Kong RGC General
Research Fund(GRF) under Grant 9042322 (CityU 11200116);
and in part by the Young Scholars Program of Shandong University (YSPSDU) under Grant 2015WLJH39.}

\IEEEcompsocitemizethanks{
\IEEEcompsocthanksitem H. Yuan, H. Fu, and J. Liu are with the School of Information Science and Engineering, Shandong University, Ji'nan 250100, China.\protect\\
Email: huiyuan@sdu.edu.cn, johy.fu@gmail.com, juliu@sdu.edu.cn 
\IEEEcompsocthanksitem J. Hou and S. Kwong are with the Department
of Computer Science, City University of Hong Kong, Kowloon, Hong
Kong.\protect\\
Email: jh.hou@cityu.edu.hk, cssamk@cityu.edu.hk}
}

\IEEEtitleabstractindextext{%
\begin{abstract}
Dynamic Adaptive Streaming over HTTP (DASH) has demonstrated to be an
emerging and promising multimedia streaming technique, owing to its
capability of dealing with the variability of networks. Rate
adaptation mechanism, a challenging and open issue, plays an
important role in DASH based systems since it affects Quality of
Experience (\emph{QoE}) of users, network utilization, etc. In this
paper, based on non-cooperative game theory, we propose a novel
algorithm to optimally allocate the limited export bandwidth of the
server to multi-users to maximize their \emph{QoE} with fairness
guaranteed. The proposed algorithm is proxy-free. Specifically, a
novel user \emph{QoE} model is derived by taking a variety of factors
into account, like the received video quality, the reference buffer
length, and user accumulated buffer lengths, etc. Then, the bandwidth
competing problem is formulated as a non-cooperation game with the
existence of Nash Equilibrium that is theoretically proven. Finally,
a distributed iterative algorithm with stability analysis is proposed
to find the Nash Equilibrium. Compared with state-of-the-art methods,
extensive experimental results in terms of both simulated and
realistic networking scenarios demonstrate that the proposed
algorithm can produce higher \emph{QoE}, and the actual buffer
lengths of all users keep nearly optimal states, i.e., moving around
the reference buffer all the time. Besides, the proposed algorithm
produces no playback interruption.
\end{abstract}

\begin{IEEEkeywords}
Non-cooperative Game, Nash Equilibrium, DASH, Bitrate Adaptation,
QoE.
\end{IEEEkeywords}}

\maketitle

\IEEEdisplaynontitleabstractindextext

%
\IEEEpeerreviewmaketitle

\begin{figure}
\setlength{\abovecaptionskip}{0.cm}
\setlength{\belowcaptionskip}{-0.cm} \centering
\includegraphics[width=8.76cm]{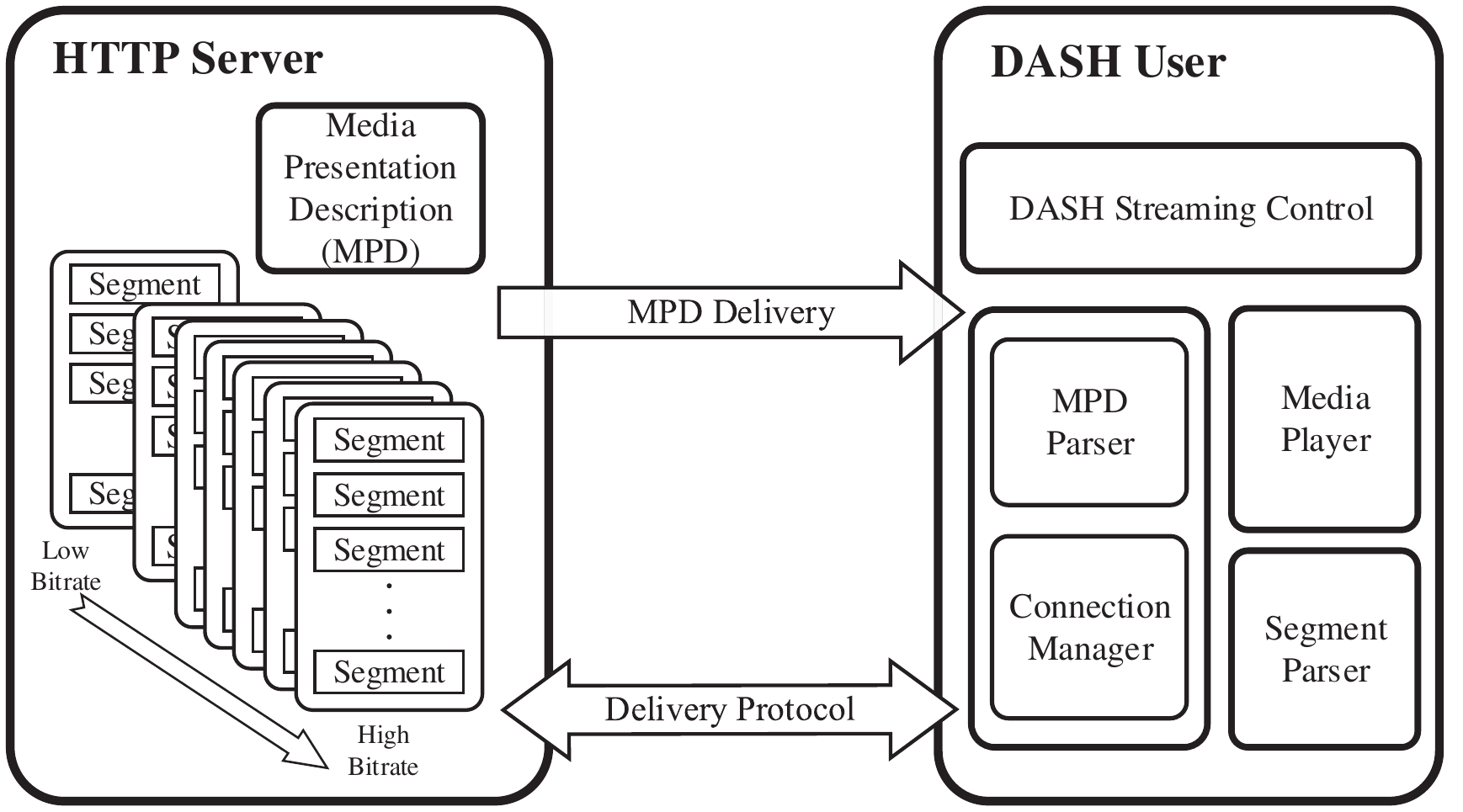}
\caption{DASH system architecture.} \label{fig1}
\end{figure}

\IEEEraisesectionheading{\section{Introduction}\label{sec:introduction}}
\IEEEPARstart{N}{owadays}, with the increase of Internet bandwidth
and the tremendous growth of web platforms, Hypertext Transfer
Protocol (HTTP) streaming has become a cost-effective method for
multimedia delivery \cite{R1}\cite{R2}. Dynamic Adaptive Streaming
over HTTP (DASH) is a typical HTTP based multimedia delivery standard
that can transmit multimedia content adaptively between multimedia
servers and users with a limited and varied network bandwidth
\cite{R3}. Fig. \ref{fig1} illustrates a typical DASH-based video
delivery system. In this system, the media content (e.g., videos or
audios) is first divided into multiple segments (or chunks) \cite{R4}
with the same display time. Each segment is then encoded/transcoded
with different bitrates (corresponding to different quality levels).
At the same time, the server generates a Media Presentation
Description (MPD) file that records the information of the available
video content, e.g., URL addresses, segment lengths, quality levels,
resolutions, etc. The users first download the MPD file from the
server using HTTP protocol, and then request segments with different
quality levels to adapt to the bandwidth variation. The main
advantage of DASH is that it can achieve bandwidth adaptation and
reduce the number of playback interruptions under fluctuating network
conditions \cite{R5}. An effective rate adaptation algorithm is
necessary in a DASH system, with which the DASH user can adaptively
request video segments with different bitrates based on the network
condition and its buffer length. However, this challenging issue is
not specified in the DASH standard. Without an effective rate
adaptation algorithm, the DASH user might suffer from frequent
interruptions. Moreover, recent studies show that the DASH user's
selfish behavior (i.e., making requests without considering other
users sharing the network resources) will result in network
underutilization (or congestion), fluctuating and unfair throughout
allocation \cite{R6}. This paper aims to develop an effective rate
adaptation algorithm to address the above issues.

Many rate adaptation methods have been proposed (see Section II).
However, most of them optimize the HTTP streaming of multiple DASH
users sharing the same network resources \emph{separately},
regardless of the influence between each other; thus, user fairness
cannot be well guaranteed. \textbf{\emph{In contrast, the proposed
rate adaptation algorithm optimizes the HTTP streaming of multiple
DASH users simultaneously who compete for higher quality video
segments from a single server with a limited export bandwidth}}.

As a branch of game theory, the non-cooperative game theory \cite{R7}
can resolve the conflicts among interacting players involved in a
certain game, in which each player behaves selfishly to optimize its
own profit usually quantified as an objective function. The
non-cooperative game can provide meaningful solutions for many
applications where the interaction among several players is
negligible and centralized approaches are not suitable
\cite{R8}\cite{R9}. The typical application of a non-cooperative game
is the oligopoly market problem in economics, in which all
corporations compete for market share of the same commodity in order
to maximize their own profits \cite{R10}, and the market share of
each corporation tends to be stable and reaches Nash Equilibrium.

In this paper, we consider formulating the rate adaptation problem
for improving user \emph{QoE}, as well as preserving user fairness,
as a non-cooperative game in which DASH users try to consume the
limited export bandwidth of the server as much as possible to
maximize their profits (i.e., \emph{QoE}). The optimal bitrate that
produces the optimal \emph{QoE} can be obtained when the Nash
Equilibrium of the contradiction problem is achieved. More
specifically, the proposed non-cooperative game theory based rate
adaptation algorithm determines the requested bitrate for a DASH user
based on both the local information (e.g., the requested bitrate of
the last segment, reference buffer length, and current buffer length)
and the global payoff variation information obtained from the server.
Each user can gradually adapt the requested bitrate to convergence.
The bitrate's convergence speed is controlled by the learning rate.
Note that there is a limitation that additional HTTP sessions between
users and servers are needed in the proposed method, which may reduce
the transmission efficiency of the streaming system. Even so,
extensive experimental results demonstrate that the proposed
algorithm can produce higher \emph{QoE}, while the actual buffer
lengths of all users move around the reference buffer all the time
when compared with state-of-the-art algorithms. Moreover, the
proposed algorithm is proxy-free and playback interruption-free. To
the best of our knowledge, this is the first time to address the DASH
rate adaptation problem using the non-cooperative game theory. The
major contributions of this paper are summarized as follows.

\begin{itemize}
\item We formulated the rate adaptation problem as a non-cooperative game
with the existence of the Nash Equilibrium that is theoretically
proven. The proposed rate adaptation algorithm optimizes streaming
for multiple DASH users simultaneously to guarantee their fairness
and improve their \emph{QoE}.
\item We proposed a novel \emph{QoE} model for the DASH users by taking the current buffer length,
reference buffer length, and video quality into account.
\item We designed an efficient distributed iterative algorithm to obtain the Nash Equilibrium
of the game by additional HTTP sessions between the server and users,
the stability of which is theoretically analyzed.
\end{itemize}

The rest of this paper is organized as follows. In Section
\uppercase\expandafter{\romannumeral2}, the related work on rate
adaptation methods for DASH is presented. In Section
\uppercase\expandafter{\romannumeral3}, a user \emph{QoE} model is
proposed, and the corresponding non-cooperative game is formulated.
Besides that, the existence of Nash Equilibrium and the stability of
the non-cooperative game are also demonstrated. Simulation and
realistic experimental results are given in Section
\uppercase\expandafter{\romannumeral4} and
\uppercase\expandafter{\romannumeral5}, respectively. Finally,
Section \uppercase\expandafter{\romannumeral6} concludes the paper
and discusses the limitations of the proposed method.


\section{Related Work}
In order to adapt to the varying bandwidths, a straightforward method
is to estimate the bandwidth or throughput of the transmission link.
\emph{Thang et al}. \cite{R11} proposed a channel throughput
estimation-based adaptive request method to deal with short-term
bandwidth fluctuations and stabilize the bitrates of segments.
\emph{Romero} \cite{R12} developed a Java client for HTTP streaming
on the Android platform and proposed a smoothed throughput estimation
method to cope with short-term fluctuations. But the user's buffer
length (evaluated by remaining playback time) has not been
considered. In \cite{R13}, a round-trip time and previous values of
the instant throughput based throughput estimation method is proposed
for adaptive streaming so as to stabilize both the bitrates of
segments and the buffer length of users. \emph{Liu et al}. \cite{R14}
developed a throughput estimation based rate adaptation method by
using the ratio of the expected segment fetch time (ESFT) and the
measured segment fetch time. However, in practical applications,
since bandwidth and throughput are affected by a lot of factors, it
is a non-trivial task to estimate them accurately. \emph{Huang et
al}. \cite{R15} showed that inaccurate throughput estimation at the
user side can cause the degeneration of the video quality. Recently,
\emph{Mao et al}. \cite{RR16} proposed a deep reinforcement learning
based rate adaptation algorithm by accurately estimating the channel
throughput accurately.

Besides, in order to improve the \emph{QoE} \cite{RR17}\cite{RR18} of
users directly, some researchers proposed dynamic bitrate selection
methods based on \emph{QoE} maximization \cite{R16}-\cite{R19}.
\emph{Zhang et al}. \cite{R16} proposed a buffer management-based
\emph{QoE} model for HTTP adaptive bitrate streaming and formulated
the adaptive request mechanisms as a constrained convex optimization
problem which is then solved by the Lagrange multiplier method.
\emph{Gheibi et al}. \cite{R17} proposed a \emph{QoE} metric by
considering the probability of interruption in media playback and the
number of initial buffered packets (initial waiting time) for
streaming media applications. However, \emph{QoE} is not only
influenced by buffer length, but also by the requested bitrate and
bitrate switching frequency, etc. \emph{Xu et al}. \cite{R18} modeled
the user \emph{QoE} as a combination of bitrate, starvation
probability of playback buffer, and continuous playback time, and
they proposed two bitrate switching algorithms based on the channel
variation and buffer length. \emph{Rodr\'{\i}guez et al}. \cite{R19}
proposed a non-reference \emph{QoE} metric for DASH by taking initial
buffer delay, temporal playback interruptions, and video resolution
changes into account. In addition, a Markov decision-based rate
adaptation scheme for DASH aiming to maximize the user \emph{QoE}
under time-varying channel conditions was proposed by \emph{Zhou et
al}. \cite{R20} in which the video quality level, bitrate switching
frequency and amplitude, buffer length, etc. are considered
comprehensively. Similarly, \emph{Mart\'{\i}n et al}. \cite{R21}
proposed a \emph{Q}-Learning-based bitrate request method to
efficiently control the selection of the segment quality by
diminishing the quality switches and the occurrence of playback
interruption. \emph{Bokani et al}. \cite{R22} proposed another Markov
Decision Process-based rate adaptation method based on
\emph{Q}-Learning to gradually learn the optimal decisions in order
to avoid playback interruption, which has been found as the most
important factor affecting user \emph{QoE}.

It is worth noting that all of the abovementioned methods are
designed based on the assumption that multiple DASH users make their
rate adaptation decisions \emph{separately}. In practical
applications, it is more common that multi-users request multimedia
content simultaneously from a single server or a relay server of a
cell. Since the export bandwidth of the server is limited, a natural
problem concerns how to allocate the limited bandwidth \emph{jointly}
to users so as to improve the performance of the whole system,
especially to guarantee the fairness of multi-users. In \cite{R23},
\emph{Jiang et al}. proposed an optimized bandwidth estimator based
on the mean of the previous bandwidths for multi-users to increase
bandwidth utilization and stabilize the buffer lengths of
multi-users. \emph{Li et al}. \cite{R24} demonstrated that the
discrete nature of the video bitrates leads to video bitrate
oscillation that negatively affects the video viewing experience and
presented a probe-and-adapt bandwidth estimation approach to further
increase the bandwidth utilization and stabilize the requested video
bitrates of multi-users. \emph{Essaili et al}. \cite{R25} proposed
\emph{QoE}-maximization based traffic and resource management in a
mobile network for multi-user adaptive HTTP streaming. However, a
proxy is needed to intercept and rewrite the user HTTP requests in
this method, which increases the system's complexity and the channel
information for each user is based on the average channel statistics
in the previous second that may not be accurate enough.


\section{Proposed Rate Adaptation Algorithm}

\begin{figure}
\setlength{\abovecaptionskip}{0.cm}
\setlength{\belowcaptionskip}{-0.cm} \centering
\includegraphics[width=8.76cm]{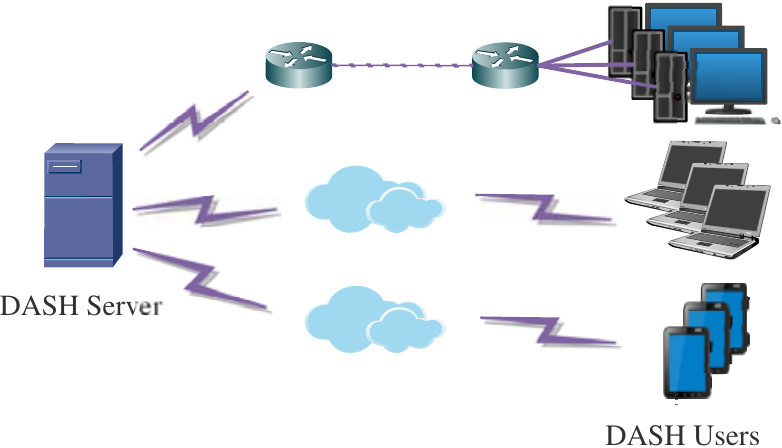}
\caption{Scenario for adaptive HTTP video delivery.} \label{fig2}
\end{figure}

As shown in Fig. \ref{fig2}, in a DASH-based video delivery system,
multi-users compete the limited export bandwidth of the server, and
the information (i.e., the requested bitrate of the last segment and
the current buffer length) of users is unknown by each other. The
DASH users first send their current buffer lengths to the server, and
then request video segments based on the payoff variation information
that is calculated by the server based on the server export bandwidth
and the buffer information of each user. We formulate the rate
adaptation algorithm into a non-cooperative game as follows.

The \textbf{players} in the game are the DASH users. The
\textbf{strategy} of each player is the requested bitrate (denoted by
$r_{i}$ for the $i$-th user). The \textbf{payoff} or \textbf{profit}
for the $i$-th user (denoted by $U_{i}$) is its \textbf{\emph{QoE}}
determined by the requested video and accumulated buffer. The
\textbf{commodity} of the competition is the video segments encoded
into different bitrates. The solution of this game is Nash
Equilibrium. Let $G=\{I,\{\mathbf{R}_{i}\},\{U_{i}(\cdot)\}\}$ denote
the non-cooperative rate adaptation game where $I=\{1,2,\cdots,N\}$
is the index set for the users in the DASH system, and
$\mathbf{R}_{i}$ and $U_{i}(\cdot)$ are the strategy space and
utility function of the $i$-th user, respectively. Each user
determines the required bitrate $r_{i}$ such that
$r_{i}\in\mathbf{R}_{i}$. Let the rate vector
$\mathbf{r}=\{r_{1},\cdots,r_{i},\cdots,r_{N}\}$ denote the outcome
of the game in terms of the requested bitrates of all users. The
resulting utility for the $i$-th user is $U_{i}(\mathbf{r})$. We will
occasionally use $U_{i}(r_{i},\mathbf{r}_{-i})$  to replace
$U_{i}(\mathbf{r})$ to indicate the dependence among DASH users,
where $\mathbf{r}_{-i}$ denotes the vector that consists of elements
of $\mathbf{r}$ without the $i$-th element, i.e.,
$\mathbf{r}_{-i}=\{r_{1},\cdots,r_{i-1},r_{i+1},\cdots,r_{N}\}$.

In the following subsections, we first propose a novel \textit{QoE}
model for DASH users by taking the current buffer length, reference
buffer length, and video quality into account. Then, the existence of
Nash Equilibrium of the non-cooperative game is theoretically proven.
Finally, a distributed iterative algorithm with stability analysis is
proposed to find the Nash Equilibrium.


\subsection{Modeling of DASH User QoE}
In a DASH based video delivery system, the user QoE depends on both
the qualities of received video segments and playback interruptions.


\emph{1) Video quality:} The video quality is directly determined by
the bitrate of the requested video.

Although there are a lot of video quality-bitrate models, most of
them can be uniformly represented as a logarithmic function of
bitrate \cite{R26}, i.e.,
\begin{equation}\label{E1}
    q_{i}(r_{i})=\alpha_{i}\log(1+\beta_{i}r_{i}),
\end{equation}
where $q_{i}$ is the quality of the received video segment measured
in peak-signal-noise-ratio (PSNR), structure similarity index metric
(SSIM) \cite{R27}, etc., and $\alpha_{i}$ and $\beta_{i}$ are
parameters depending on video content. Therefore, without loss of
generality, Eq. \eqref{E1} is used to evaluate the quality of
received video segments in the proposed \emph{QoE} model.

\emph{2) Playback interruption:} We evaluate the influence of
playback interruptions on user \emph{QoEs} by explicitly modeling the
relationship between the estimated buffer length and requested video
bitrate. Usually, for the $i$-th user, the buffer variation can be
calculated as the difference between the cumulated buffer length
(i.e., the remaining video playback time) caused by the downloaded
video segment and the consumed buffer length caused by the played
video content during the download time \cite{R16}\cite{R28}, i.e.,
\begin{equation}\label{E2}
   \Delta b_{i}(r_{i})=T-T\cdot{r_{i}/B_{W}^{i}},
\end{equation}
where $\Delta b_{i}(r_{i})$ is the buffer variation caused by the
requested video bitrate $r_{i}$, $T$ is the length of a video
segment, and $B_{W}^{i}$ is the available channel bandwidth of the
$i$-th user. Unfortunately, as aforementioned, the channel state of
each user is unknown, resulting in Eq. \eqref{E2} being worthless.
But the export bandwidth of the server is known to users. When the
summation of the requested video bitrates of all users is larger than
the export bandwidth of the server, the download time of all users
will increase since the throughput of the system increases.

\begin{figure}
\setlength{\abovecaptionskip}{0.cm}
\setlength{\belowcaptionskip}{-0.cm} \centering
\includegraphics[width=7cm]{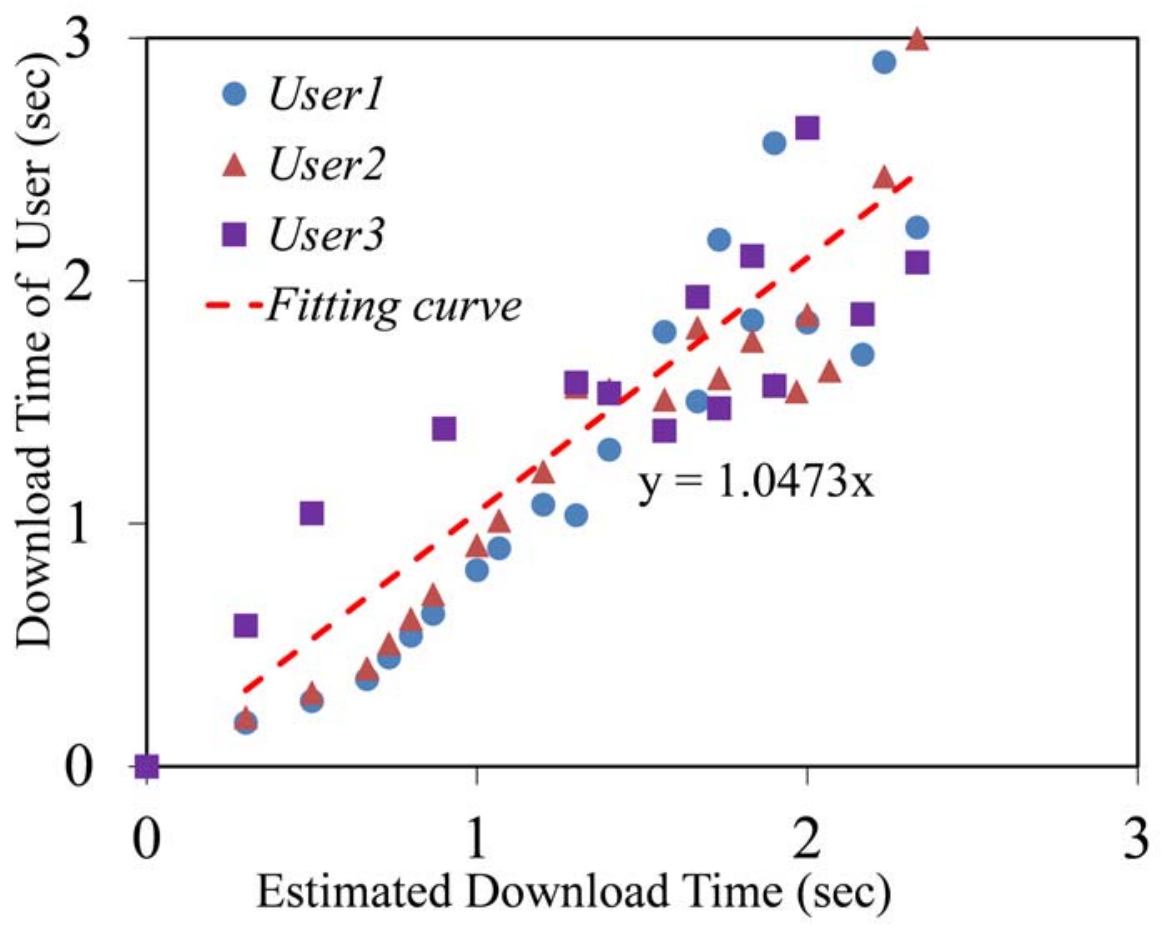}
\caption{Illustration of the relationship between the actual download
time and estimated download time using
$T\cdot\sum\nolimits_{j=1}^{N}{r_{j}/B_{W}}$ of each user. Here,
three users with varied channel throughputs compete the fixed export
bandwidth of a server, and the correlation coefficient is 0.8658.}
\label{fig3}
\end{figure}

Therefore, we use the total requested video bitrates of all the users
over the export bandwidth of the server to estimate the download time
for each user, and the \textbf{\emph{average buffer variation}} of
all users in the system is
\begin{equation}\label{E3}
   \Delta b(\mathbf{r})=T-\omega\cdot
   T\cdot\sum\nolimits_{i=1}^{N}{r_{i}/B_{W}},
\end{equation}
where $\omega$ is a coefficient, and $B_{W}$ is the export bandwidth
of a server. Such a simple approximation can facilitate the following
theoretical analysis with reasonable accuracy. For a single user, the
relationship between the actual download time and the estimated
download time may not exactly be linear, as shown in "\emph{User1}",
"\emph{User2}", and "\emph{User3}" in Fig. \ref{fig3}. But for all
three users, the overall relationship between the actual download
time and estimated download time of all the users can be modeled
using a proportional function with the correlation coefficient of
0.8658.

Furthermore, after the current video segment was downloaded, the
estimated buffer length of the $i$-th user considering other users
denoted as $b_{i}^{est}(r_{i},\mathbf{r}_{-i})$ can be derived by
integrating $\Delta b(\mathbf{r})$ over $r_{i}$:
\begin{equation}\label{E4}
\begin{split}
   &b_{i}^{est}(r_{i},\mathbf{r}_{-i})\\
   &=\int\Delta b(\mathbf{r}){\rm
   d}r_{i}\\
   &=T\cdot r_{i}-\omega\cdot T\cdot \left(\frac{1}{2}r_{i}^{2}+r_{i}\sum\nolimits_{\substack{j=1\\j\neq
   i}}^{N}r_{j}\right)/B_{W}+b_{0}\\
   &=\Phi(r_{i})-\omega\cdot\Psi(\mathbf{r})+b_{0},
\end{split}
\end{equation}
where $b_{0}$ is a constant, denoting the initial average buffer
length of all the users, $\Phi(r_{i})=T\cdot r_{i}$ is the benefit
gained from accumulated buffer, and $\Psi(\mathbf{r})=T\cdot
\left(\frac{1}{2}r_{i}^{2}+r_{i}\sum\nolimits_{\substack{j=1\\j\neq
   i}}^{N}r_{j}\right)/B_{W}$
represents the \textbf{system penalty} (buffer consumption) caused by
the requested bitrates of all the users.

In order to ensure the buffer is equipped with an optimal state,
i.e., the buffer length should be kept within a certain reference
level, an adjustment factor (denoted as $A_{f}$) is adopted to modify
the revenue function $\Phi(r_{i})$, i.e.,
\begin{equation}\label{E5}
   \Phi'(r_{i})=A_{f}\cdot\Phi(r_{i})=A_{f}\cdot T\cdot r_{i}.
\end{equation}
Considering that a larger (resp. smaller) $b_{curr}$ indicates more
aggressive (resp. defensive) behaviors in requesting video segments,
$A_{f}$ is defined as a monotonically increasing function of the
difference between the current buffer length $b_{curr}$ and the
reference buffer length $b_{ref}$ \cite{R28}:
\begin{equation}\label{E6}
   A_{f}=2\cdot\frac{e^{p(b_{curr}-b_{ref}})}{1+e^{p(b_{curr}-b_{ref})}},
\end{equation}
where $p>0$ is a constant, $b_{ref}$ is the predefined reference
buffer length, and $b_{curr}$ is the current known buffer length
before the video segment was downloaded for a certain user,
respectively. The difference between $b_{ref}$ and $b_{curr}$ is an
indicator to control the requested bitrates. When $b_{curr}>b_{ref}$,
$A_{f}$ is larger than 1, it indicates that the previously received
video quality may be low. Then, the bitrates requested by users will
be increased to achieve the optimal utility. When $b_{curr}<b_{ref}$,
$A_{f}$ is smaller than 1. This indicates that the probability of
playback interruption is large. Then, the bitrates requested by users
will be decreased to achieve the optimal utility.

Accordingly, the estimated buffer of the $i$-th user with respect to
all the other users can be rewritten as
\begin{equation}\label{E7}
\begin{split}
   &b_{i}^{est}(r_{i},\mathbf{r}_{-i})\\
   &=\Phi'(r_{i})-\omega\cdot\Psi(\mathbf{r})+b_{0}\\
   &=2\cdot\frac{e^{p(b_{curr}-b_{ref})}}{1+e^{p(b_{curr}-b_{ref})}}
   \cdot T\cdot r_{i}\\
   &-\omega\cdot T\cdot\left(\frac{1}{2}r_{i}^{2}+r_{i}\sum\nolimits_{\substack{j=1\\j\neq
   i}}^{N}r_{j}\right)/B_{W}+b_{0}.
\end{split}
\end{equation}

Finally, we define the utility function of user $i$ that considers
the influence on other DASH users sharing the same network resources
as the linear combination of the quality function in \eqref{E1} and
the buffer function in \eqref{E7}, i.e.,
\begin{equation}\label{E8}
\begin{split}
   &U_{i}(r_{i},\mathbf{r}_{-i})\\
   &=q_{i}(r_{i})+\mu\cdot b_{i}^{est}(r_{i},\mathbf{r}_{-i})\\
   &=\alpha_{i}\log(1+\beta_{i}r_{i})+\mu\left(2\cdot\frac{e^{p(b_{curr}-b_{ref})}}{1+e^{p(b_{curr}-b_{ref})}}
   \cdot T\cdot r_{i}\right)\\
   &-\mu\cdot\omega\cdot T\cdot\left(\frac{1}{2}r_{i}^{2}+r_{i}\sum\nolimits_{\substack{j=1\\j\neq i}}^{N}r_{j}\right)/B_{W}+\mu\cdot
   b_{0}\\
   &=\alpha_{i}\log(1+\beta_{i}r_{i})+\mu\left(2\cdot\frac{e^{p(b_{curr}-b_{ref})}}{1+e^{p(b_{curr}-b_{ref})}}
   \cdot T\cdot r_{i}\right)\\
   &-\nu\cdot T\cdot\left(\frac{1}{2}r_{i}^{2}+r_{i}\sum\nolimits_{\substack{j=1\\j\neq i}}^{N}r_{j}\right)/B_{W}+\mu\cdot
   b_{0},
\end{split}
\end{equation}
where $\mu>0$ is a weight to balance the two parts, and
$\nu=\mu\cdot\omega$.


\subsection{Proof of the Existence of Nash Equilibrium}
The Nash Equilibrium of a game is a strategy profile with the
property in which no player can increase its utility by choosing a
different action when the other players' actions are given
\cite{R29}. A Nash Equilibrium exists for a game $G$ if the following
two conditions are met:

a)  the strategy space $\mathbf{R}_{i}$ is a non-empty, convex, and
compact subset of Euclidean space $\mathbf{R}^{N}$;

b)  the utility $U_{i}(\mathbf{r})$ is continuous in $\mathbf{r}$ and
(at least) quasi-concave in $r_{i}$.

For the proposed non-cooperative game, the strategy space is composed
of the user requested bitrates with the range of $[0,r^{max}]$, where
$r^{max}$ is the maximum requested video bitrate of the $i$-th user.
Thereby, there is no doubt that $\mathbf{R}_{i}$ is a non-empty and
compact subset of Euclidean space $\mathbf{R}^{N}$. According to the
definition of the convex set \cite{R30}, for any
$r_{x},r_{y}\in\mathbf{R}_{i}$ and any $\zeta$ with
$0\leq\zeta\leq1$, we have $0\leq\zeta r_{x}\leq \zeta r^{max}$ and
$0\leq(1-\zeta)r_{y}\leq(1-\zeta)r^{max}$. Then, we can get
$0\leq\zeta r_{x}+(1-\zeta)r_{y}\leq r^{max}$. Therefore, $\zeta
r_{x}+(1-\zeta)r_{y}\in\mathbf{R}_{i}$, indicating $\mathbf{R}_{i}$
is a convex set. Thus, condition a) is satisfied. For the utility
function in \eqref{E8}, it is obviously a continuous function in
terms of $\mathbf{r}$. Besides, the second derivatives of
$U_{i}(\mathbf{r})$ with respect to all the $r_{i}$ are
\begin{equation}\label{E9}
\begin{cases}
    \frac{\partial^{2}U_{i}}{\partial r_{i}^{2}}=-\frac{\alpha\beta^{2}}{(1+\beta r_{i})^{2}}-\frac{\nu\cdot T}{B_{W}},\\
    \frac{\partial^{2}U_{i}}{\partial r_{i}\partial r_{j}}=-\frac{\nu\cdot
    T}{B_{W}}|_{i\neq j},
\end{cases}
\end{equation}
which are negative because $\alpha$, $\beta$, $\nu$, $T$, and $B_{W}$
are non-negative. Accordingly, the utility $U_{i}(\mathbf{r})$ is a
strictly concave function of $r_{i}$ (for all $i$) \cite{R30}.
Therefore, there must exist a Nash Equilibrium in the proposed rate
adaptation game.

For a non-cooperative game, the Nash Equilibrium can be achieved by
jointly maximizing the utility functions of all players, i.e., the
corresponding best response function of a player is defined as the
strategy of the player with those of other players fixed:
\begin{equation}\label{E10}
    \mathbf{B}_{i}(\mathbf{r}_{-i})=\arg\max_{r_{i}\in\mathbf{R}_{i}}U_{i}(r_{i},\mathbf{r}_{-i}).
\end{equation}
When the Nash Equilibrium is achieved, the strategies of all players
can be represented as
$\mathbf{r}^{\ast}=\{r_{1}^{\ast},r_{2}^{\ast},\cdots,r_{N}^{\ast}\}$,
where $r_{i}^{\ast}=\mathbf{B}_{i}(\mathbf{r}_{-i}^{\ast})$ is the
optimal strategy of the $i$-th user and
$\mathbf{r}_{-i}^{\ast}=\{r_{1}^{\ast},\cdots,r_{i-1}^{\ast},r_{i+1}^{\ast},\cdots,r_{N}^{\ast},\}$
is the set of the Nash Equilibrium of all users except user $i$.


\subsection{Distributed Iterative Algorithm for Nash Equilibrium}
Theoretically, the Nash Equilibrium can be obtained by solving the
following equations:
\begin{equation}\label{E11}
\begin{aligned}
    \frac{\partial U_{i}(\mathbf{r})}{\partial
    r_{i}}=&\frac{\alpha_{i}\beta_{i}}{1+\beta_{i}r_{i}}+\mu\cdot
    T\cdot\frac{2e^{p(b_{curr}-b_{ref})}}{1+e^{p(b_{curr}-b_{ref})}}\\
    &-\nu\cdot
    T\cdot\frac{\sum\nolimits^{N}_{j=1}r_{j}}{B_{W}}=0, \quad\forall i.
\end{aligned}
\end{equation}
Taking a DASH system with only 2 users as an example, we have,
\begin{equation}\label{E12}
\begin{cases}
    \frac{Z_{1,1}}{1+\beta_{1}r_{1}}+Z_{2,1}-Z_{3}(r_{1}+r_{2})=0,\\
    \frac{Z_{1,2}}{1+\beta_{2}r_{2}}+Z_{2,2}-Z_{3}(r_{1}+r_{2})=0,
\end{cases}
\end{equation}
where
\begin{equation}\label{E13}
\begin{cases}
    Z_{1,1}=\alpha_{1}\beta_{1},Z_{2,1}=\mu\cdot T\cdot\frac{2e^{p(b_{curr,1}-b_{ref})}}{1+e^{p(b_{curr,1}-b_{ref})}},\\
    Z_{1,2}=\alpha_{2}\beta_{2},Z_{2,2}=\mu\cdot
    T\cdot\frac{2e^{p(b_{curr,2}-b_{ref})}}{1+e^{p(b_{curr,2}-b_{ref})}},\\
    Z_{3}=\nu\cdot T/B_{W}.
\end{cases}
\end{equation}
Assume the two users request the same video and have the same
channel condition (i.e., two identical users), we have
$Z_{1,1}=Z_{1,2}=Z_{1}$ and $Z_{2,1}=Z_{2,2}=Z_{2}$. Then, the Nash
Equilibrium can be expressed as
\begin{equation}\label{E14}
    r_{1}^{\ast}=r_{2}^{\ast}=\frac{-(2Z_{3}-\beta_{i}Z_{2})+\sqrt{(2Z_{3}+\beta_{i}Z_{2})^{2}+8\beta_{i}Z_{1}Z_{3}}}{4\beta_{i}Z_{3}}.
\end{equation}

From the above analysis, to determine the requested video bitrate for
a certain user, the strategies of the other users must be available.
However, such user strategies and information are unknown to each
other in a practical DASH system. In order to adjust the requested
bitrate $r_{i}$ for the $i$-th user, we propose to employ its own
information (i.e., the requested bitrate of the last segment and the
current buffer length) and communicate with the server to obtain the
payoff variation that is induced by the varied download time.
Therefore, the requested video bitrate $r_{i}$ can be updated based
on the sub-gradient method \cite{R31}\cite{R32}\cite{R33}:
\begin{equation}\label{E15}
    r_{i}(t+1)=r_{i}(t)+\theta_{i}r_{i}(t)\frac{\partial U_{i}(\mathbf{r})}{\partial
    r_{i}(t)},
\end{equation}
where $\theta_{i}>0$ is the speed adjustment parameter (i.e.,
learning rate) of user $i$.

In an actual system, the value of $\partial
U_{i}(\mathbf{r})/\partial
    r_{i}(t)$
(i.e., the payoff variation information in Fig. \ref{fig2}) can be
estimated by the server and transmitted to the user as,
\begin{equation}\label{E16}
    \frac{\partial U_{i}(\mathbf{r})}{\partial
    r_{i}(t)}\approx\frac{U_{i}^{+}(\mathbf{r}^{+})-U_{i}^{-}(\mathbf{r}^{-})}{2\varepsilon},
\end{equation}
with
\begin{equation}\label{E17}
\begin{cases}
   \mathbf{r}^{+}=\left\{r_{1}(t),\cdots,r_{i}(t)+\varepsilon,\cdots,r_{N}(t)\right\},\\
   \mathbf{r}^{-}=\left\{r_{1}(t),\cdots,r_{i}(t)-\varepsilon,\cdots,r_{N}(t)\right\},
\end{cases}
\end{equation}
where $\varepsilon$ is an especially small value (e.g.,
$\varepsilon=0.0001$). When the Nash Equilibrium is achieved, we have
$r_{i}(t+1)=r_{i}(t)$ for any $i$, i.e.,
\begin{equation}\label{E18}
\begin{cases}
   \textbf{r}(t+1)=\textbf{r}(t),\\
   \frac{\partial U_{i}(\mathbf{r})}{\partial
    r_{i}(t)}=0.
\end{cases}
\end{equation}


\subsection{Stability Analysis for the Distributed Iterative Algorithm}
The stability of the distributed strategy update algorithm
\eqref{E15} is analyzed by using the \emph{Routh-Hurvitz} stability
condition \cite{R34}\cite{R35}\cite{R36}\cite{R37}, which judges the
distribution of the eigenvalues (denoted as $\lambda_{i}$) of the
\textbf{\emph{Jacobian matrix}}. That is, if all of the eigenvalues
are inside a unit circle of the complex plane (i.e.,
$|\lambda_{i}|<1$ ), the Nash Equilibrium point is stable. Taking a
DASH system with only two users as an example, the
\textbf{\emph{Jacobian matrix}} can be expressed as,
\begin{equation}\label{E19}
    \mathbf{J}(r_{1},r_{2})=\left[\begin{array}{ccc}\frac{\partial r_{1}(t+1)}{\partial
    r_{1}(t)}&\frac{\partial r_{1}(t+1)}{\partial
    r_{2}(t)}\\
    \frac{\partial r_{2}(t+1)}{\partial
    r_{1}(t)}&\frac{\partial r_{2}(t+1)}{\partial
    r_{2}(t)}\end{array}\right]=\left[\begin{array}{ccc}j_{1,1}&j_{1,2}\\
    j_{2,1}&j_{2,2}\end{array}\right],
\end{equation}
where
\begin{equation}\label{E20}
\begin{cases}
    j_{1,2}=-\theta_{1}Z_{3}r_{1}\\
    j_{2,1}=-\theta_{2}Z_{3}r_{2}\\
    j_{1,1}=1+\theta_{1}\left(-\frac{\beta_{1}Z_{1,1}r_{1}}{(1+\beta_{1}r_{1})^{2}}+\frac{Z_{1,1}}{1+\beta_{1}r_{1}}+Z_{2,1}-Z_{3}(2r_{1}+r_{2})\right)\\
    j_{2,2}=1+\theta_{2}\left(-\frac{\beta_{2}Z_{1,2}r_{2}}{(1+\beta_{2}r_{2})^{2}}+\frac{Z_{1,2}}{1+\beta_{2}r_{2}}+Z_{2,2}-Z_{3}(r_{1}+2r_{2})\right)
\end{cases}
\end{equation}
The two eigenvalues can be obtained by solving the characteristic
equation:
\begin{equation}\label{E21}
    \lambda^{2}-\lambda(j_{1,1}+j_{2,2})+(j_{1,1}j_{2,2}-j_{1,2}j_{2,1})=0,
\end{equation}
whose solution is
\begin{equation}\label{E22}
\begin{cases}
    \lambda_{1}=\frac{(j_{1,1}+j_{2,2})+\sqrt{(j_{1,1}-j_{2,2})^{2}+4j_{1,2}j_{2,1}}}{2},\\
    \lambda_{2}=\frac{(j_{1,1}+j_{2,2})-\sqrt{(j_{1,1}-j_{2,2})^{2}+4j_{1,2}j_{2,1}}}{2}.
\end{cases}
\end{equation}
Assume the two users are identical (i.e.,
$Z_{1,1}=Z_{1,2}=Z_{1}=\alpha\beta$ and $Z_{2,1}=Z_{2,2}=Z_{2}$), and
when the Nash equilibrium point is achieved (i.e.,
$r_{1}^{\ast}=r_{2}^{\ast}$) and the buffer length is in a steady
state (i.e., $b_{curr}=b_{ref}$ and $Z_{2,1}=Z_{2,2}=Z_{2}=\mu\cdot
T$), we can derive that $j_{1,1}=j_{2,2}$, $j_{1,2}=j_{2,1}$,
$\lambda_{1}=j_{1,1}-j_{1,2}$, and $\lambda_{2}=j_{1,1}+j_{1,2}$.
Therefore, the condition to ensure the stability of the proposed
algorithm is expressed as
\begin{equation}\label{E23}
\begin{cases}
    -1<j_{1,1}-j_{1,2}<1,\\
    -1<j_{1,1}+j_{1,2}<1.
\end{cases}
\end{equation}
For $|j_{1,1}-j_{1,2}|<1$, substituting Eq. \eqref{E20} into
\eqref{E23}, we have
\begin{equation}\label{E24}
\begin{aligned}
    -2<&\theta_{1}\left[-\frac{\beta Z_{1}r_{1}}{(1+\beta r_{1})^{2}}+\frac{Z_{1}}{1+\beta
    r_{1}}+Z_{2}-Z_{3}(2r_{1}+r_{2})\right]\\
    &+\theta_{1}Z_{3}r_{1}<0.
\end{aligned}
\end{equation}
At the Nash Equilibrium point, since $\theta_{1}=
\theta_{2}=\theta^{\ast}$, and $r_{1}=r_{2}=r^{\ast}$, Eq.
\eqref{E24} can be rewritten as
\begin{equation}\label{E25}
\begin{aligned}
    -\frac{2}{\theta^{\ast}}<-\frac{\beta Z_{1}r^{\ast}}{(1+\beta r^{\ast})^{2}}+\frac{Z_{1}}{1+\beta
    r^{\ast}}+Z_{2}-2Z_{3}r^{\ast}<0.
\end{aligned}
\end{equation}
Furthermore, Eq. \eqref{E25} can be simplified as
\begin{equation}\label{E26}
\begin{cases}
    Z_{1}+Z_{2}(1+\beta r^{\ast})^{2}<2Z_{3}r^{\ast}(1+\beta r^{\ast})^{2},\\
    Z_{1}+\left(Z_{2}+\frac{2}{\theta^{\ast}}\right)(1+\beta r^{\ast})^{2}>2Z_{3}r^{\ast}(1+\beta r^{\ast})^{2}.
\end{cases}
\end{equation}
Substituting $T = 2$, $Z_{1}$, $Z_{2}$, and $Z_{3}$ into \eqref{E26},
we can obtain
\begin{equation}\label{E27}
\begin{cases}
    \alpha\beta+2\mu(1+\beta r^{\ast})^{2}<\frac{4\nu}{B_{W}}r^{\ast}(1+\beta r^{\ast})^{2},\\
    \alpha\beta+\left(2\mu+\frac{2}{\theta^{\ast}}\right)(1+\beta r^{\ast})^{2}>\frac{4\nu}{B_{W}}r^{\ast}(1+\beta r^{\ast})^{2}.
\end{cases}
\end{equation}
Similarly, for $|j_{1,1}+j_{1,2}|<1$, the stability condition is,
\begin{equation}\label{E28}
\begin{cases}
    \alpha\beta+2\mu(1+\beta r^{\ast})^{2}<\frac{8\nu}{B_{W}}r^{\ast}(1+\beta r^{\ast})^{2},\\
    \alpha\beta+\left(2\mu+\frac{2}{\theta^{\ast}}\right)(1+\beta r^{\ast})^{2}>\frac{8\nu}{B_{W}}r^{\ast}(1+\beta r^{\ast})^{2}.
\end{cases}
\end{equation}
Since $\alpha$, $\beta$, $\mu$, $\nu$, $\theta^{\ast}$, and $B_{W}$
are all positive, we can conclude that the proposed distributed
iterative updating algorithm is stable if the following conditions
are satisfied:
\begin{equation}\label{E29}
\begin{cases}
    \alpha\beta+2\mu(1+\beta r^{\ast})^{2}<\frac{4\nu}{B_{W}}r^{\ast}(1+\beta r^{\ast})^{2},\\
    \alpha\beta+\left(2\mu+\frac{2}{\theta^{\ast}}\right)(1+\beta r^{\ast})^{2}>\frac{8\nu}{B_{W}}r^{\ast}(1+\beta r^{\ast})^{2}.
\end{cases}
\end{equation}

When there are more than two users in the system, \eqref{E30} must be
satisfied for user $i$,
\begin{equation}\label{E30}
    \frac{Z_{1,i}}{1+\beta_{i}r_{i}}+Z_{2,i}-Z_{3}\left(\sum^{N}_{j=1}r_{j}\right)=0,
\end{equation}
Equation (30) of all users can be expressed by matrix format as
\begin{equation}\label{E31}
    \mathbf{Z}_{1}+(\mathbf{1}+\mathbf{r}\cdot\upbeta)\mathbf{Z}_{2}=Z_{3}(\mathbf{r}\cdot \mathbf{1}^{T})(\mathbf{1}+\mathbf{r}\cdot\upbeta),
\end{equation}
where
\begin{equation}\label{E32}
\begin{cases}
    \mathbf{r}=[r_{1}\cdots r_{i}\cdots r_{N}],\\
    \mathbf{1}=[1\cdots 1\cdots 1],\\
    \mathbf{Z}_{1}=[Z_{1,1}\cdots Z_{1,i}\cdots Z_{1,N}],\\
    \upbeta=diag(\beta_{1},\cdots,\beta_{N}),\\
    \mathbf{Z}_{2}=diag(Z_{2,1},\cdots,Z_{2,N}).
\end{cases}
\end{equation}
The \textit{\textbf{Jacobian matrix}} of the Nash Equilibrium for
multi-users is given as
\begin{equation}\label{E33}
    \mathbf{J}=\left[\begin{array}{ccc}\frac{\partial r_{1}(t+1)}{\partial
    r_{1}(t)}&\cdots&\frac{\partial r_{1}(t+1)}{\partial
    r_{N}(t)}\\
    \vdots&\ddots&\vdots\\
    \frac{\partial r_{N}(t+1)}{\partial
    r_{1}(t)}&\cdots&\frac{\partial r_{N}(t+1)}{\partial
    r_{N}(t)}\end{array}\right].
\end{equation}
Then, similar to the DASH system with only 2 users, the local
stability condition can also be analyzed.

\begin{table}[htbp]
  \centering
    \begin{tabular}{l}
    \toprule
    \textbf{\textit{Algorithm 1}: Distributed Iterative Algorithm of Rate Adaptation}\\
    \midrule
       1:  Initially, all  users request  the bitrate $r(1)=0.1$Mbps to the server \\
           \quad so as to quickly establish the predefined initial buffer
           length\\
           \quad (e.g. 2s).\\
       2:  \textit{\textbf{while}} $n\leq S$ (\textit{S} is the total number of requested segments) \textit{\textbf{do}}\\
       3:      \quad Each user sends the payoff request to the server;\\
       4:      \quad The server sends the payoff information to corresponding user;\\
       5:      \quad  The users update the requested bitrate $r(n)$ according to (15)\\
               \qquad and request video segments from the server;\\
       6:      \quad The server sends the video segments to the users;\\
       7:      \quad Update the buffer information $b_{curr}(n)$;\\
       8:      \quad $n=n+1$;\\
       9: \textit{\textbf{end while}}\\
    \bottomrule
    \end{tabular}
  \label{tab:addlabel}
\end{table}


\emph{\textbf{Algorithm 1}} shows the   detailed procedure of the
proposed method. The DASH users first request the bitrate of 0.1Mbps
to quickly establish the initial buffer length. Then, the users send
their buffer information to the server. Thirdly, the server
calculates and sends the payoff variation information for each user
based on the export bandwidth and the buffer lengths of users.
Finally, the users update the requested bitrates and request video
segments from the server.


\section{Simulation Results}

\subsection{Simulation Setup}
To verify the performance of the proposed method, the
\textit{\textbf{Lib-DASH}} platform \cite{R38}\cite{R39} is used.
Users request video segments from the server, which is hosting an
Apache HTTP Web server \cite{R40}. And the server export bandwidth is
controlled by \textit{DummyNet} \cite{R41}. The video dataset
includes \textit{BigBuckBunny} \cite{R42}\cite{R43},
\textit{ElephantsDream} \cite{R44}, and \textit{SitaSingstheBlues}
\cite{R45}\cite{R46}. Each video was encoded by \textit{FFMPEG}
\cite{R47} with 20 various bitrates from low to high, as shown in
TABLE \ref{table1}. Fig. \ref{fig4} shows the two parameters of the
quality model in (1) (i.e., $\alpha$ and $\beta$) for each video. The
parameters $\alpha$ and $\beta$ are obtained by fitting Eq.
\eqref{E1} using the actual qualities and bitrates of each segment.
The lengths of each video segment and the initial buffer of each user
are set as 2s. For the proposed method, all users are equipped with
the same learning rate in \eqref{E15}, i.e.,
$\theta_{1}=\cdots=\theta_{N}=\theta$, for all $i$, to ensure the
synchronization of the convergence of the distributed algorithm among
all users in the system. Meanwhile, the initial requested bitrates of
all users are set as 0.1 Mbps.

\begin{table}[htbp]
\setlength{\abovecaptionskip}{0.cm}
\setlength{\belowcaptionskip}{-0.cm}
  \centering
  \caption{Detailed Information of the Tested DASH Dataset}
\includegraphics[width=7cm]{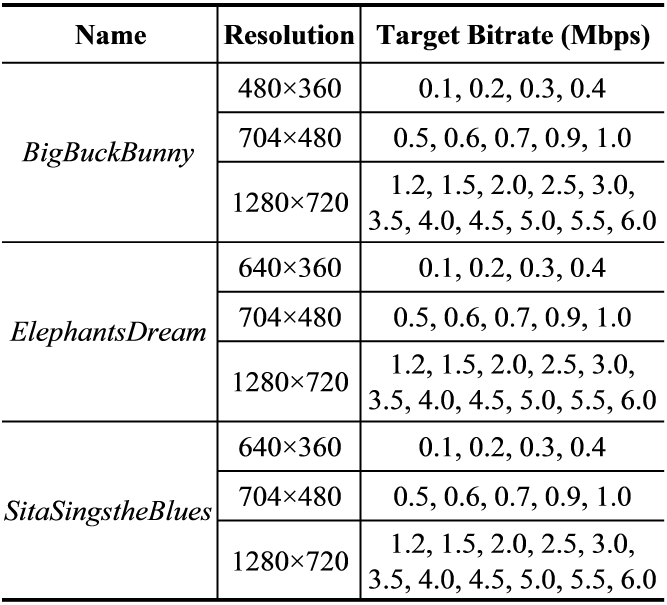}
\label{table1}
\end{table}


The proposed algorithm is validated under four cases:

\textit{\textbf{Case 1}}, two identical users
(\textit{\textbf{without limitations on user channel throughputs}})
request the same video content (i.e., \textit{BigBuckBunny}) with a
\textit{\textbf{fixed}} server export bandwidth;

\textit{\textbf{Case 2}}, two identical users
(\textit{\textbf{without limitations on user channel throughputs}})
request the same video content (i.e., \textit{BigBuckBunny}) with a
\textit{\textbf{varied}} server export bandwidth;

\textit{\textbf{Case 3}}, three users (\textit{\textbf{with fixed
limitations on user channel throughputs}}) request different video
contents (i.e., \textit{User1}, \textit{User2}, and \textit{User3}
request \textit{BigBuckBunny}, \textit{ElephantsDream}, and
\textit{SitaSingstheBlues}, respectively) with a
\textit{\textbf{varied}} server export bandwidth;

\textit{\textbf{Case 4}}, three users (\textit{\textbf{with random
limitations on user channel throughputs}}) request different video
contents with both \textit{\textbf{fixed}} and
\textit{\textbf{varied}} server export bandwidth. Besides we also
compare the \textit{\textbf{Proposed}} method with three algorithms,
i.e., the \textit{Quality\_First} method (\textit{\textbf{QF}})
\cite{R48}, \textit{Buffer\_First} method (\textit{\textbf{BF}})
\cite{R48}, and \textit{QoE-based Buffer-aware Resource Allocation}
method (\textit{\textbf{QBA}}) \cite{R25}.
\begin{figure}
\setlength{\abovecaptionskip}{0.cm}
\setlength{\belowcaptionskip}{-0.cm} \centering \subfigure[]{
\label{fig4:subfig:a} 
\includegraphics[height=2.7cm]{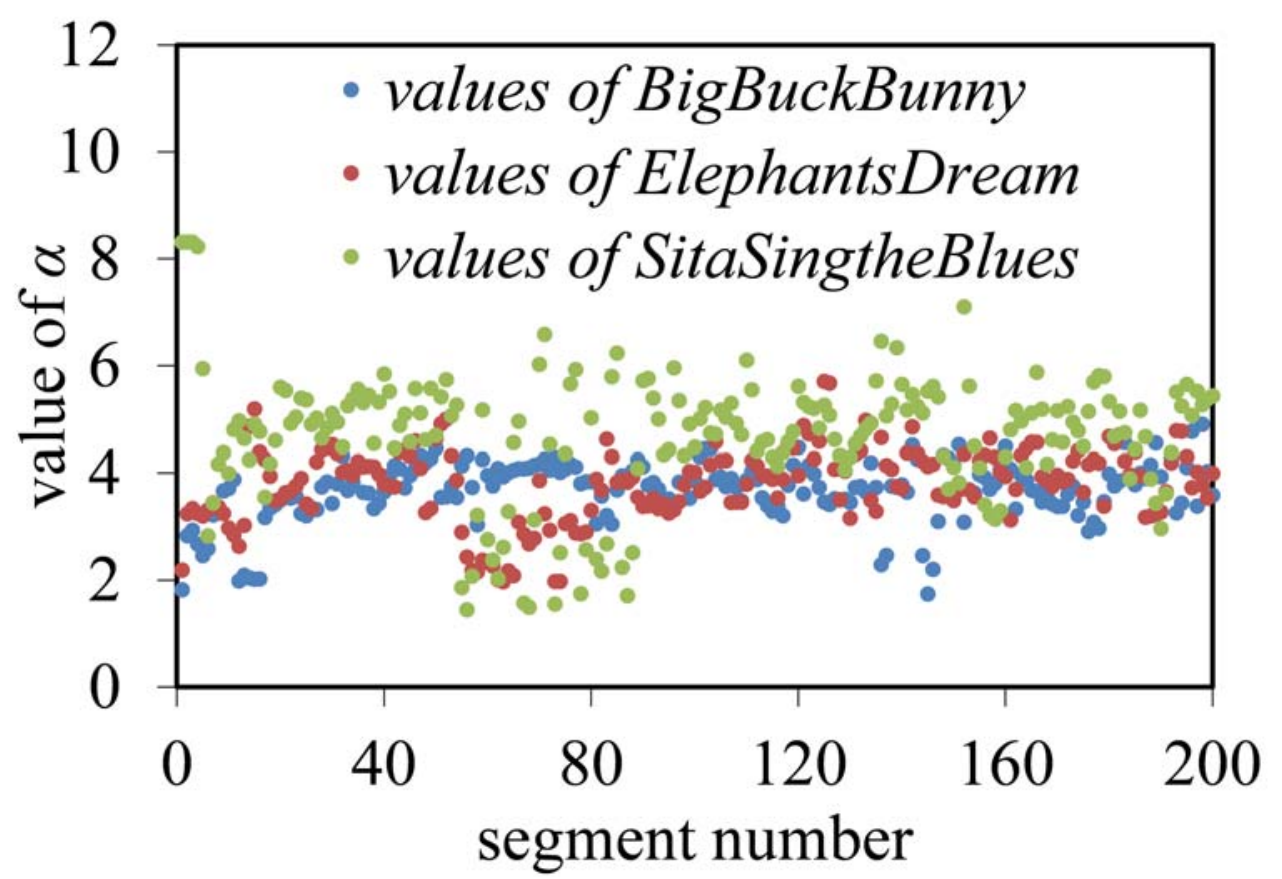}}
\subfigure[]{
\label{fig4:subfig:b} 
\includegraphics[height=2.7cm]{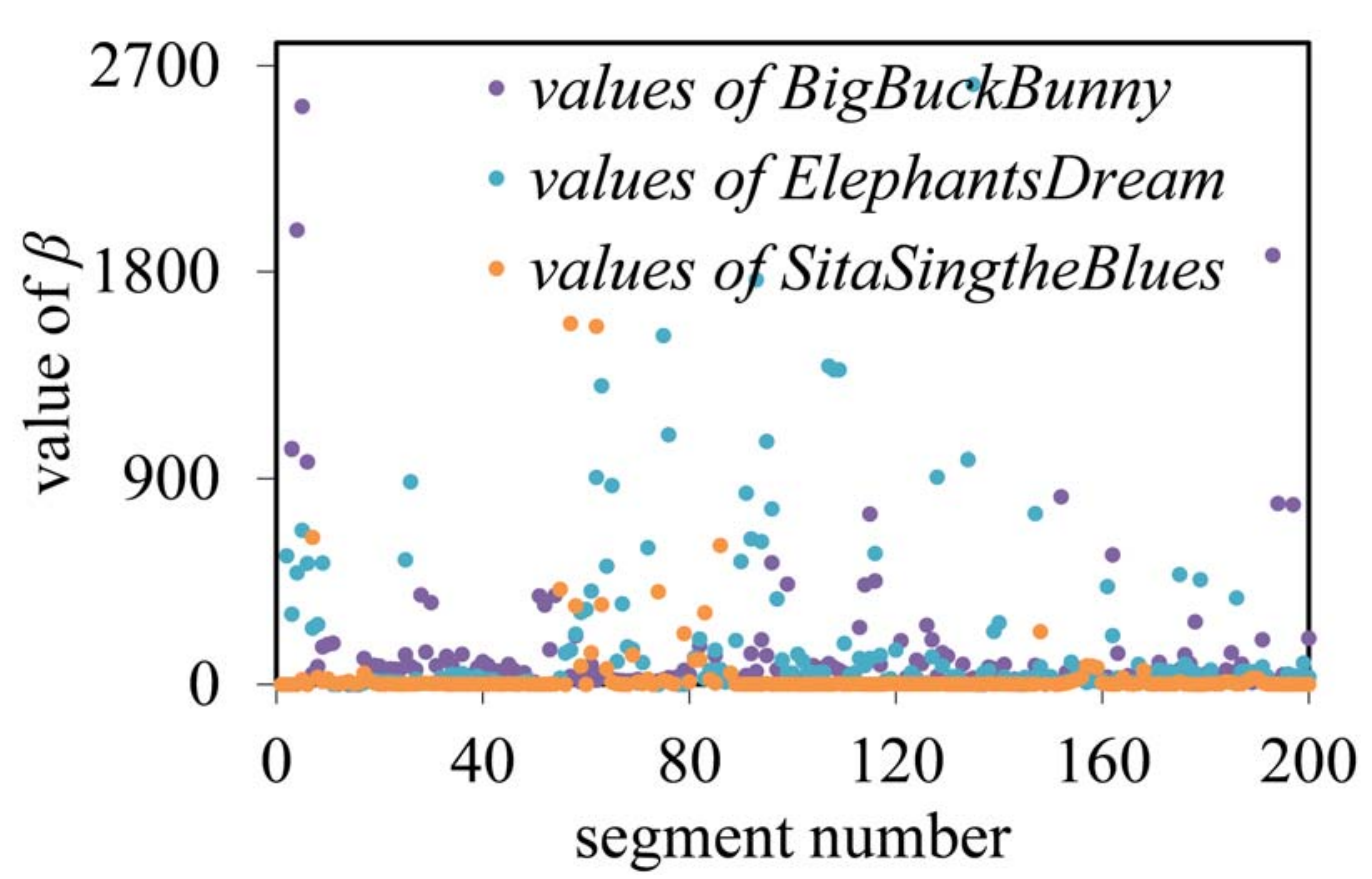}}
\caption{Quality model parameters (a) $\alpha_{1}$, $\alpha_{2}$,
$\alpha_{3}$ and (b) $\beta_{1}$, $\beta_{2}$, $\beta_{3}$ of the
video datasets in the simulation.}
\label{fig4} 
\end{figure}

\begin{figure}
\setlength{\abovecaptionskip}{0.cm}
\setlength{\belowcaptionskip}{-0.cm} \centering \subfigure[]{
\label{fig5:subfig:a} 
\includegraphics[width=4.25cm]{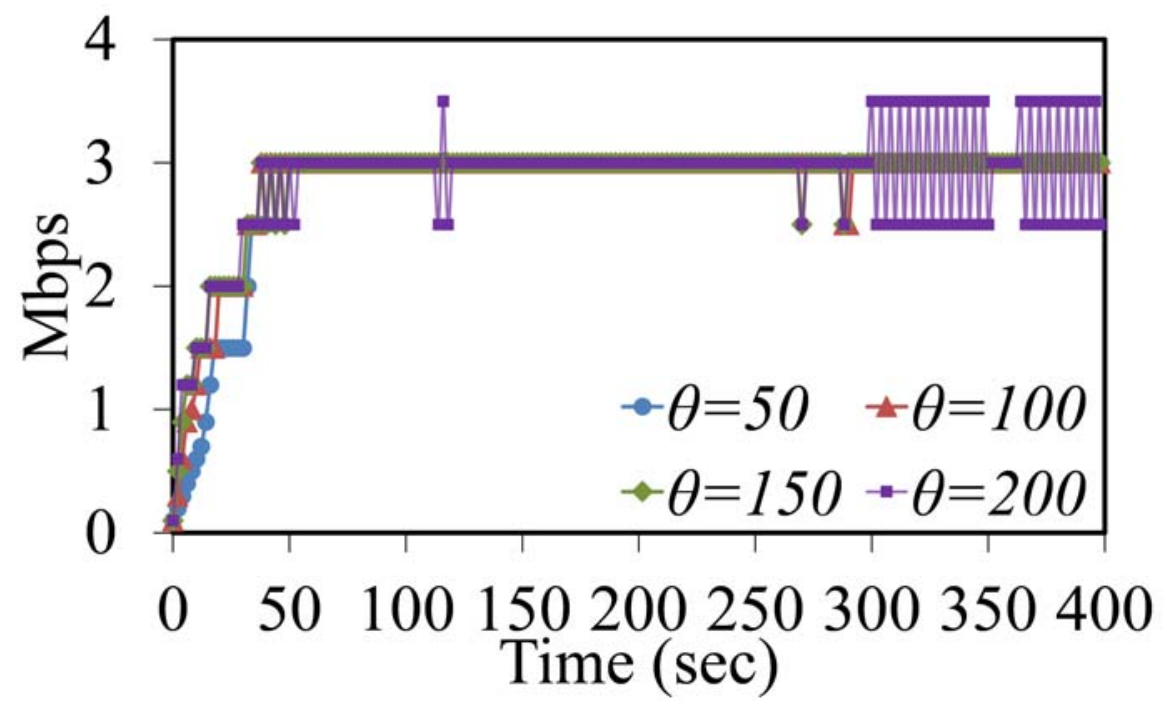}}
\subfigure[]{
\label{fig5:subfig:b} 
\includegraphics[width=4.25cm]{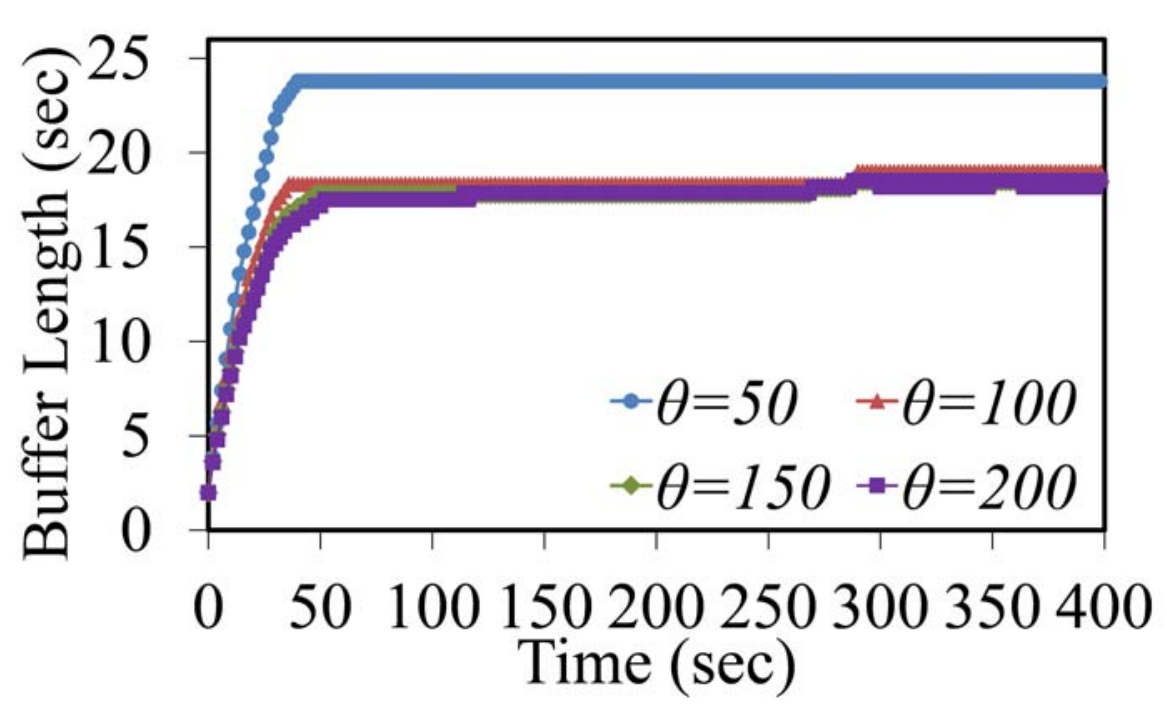}}
\caption{Results of case 1, in which two identical users compete the
server export bandwidth of 6Mbps, and request the \emph{BigBuckBunny}
video sequence with learning rate $\theta=50$, 100, 150, and 200.
Here, $\mu=0.003$, $\nu=0.0041$, $\alpha=2.15$, $\beta=0.0827$,
$r^{\ast}=3$ Mbps, and $B_{W}=6$ Mbps and the inequality relation of
(29) is tenable. (a) dynamic behavior of requested bitrates, (b)
actual buffer lengths of the two users. \emph{\textbf{Note that the
states of the two users are identical}}.}
\label{fig5} 
\end{figure}

\begin{figure}
\setlength{\abovecaptionskip}{0.cm}
\setlength{\belowcaptionskip}{-0.cm} \centering \subfigure[]{
\label{fig6:subfig:a} 
\includegraphics[width=4.25cm]{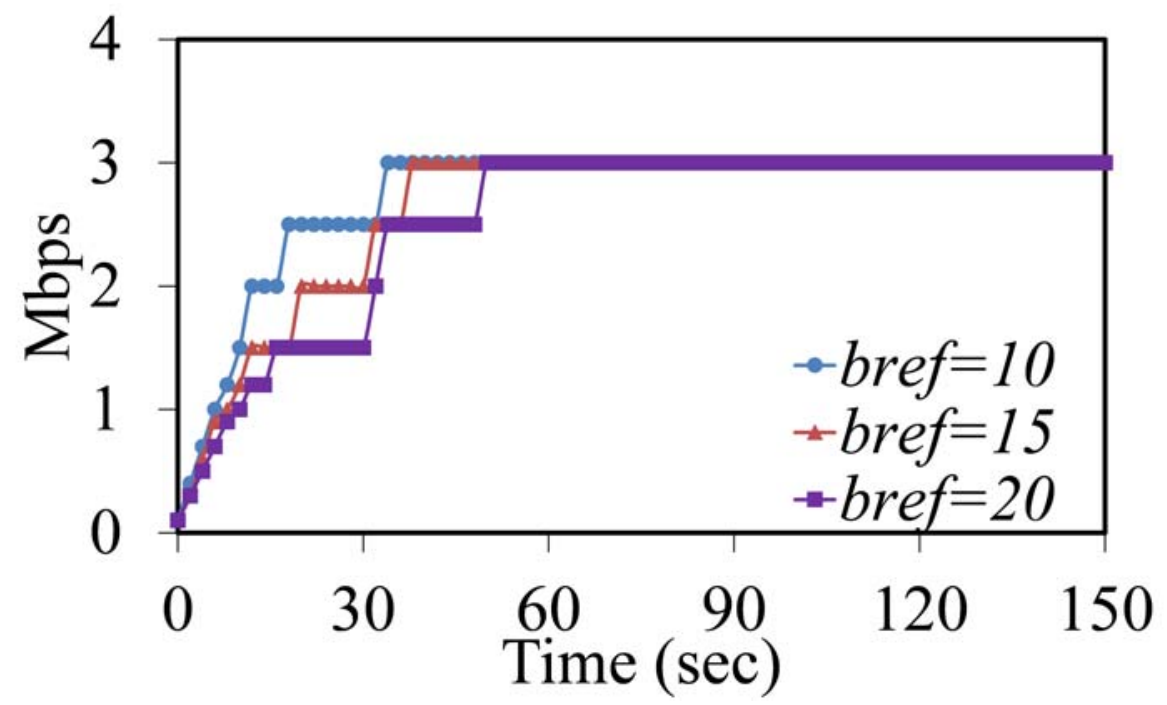}}
\subfigure[]{
\label{fig6:subfig:b} 
\includegraphics[width=4.25cm]{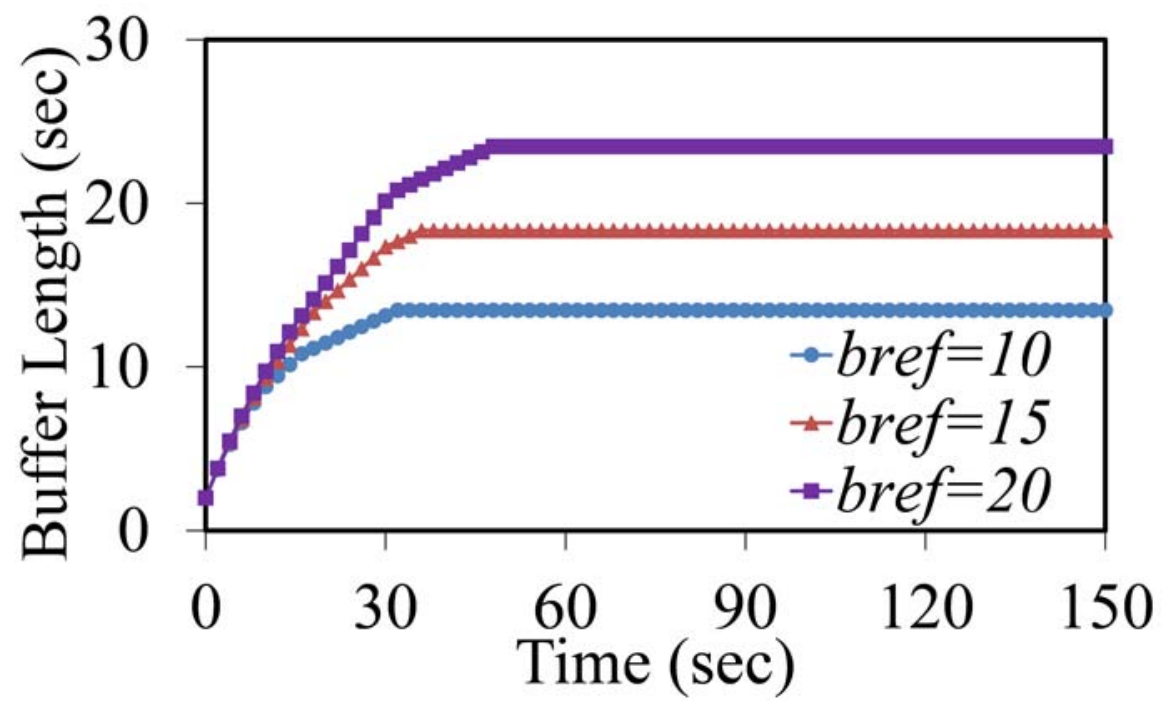}}
\caption{Results of case 1, in which two identical users compete the
server export bandwidth of 6Mbps, and request the \emph{BigBuckBunny}
video sequence with learning rate $\theta=100$. Here, $\mu=0.003$,
$\nu=0.0041$, $\alpha=2.15$, $\beta=0.0827$, $r^{\ast}=3$ Mbps, and
$B_{W}=6$ Mbps and the inequality relation of (29) is tenable. (a)
dynamic behavior of requested bitrates, (b) actual buffer lengths of
the two users with the reference buffer lengths of 10s, 15s, and 20s
respectively. \emph{\textbf{Note that the states of the two users are
identical}}.}
\label{fig6} 
\end{figure}

\begin{table}[htbp]
\setlength{\abovecaptionskip}{0.cm}
\setlength{\belowcaptionskip}{-0.cm}
  \centering
  \caption{Comparisons of Average Quality and Average Number of Switches of the Proposed Method with Different Learning Rates under Case 1}
    \begin{tabular}{ccccc}
    \toprule
    \textbf{Learning Rate $\theta$} & \textbf{50} & \textbf{100} & \textbf{150} & \textbf{200} \\
    \midrule
    \textbf{Average Bitrate (Mbps)} & 2.822  & 2.858  & 2.862  & 2.862  \\
    \midrule
    \textbf{Average PSNR (dB)} & 44.602  & 44.700  & 44.683  & 44.654  \\
    \midrule
    \textbf{Average SSIM} & 0.996  & 0.997  & 0.997  & 0.997  \\
    \midrule
    \textbf{Average Number of Switches} & 12    & 11    & 17    & 67  \\
    \bottomrule
    \end{tabular}%
  \label{table2}%
\end{table}%

\subsection{Results of Case 1}
In this case, two users compete the limited server export bandwidth
of size 6Mbps. The reference buffer length is set as 15s. Fig.
\ref{fig5} shows the requested bitrates and buffer lengths of the two
users by the proposed method with different learning rates. We can
observe that the Nash Equilibrium is achieved at $r^{\ast}=3$ Mbps,
and the actual buffer length converges to 15-20s except for the case
where the learning rate equals 50. When the learning rate increases
(e.g., $\theta=200$), the requested bitrate varies severely; however,
the reference buffer is achieved more accurately as shown in Fig.
\ref{fig5:subfig:b}. TABLE \ref{table2} compares the average
requested bitrate and bitrate switching frequency of the proposed
method with different learning rates ($\theta$). We can observe that
the average requested bitrates are similar, while the minimum bitrate
switching frequency is achieved when $\theta=100$.

Taking the learning rate of $\theta=100$ as an example, Fig.
\ref{fig6} shows that the Nash Equilibrium is achieved in a slower
pace with the reference buffer increasing. Specifically, the
converged buffer lengths are 13.32s, 17.93s, and 22.68s when the
reference buffer lengths are set as 10s, 15s, and 20s, respectively.
The reason is that the buffer needs more time to accumulate.

\subsection{Results of Case 2}
\begin{figure}
\setlength{\abovecaptionskip}{0.cm}
\setlength{\belowcaptionskip}{-0.cm} \centering \subfigure[]{
\label{fig7:subfig:a} 
\includegraphics[width=4.25cm]{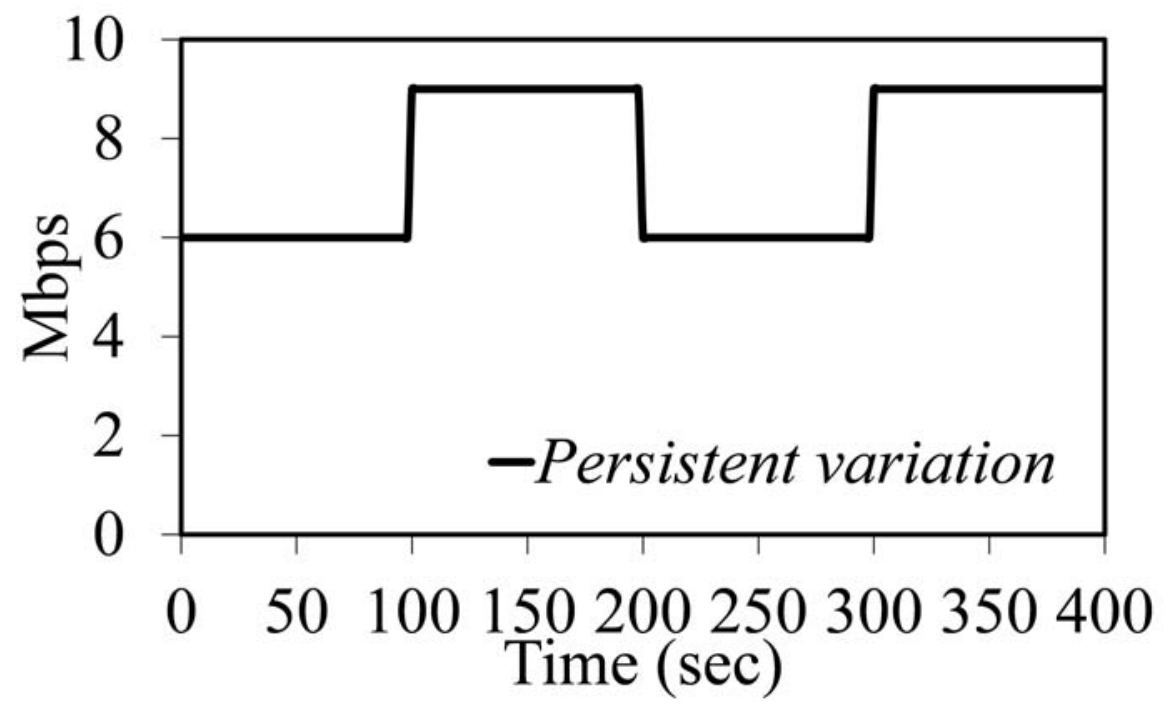}}
\subfigure[]{
\label{fig7:subfig:b} 
\includegraphics[width=4.25cm]{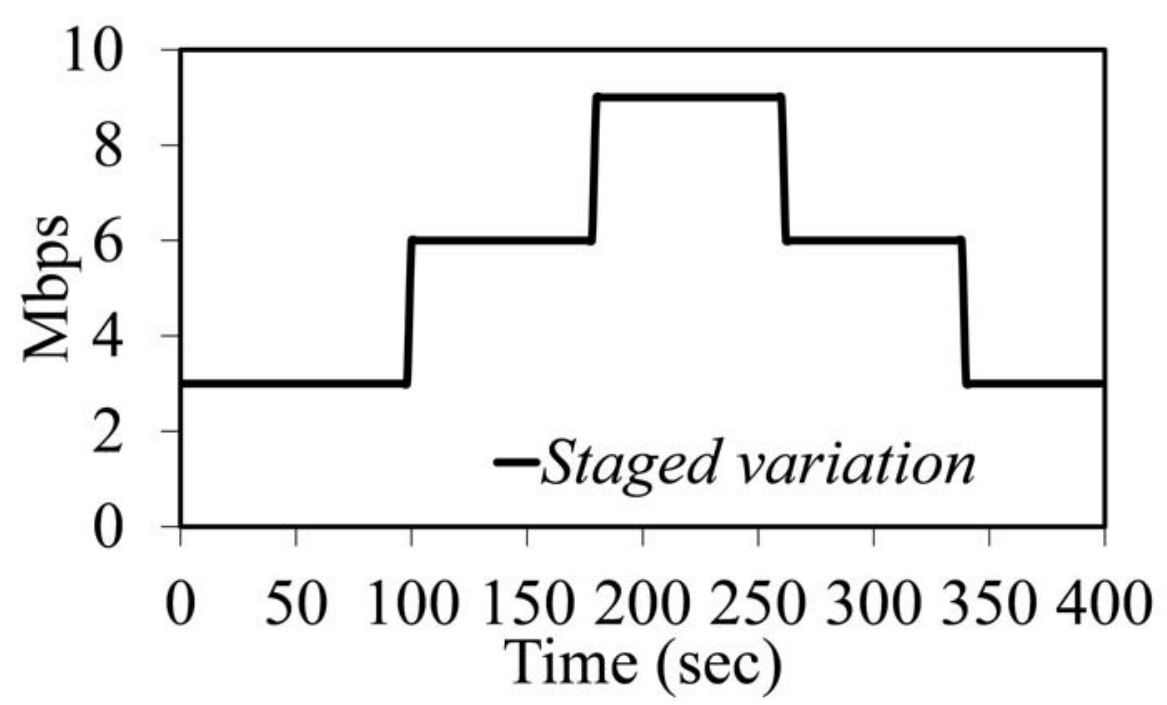}}
\subfigure[]{
\label{fig7:subfig:c} 
\includegraphics[width=4.25cm]{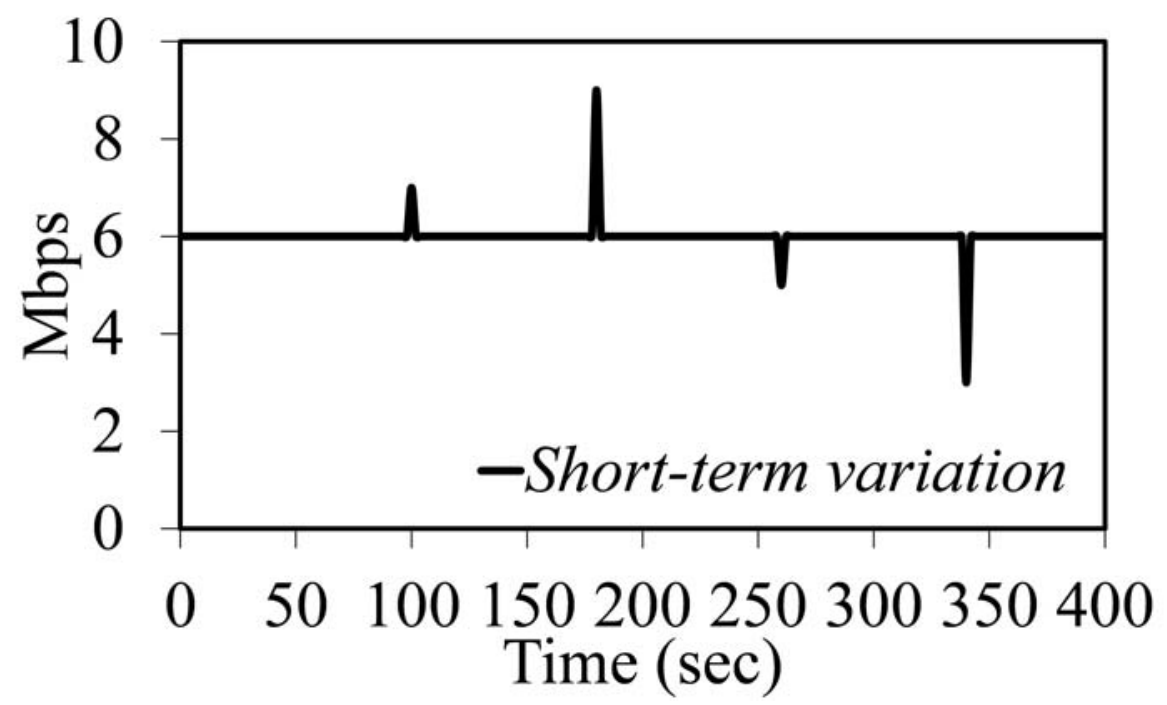}}
\caption{Three kinds of server bandwidth variations, (a) persistent
variation, (b) staged variation, (c) short-term variation.}
\label{fig7} 
\end{figure}

We verify the proposed method with a varied server export bandwidth
that is realized via three types of variations \cite{R49}, i.e.,
persistent variation, staged variation, and short-term variation, as
shown in Fig. \ref{fig7}. The persistent and staged bandwidth
variations (both increment and decrement) that last for tens of
seconds appear frequently in practice when the cross traffic in the
path's bottleneck varies significantly due to arriving or departing
traffic of some users. A good rate adaptation method should adapt to
such variations by decreasing or increasing the requested bitrate.
The short-term variation that lasts for only a few seconds is usually
caused by burst change of channel states. For such short-term
variations, the user should be able to keep requested bitrate stable
to avoid unnecessary bitrate variations.

\begin{figure}
\setlength{\abovecaptionskip}{0.cm}
\setlength{\belowcaptionskip}{-0.cm} \centering \subfigure[]{
\label{fig8:subfig:a} 
\includegraphics[width=4.25cm]{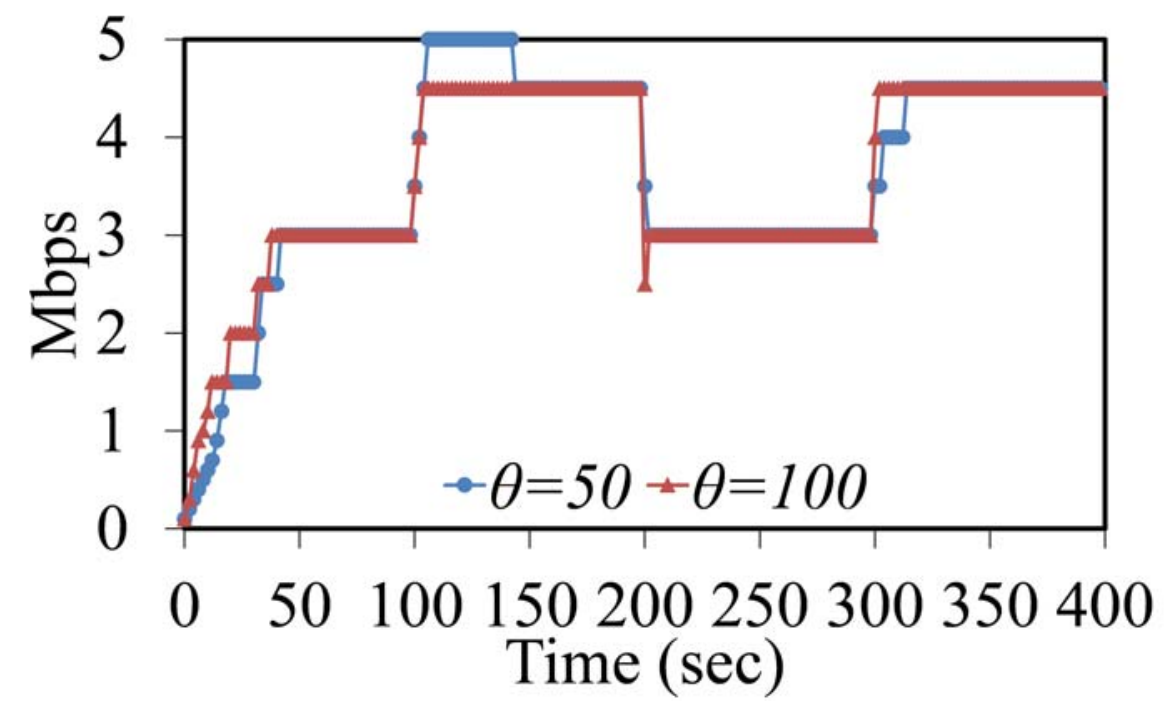}}
\subfigure[]{
\label{fig8:subfig:b} 
\includegraphics[width=4.25cm]{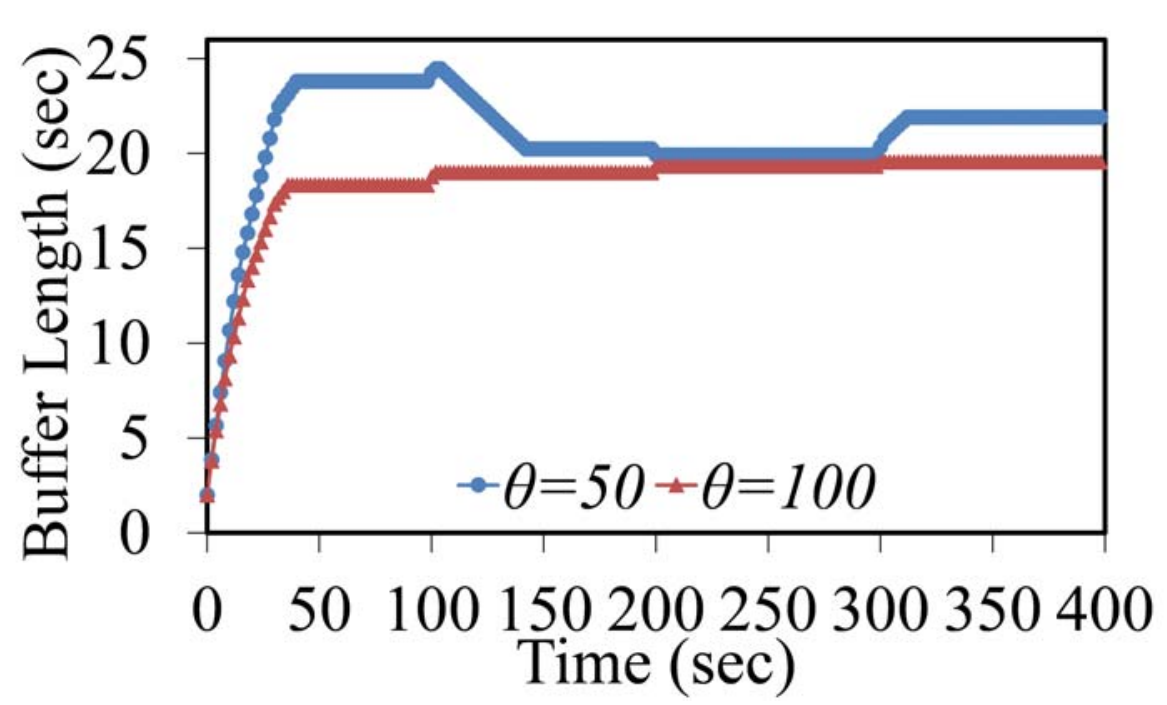}}
\caption{Results of case 2, in which two identical users compete the
server export bandwidth with \textit{\textbf{persistent variation}},
and request the \emph{BigBuckBunny} video sequence with learning rate
$\theta=50$ and 100 respectively. Here, $\mu=0.003$, $\nu=0.0041$.
(a) dynamic behavior of requested bitrates, (b) actual buffer lengths
of the two users with the reference buffer length of 15s. The initial
server export bandwidth is 6Mbps. \emph{\textbf{Note that the states
of the two users are identical}}.}
\label{fig8} 
\end{figure}

Fig. \ref{fig8} shows the requested bitrates and buffer lengths of
two users by the proposed method when the server export bandwidth
exhibits persistent variation. We can observe that the requested
bitrate increases to the Nash Equilibrium rapidly, and the reference
buffer is also reached before $t=50$s. When the server export
bandwidth varies to 9 Mbps at the time of 100s, the requested bitrate
with $\theta=100$ quickly increases to 4.5 Mbps, while the requested
bitrate with $\theta=50$ firstly increases to 5 Mbps, which consumes
about 5s playout buffer before dropping to 4.5 Mbps. When the
available bandwidth decreases back to 6 Mbps at the time of 200s, the
requested bitrate with $\theta=100$ drops to 3 Mbps directly, while
the requested bitrate with $\theta=50$ firstly drops to 3.5 Mbps.
When the available bandwidth increases back to 9 Mbps at the time of
300s, the requested bitrate with $\theta=50$ steps to 4.5 Mbps, while
the requested bitrate with $\theta=100$ increases to 4.5 Mbps
directly. We can conclude that a larger learning rate can follow the
bandwidth variation accurately and keep the buffer lengths more
stable, while the requested bitrate is more stable for a smaller
learning rate.

\begin{figure}
\setlength{\abovecaptionskip}{0.cm}
\setlength{\belowcaptionskip}{-0.cm} \centering \subfigure[]{
\label{fig9:subfig:a} 
\includegraphics[width=4.25cm]{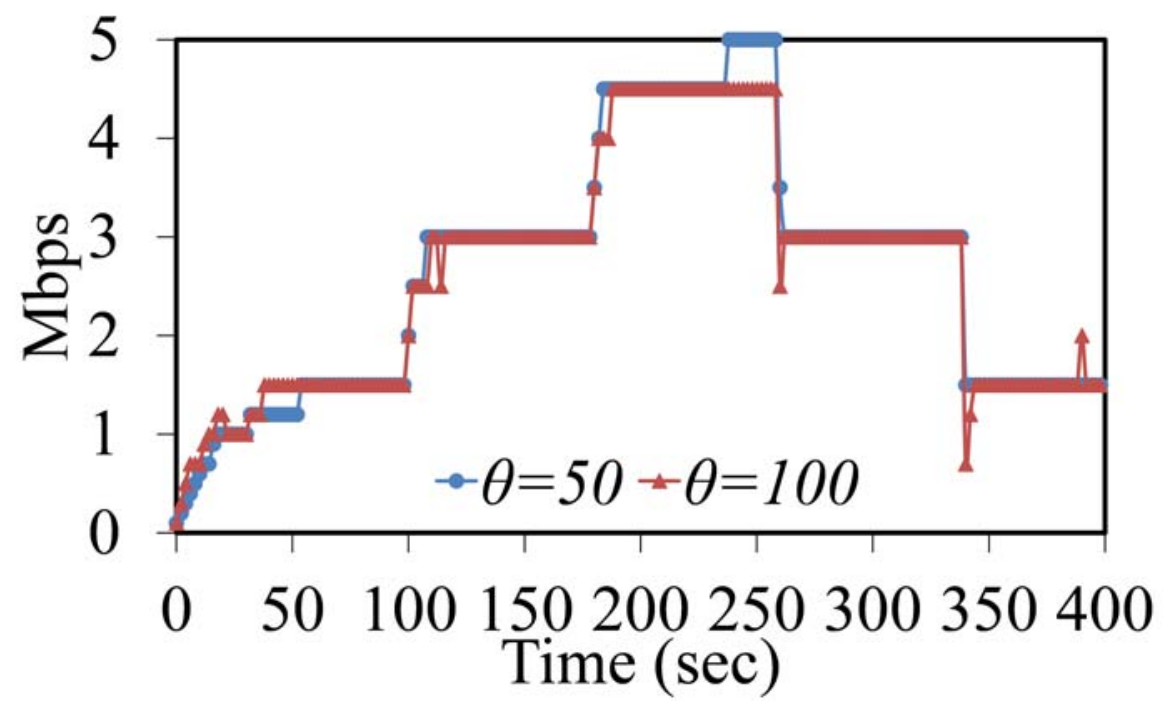}}
\subfigure[]{
\label{fig9:subfig:b} 
\includegraphics[width=4.25cm]{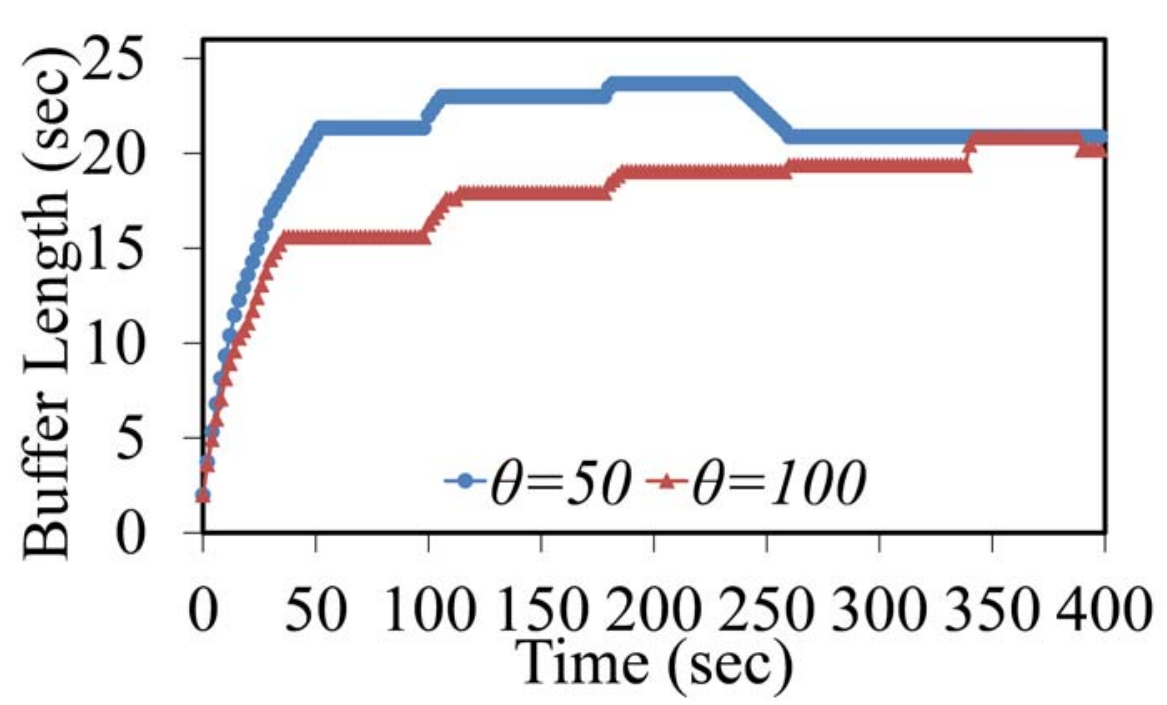}}
\caption{Results of case 2, in which two identical users compete the
server export bandwidth with \textit{\textbf{staged variation}}, and
request the \emph{BigBuckBunny} video sequence with learning rate
$\theta=50$ and 100 respectively. Here, $\mu=0.003$, $\nu=0.0041$.
(a) dynamic behavior of requested bitrates, (b) actual buffer lengths
of the two users with the reference buffer length of 15s. The initial
server export bandwidth is 6Mbps. \emph{\textbf{Note that the states
of the two users are identical}}.}
\label{fig9} 
\end{figure}

Fig. \ref{fig9} shows the requested bitrates and  buffer lengths of
two users by the proposed method when the server export bandwidth
exhibits staged variation. Similarly, we can also observe that the
requested bitrate with a smaller learning rate (e.g., $\theta=50$) is
more stable and the bitrate fluctuations are smaller at the time of
bandwidth changing (i.e., at 100s, 180s, 260s, and 340s). However,
the buffer length is further away from the reference buffer length
than the learning rate of $\theta=100$. Nevertheless, the Nash
Equilibrium can be achieved under both cases.

\begin{figure}
\setlength{\abovecaptionskip}{0.cm}
\setlength{\belowcaptionskip}{-0.cm} \centering \subfigure[]{
\label{fig10:subfig:a} 
\includegraphics[width=4.25cm]{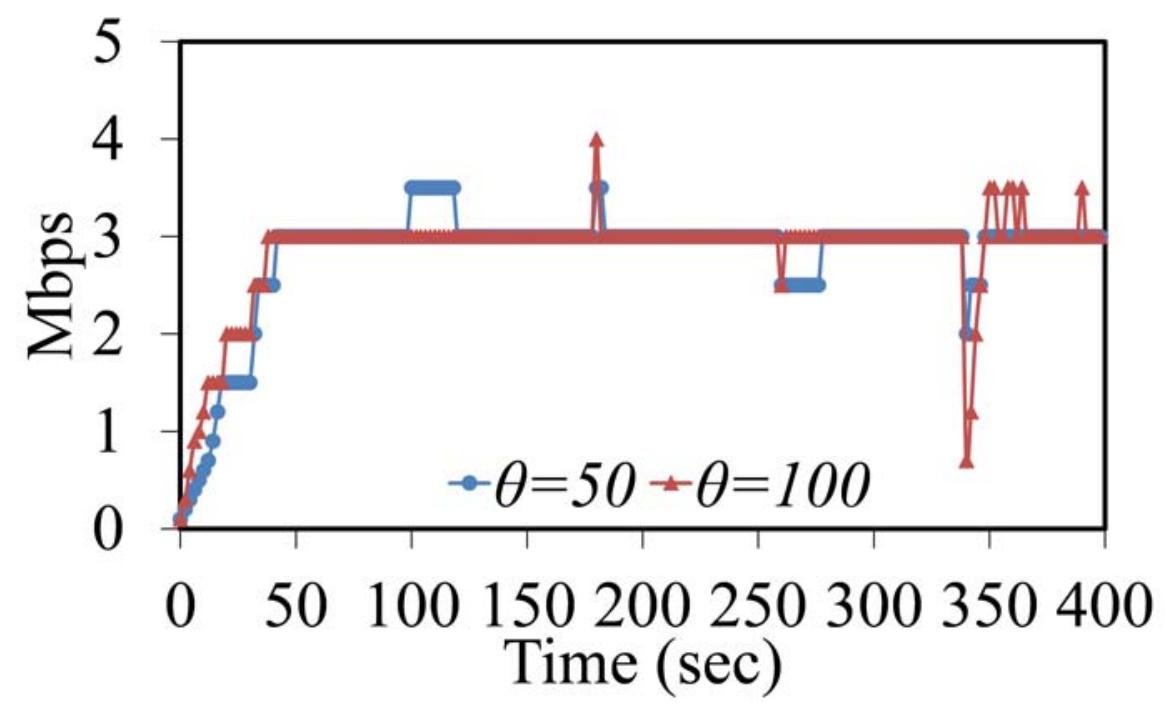}}
\subfigure[]{
\label{fig10:subfig:b} 
\includegraphics[width=4.25cm]{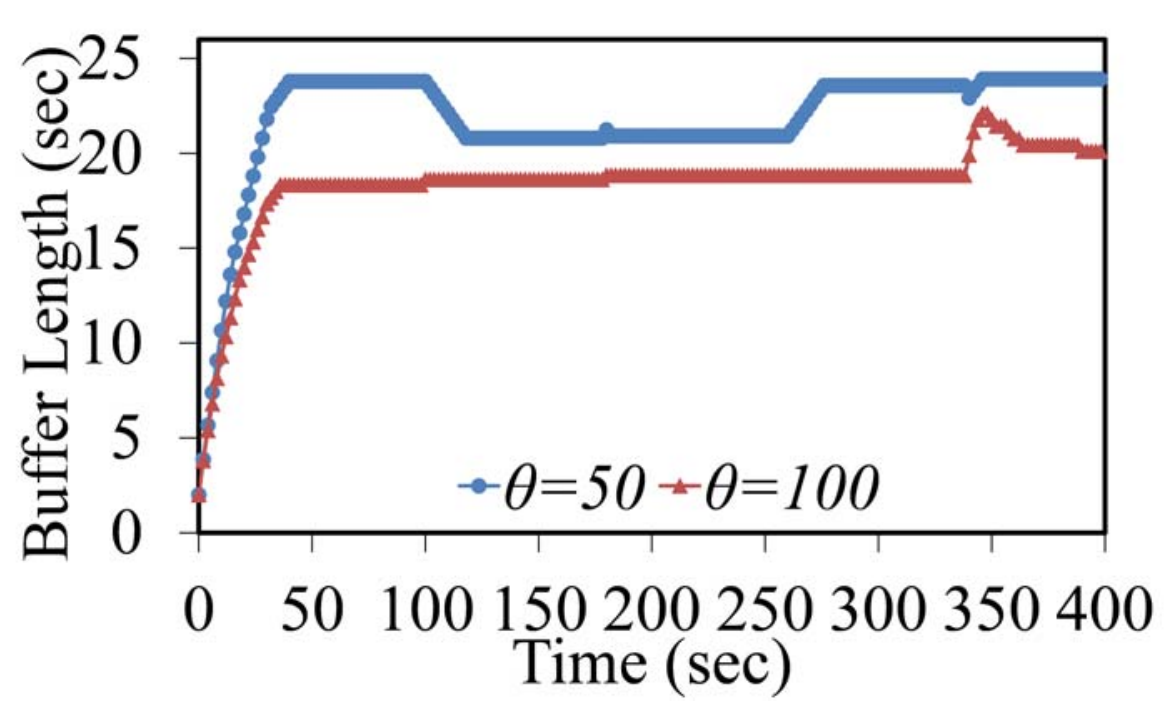}}
\caption{Results of case 2, in which two identical users compete the
server export bandwidth with \textit{\textbf{short-term variation}},
and request the \emph{BigBuckBunny} video sequence with learning rate
$\theta=50$ and 100 respectively. Here, $\mu=0.003$, $\nu=0.0041$.
(a) dynamic behavior of requested bitrates, (b) actual buffer lengths
of the two users with the reference buffer length of 15s. The initial
server export bandwidth is 6Mbps. \emph{\textbf{Note that the states
of the two users are identical}}.}
\label{fig10} 
\end{figure}

For the case of short-term server export bandwidth variation, the
requested bitrate with a smaller learning rate ($\theta=50$) is more
stable, as shown in Fig. \ref{fig10}. The requested bitrate with a
smaller learning rate can avoid abrupt short-term changing by
accumulating/consuming buffered video segments when encountering a
small amplitude bandwidth variation, e.g., at the time of 100s and
260s, as shown in Fig. \ref{fig10:subfig:a}. But, the difference
between the actual buffer length and the reference buffer length of
$\theta=50$ is larger than that of the larger learning rate, i.e.,
$\theta=100$, as shown in Fig. \ref{fig10:subfig:b}.

\subsection{Results of Case 3}
\begin{figure}
\setlength{\abovecaptionskip}{0.cm}
\setlength{\belowcaptionskip}{-0.cm} \centering \subfigure[]{
\label{fig11:subfig:a} 
\includegraphics[width=4.25cm]{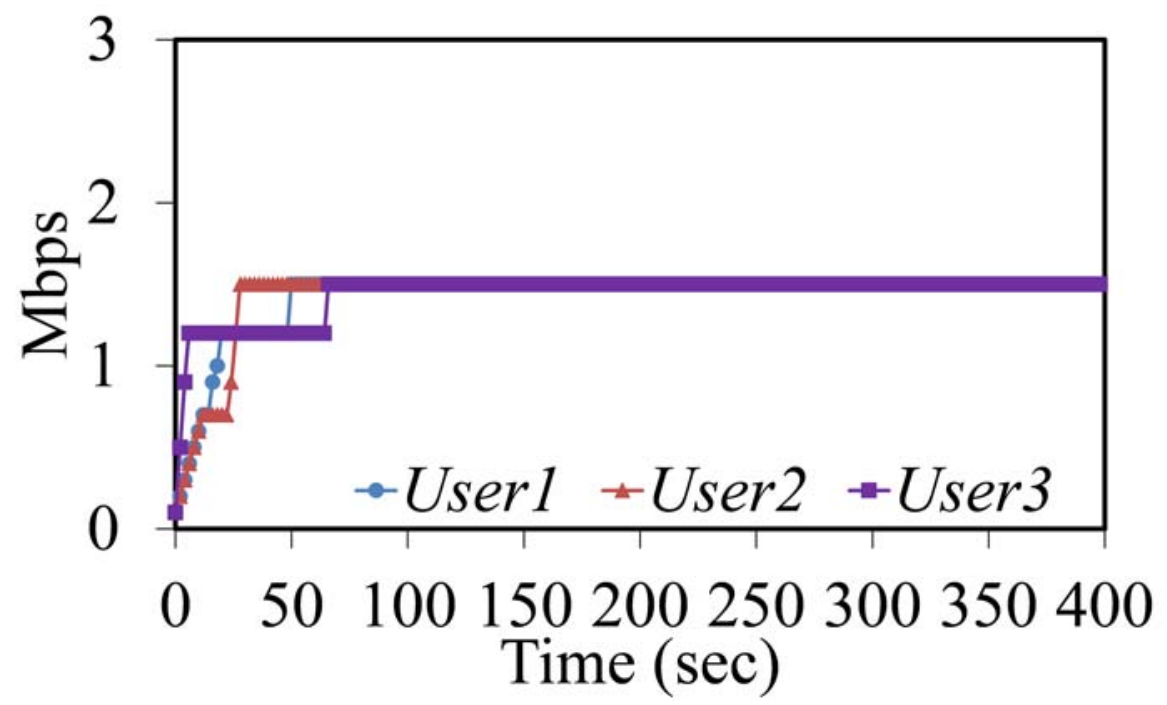}}
\subfigure[]{
\label{fig11:subfig:b} 
\includegraphics[width=4.25cm]{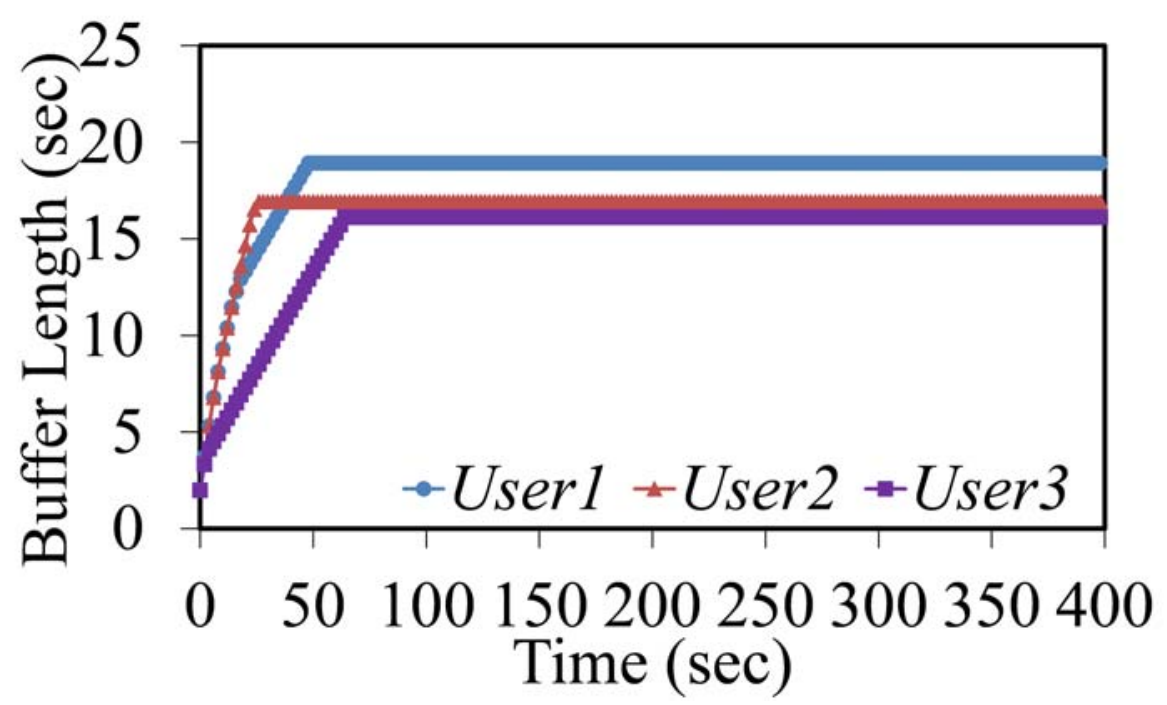}}
\subfigure[]{
\label{fig11:subfig:c} 
\includegraphics[width=4.25cm]{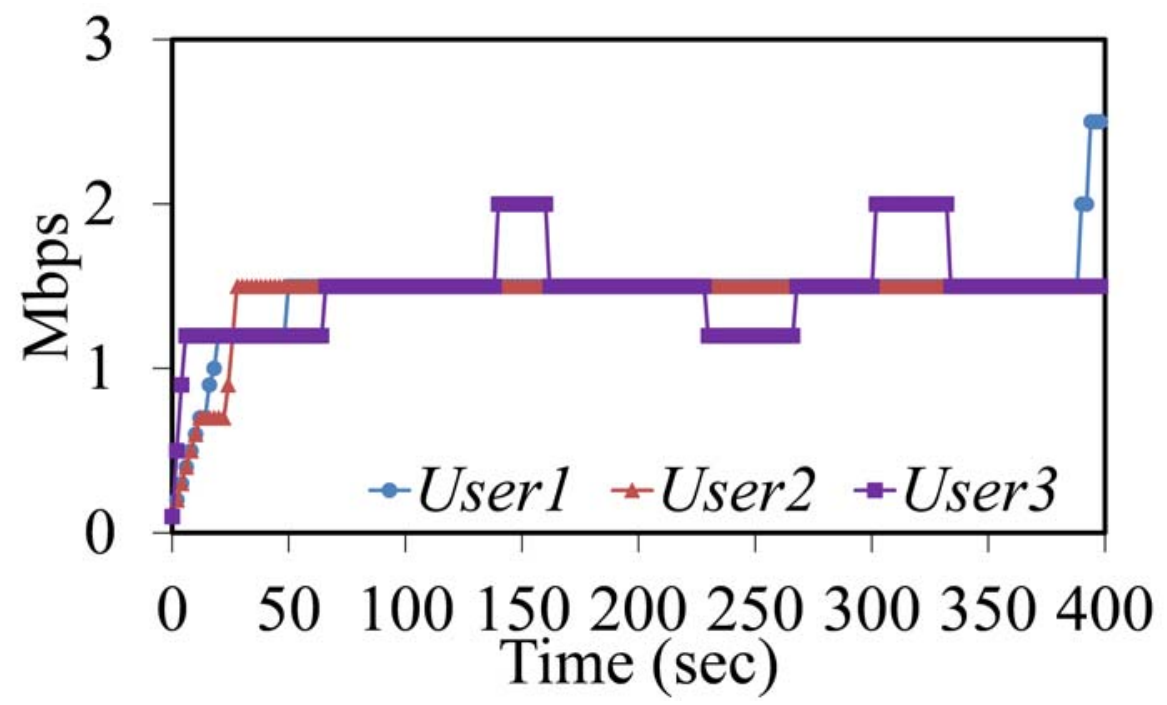}}
\subfigure[]{
\label{fig11:subfig:d} 
\includegraphics[width=4.25cm]{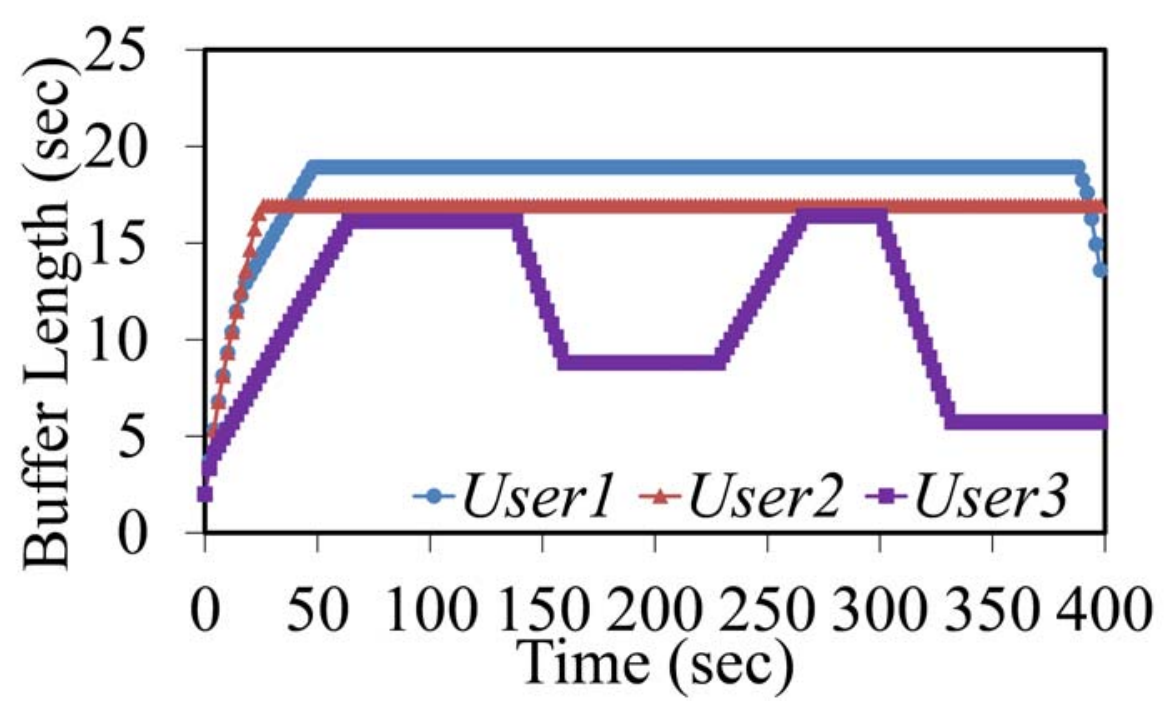}}
\subfigure[]{
\label{fig11:subfig:e} 
\includegraphics[width=4.25cm]{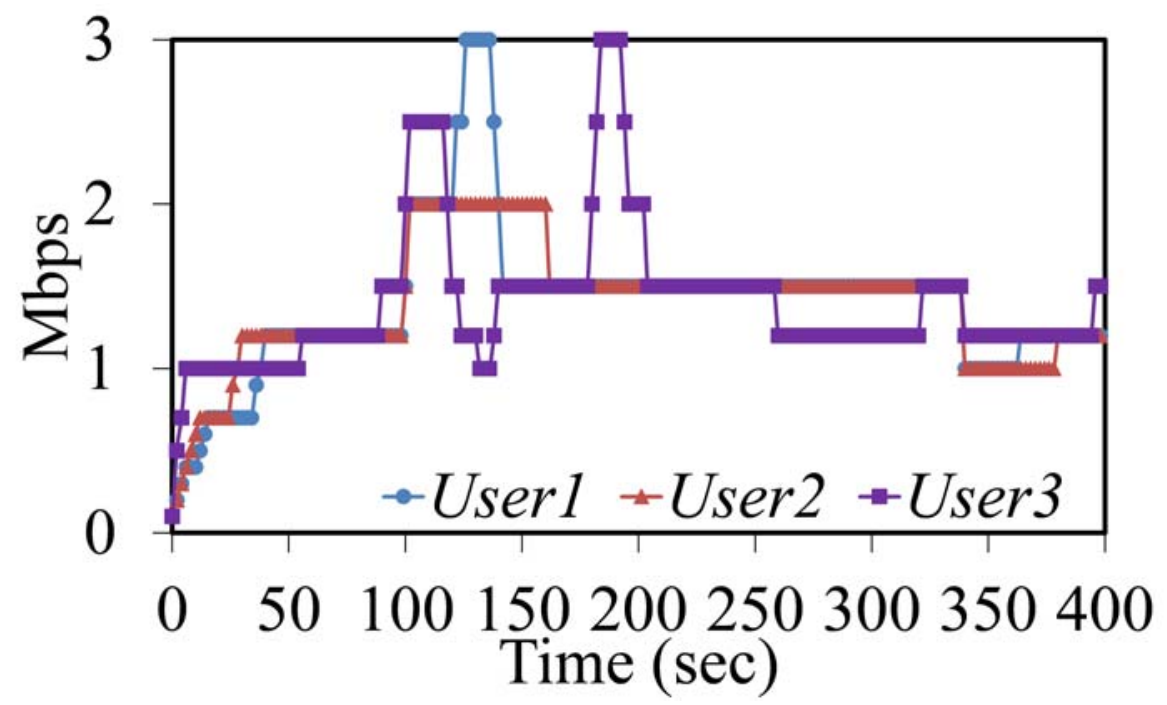}}
\subfigure[]{
\label{fig11:subfig:f} 
\includegraphics[width=4.25cm]{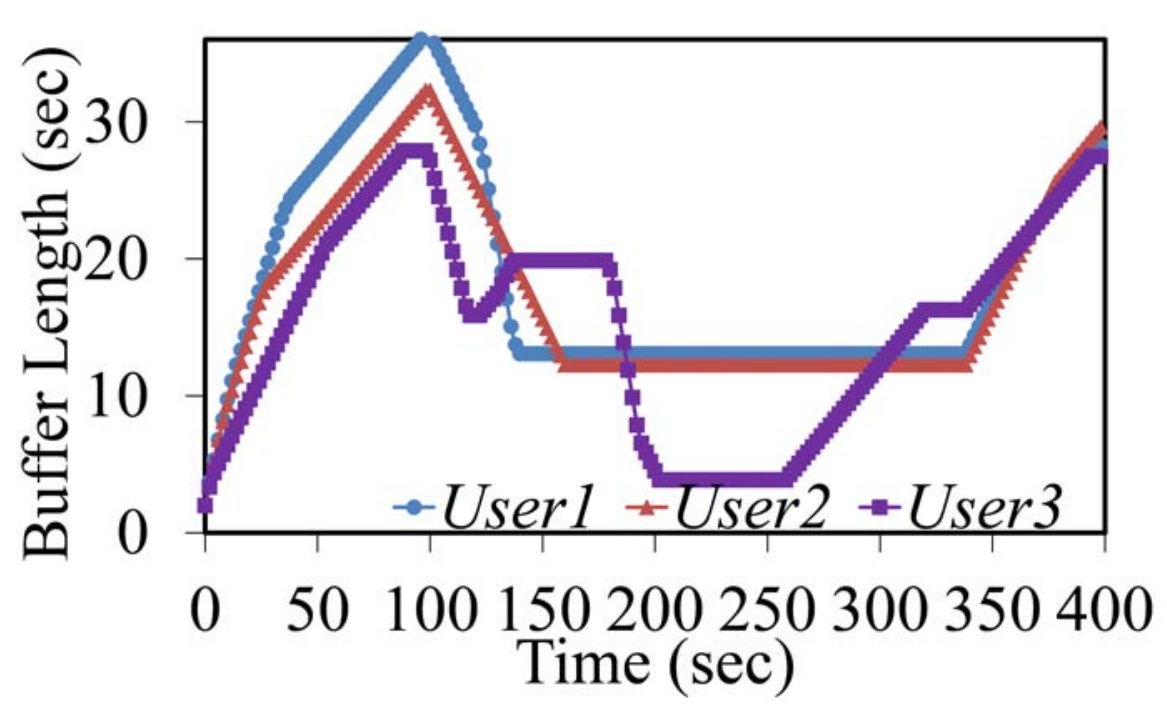}}
\subfigure[]{
\label{fig11:subfig:g} 
\includegraphics[width=4.25cm]{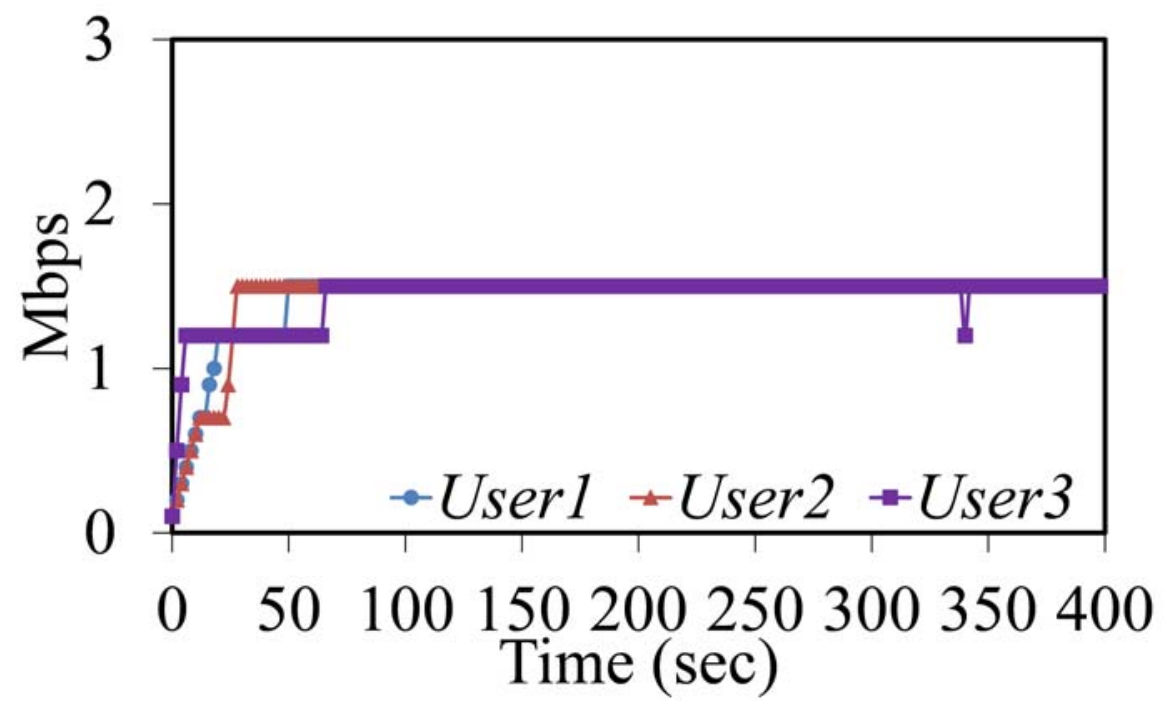}}
\subfigure[]{
\label{fig11:subfig:h} 
\includegraphics[width=4.25cm]{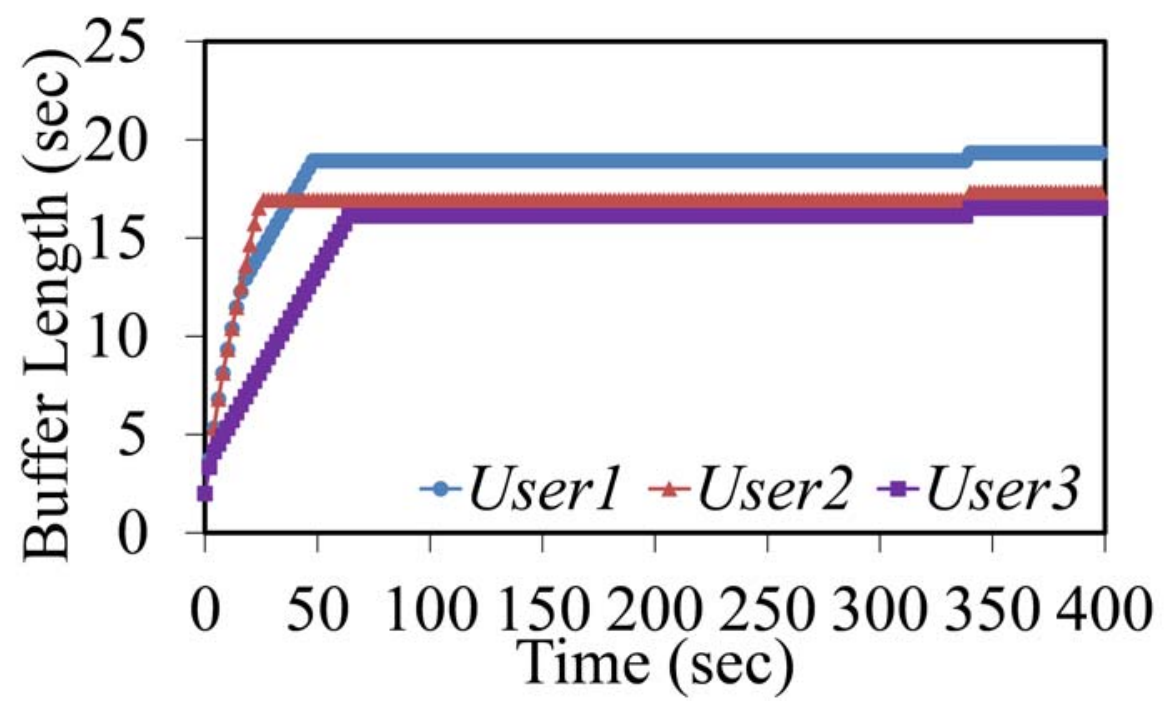}}
\caption{Results of case 3, in which three users (with
\textit{\textbf{fixed and limited}} throughputs of 1.5 Mbps) compete
the server export bandwidth (6 Mbps), and request
\emph{BigBuckBunny}, \textit{ElephantsDream}, and
\textit{SitaSingstheBlues}, respectively, with learning rate
$\theta=50$. Here, $\mu=0.003$, $\nu=0.0041$. (a), (c), (e), and (g)
show the requested bitrates for fixed, persistent variation, staged
variation, and short-term variation of server export bandwidth. (b),
(d), (f), and (h) show the corresponding buffer length of each user.}
\label{fig11} 
\end{figure}

This subsection presents the performance of the proposed algorithm
under \emph{\textbf{Case 3}}, in which three users request three
different videos (i.e., \emph{BigBuckBunny} for \emph{User1},
\emph{ElephantsDream} for \emph{User2}, and \emph{SitaSingstheBlues}
for \emph{User3}) with a varied server export bandwidth. Note that
the channel throughput of each user is limited up to 1.5 Mbps, which
is unknown to the users, and the server export bandwidth is set as 6
Mbps. The reference buffer length is set as 15s. From Fig.
\ref{fig11}, we can observe that the Nash Equilibrium is achieved
with $r_{1}^{*}=1.5$ Mbps, $r_{2}^{*}=1.5$ Mbps, and $r_{3}^{*}=1.5$
Mbps, for different server export bandwidth variation scenarios, and
no playback interruption occurs.

\subsection{Results of Case 4}
In addition, we also compare the proposed method (denoted by
\textit{\textbf{Proposed}}) with other methods (i.e.,
\textit{\textbf{BF}}, \textit{\textbf{QF}}, and
\textit{\textbf{QBA}}) with \textit{\textbf{random}} user channel
throughput limitations that are also unknown for each user.

\begin{figure}
\setlength{\abovecaptionskip}{0.cm}
\setlength{\belowcaptionskip}{-0.cm} \centering \subfigure[]{
\label{fig12:subfig:a} 
\includegraphics[width=4.25cm]{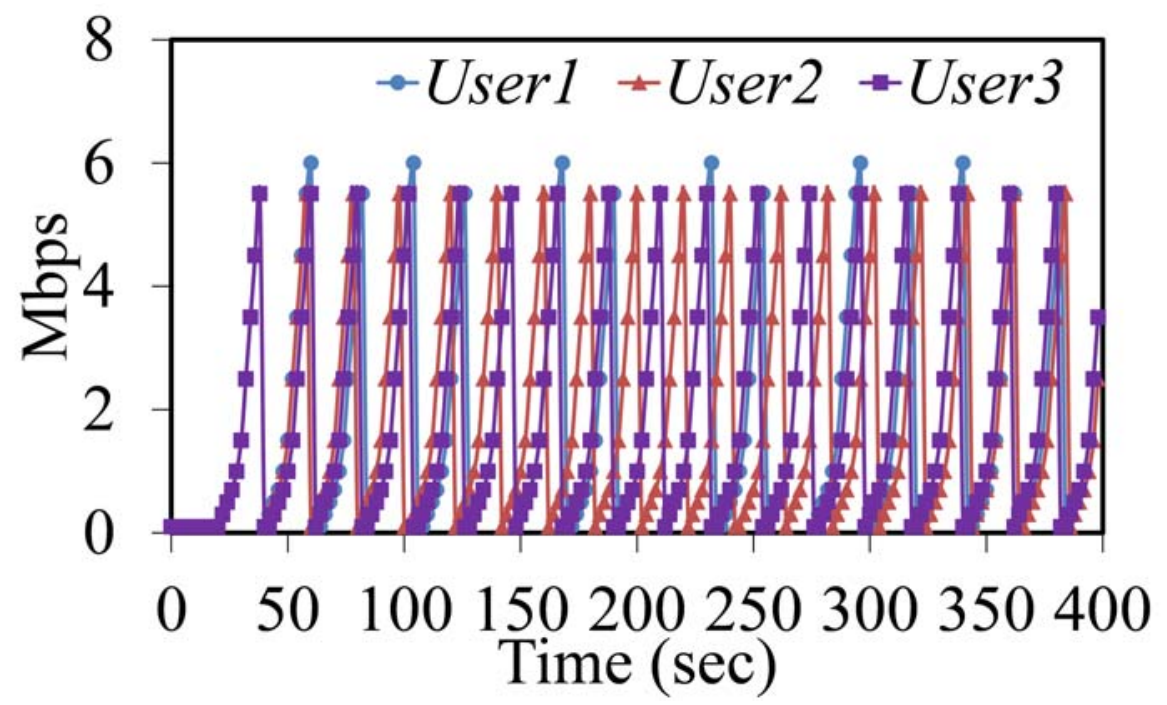}}
\subfigure[]{
\label{fig12:subfig:b} 
\includegraphics[width=4.25cm]{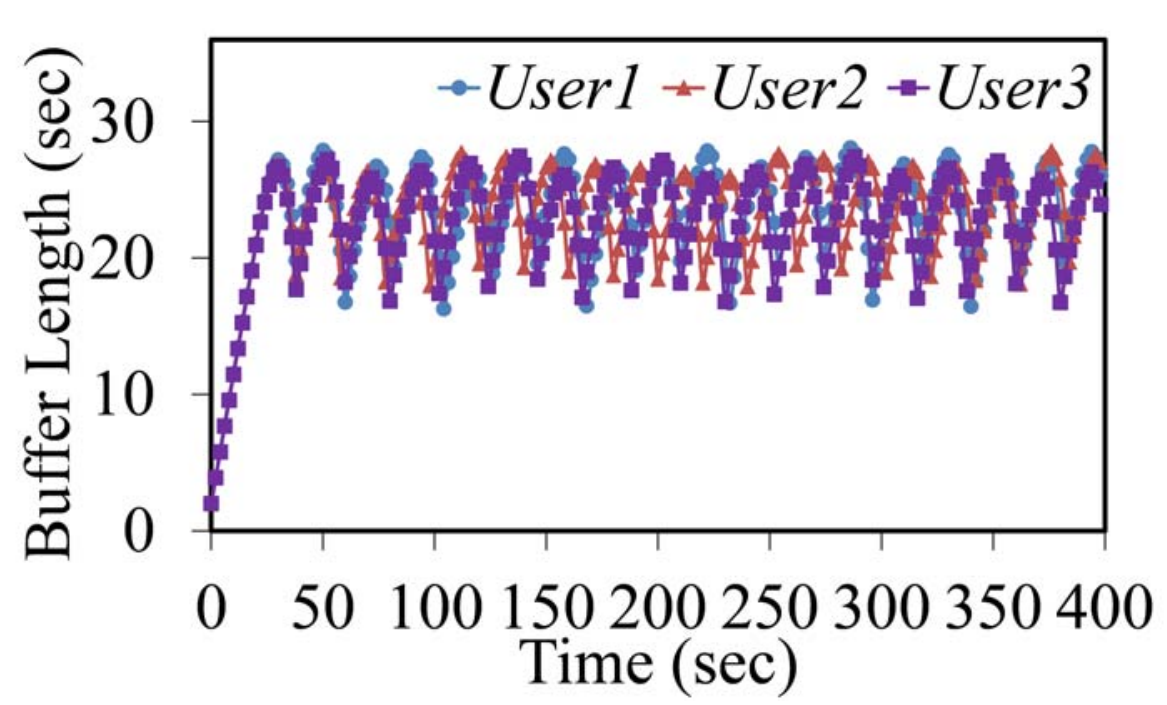}}
\subfigure[]{
\label{fig12:subfig:c} 
\includegraphics[width=4.25cm]{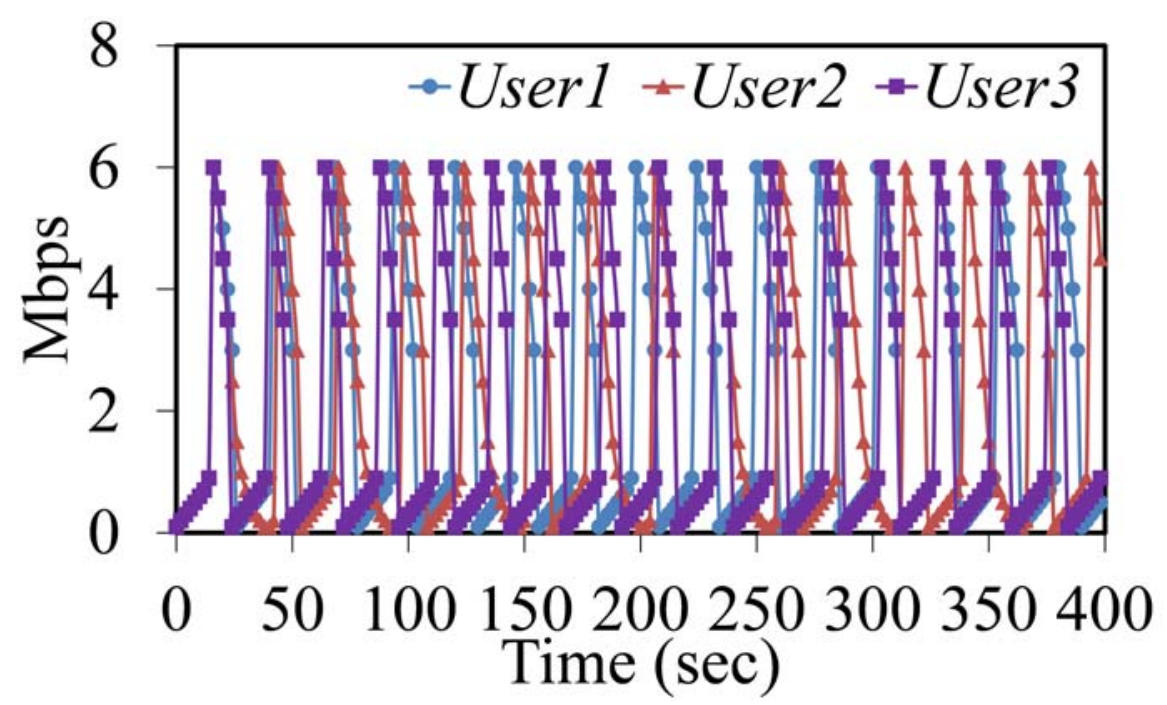}}
\subfigure[]{
\label{fig12:subfig:d} 
\includegraphics[width=4.25cm]{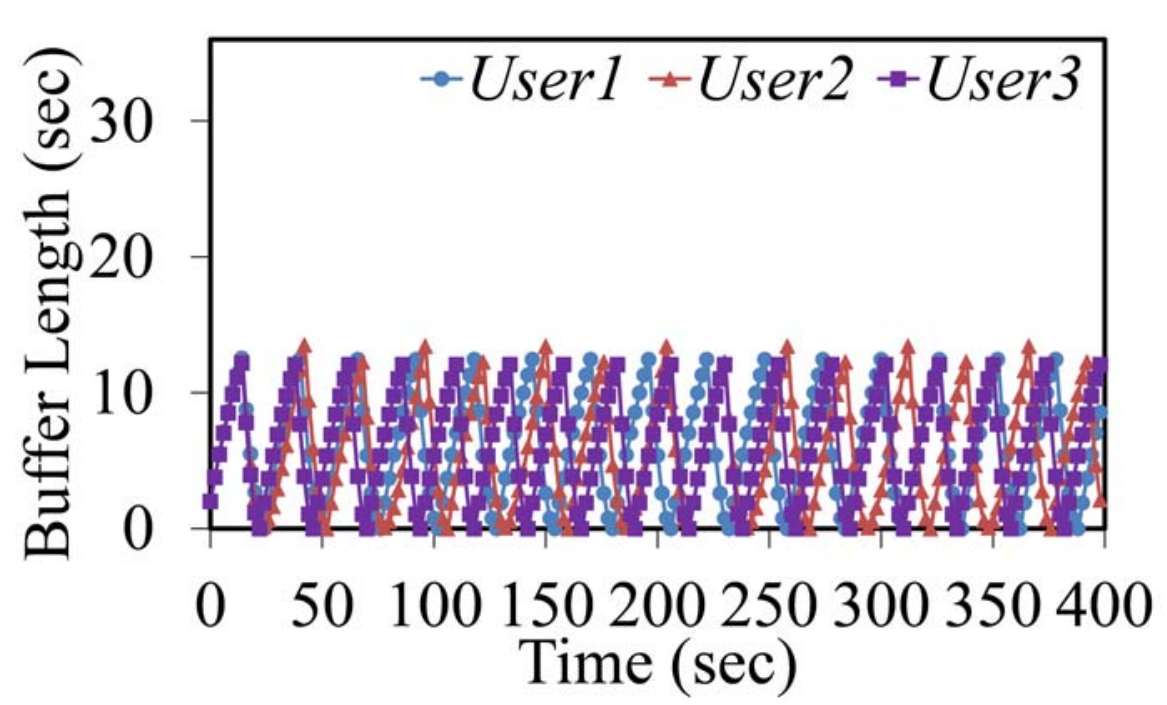}}
\subfigure[]{
\label{fig12:subfig:e} 
\includegraphics[width=4.25cm]{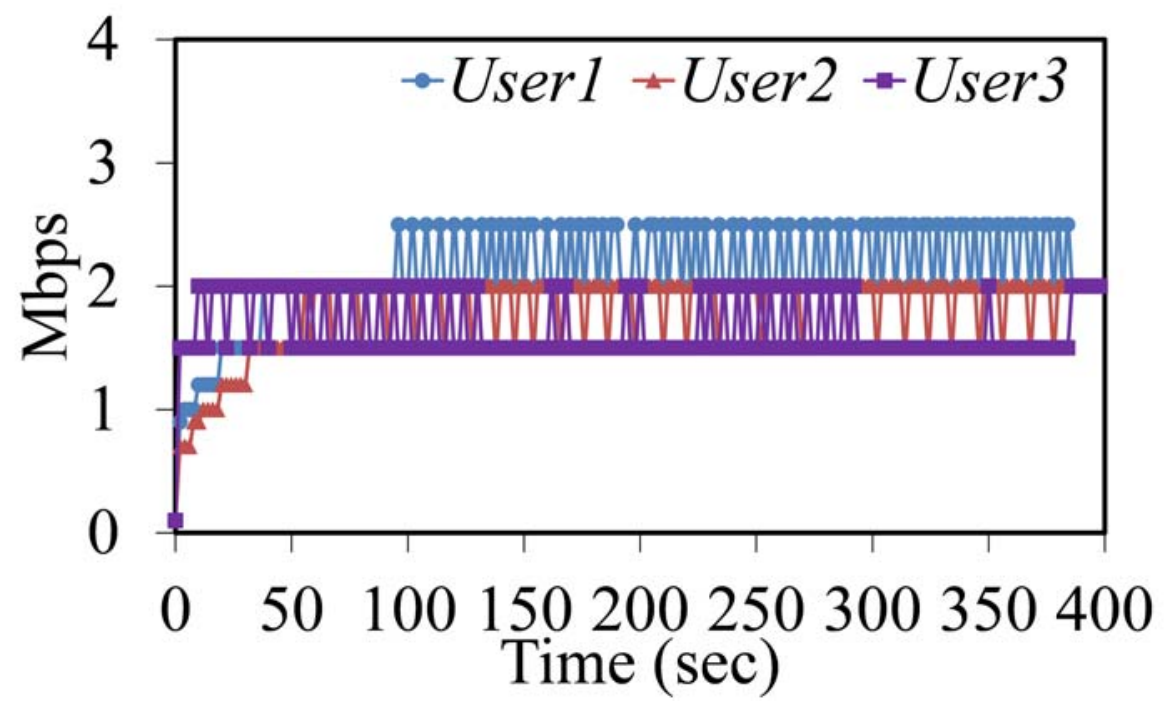}}
\subfigure[]{
\label{fig12:subfig:f} 
\includegraphics[width=4.25cm]{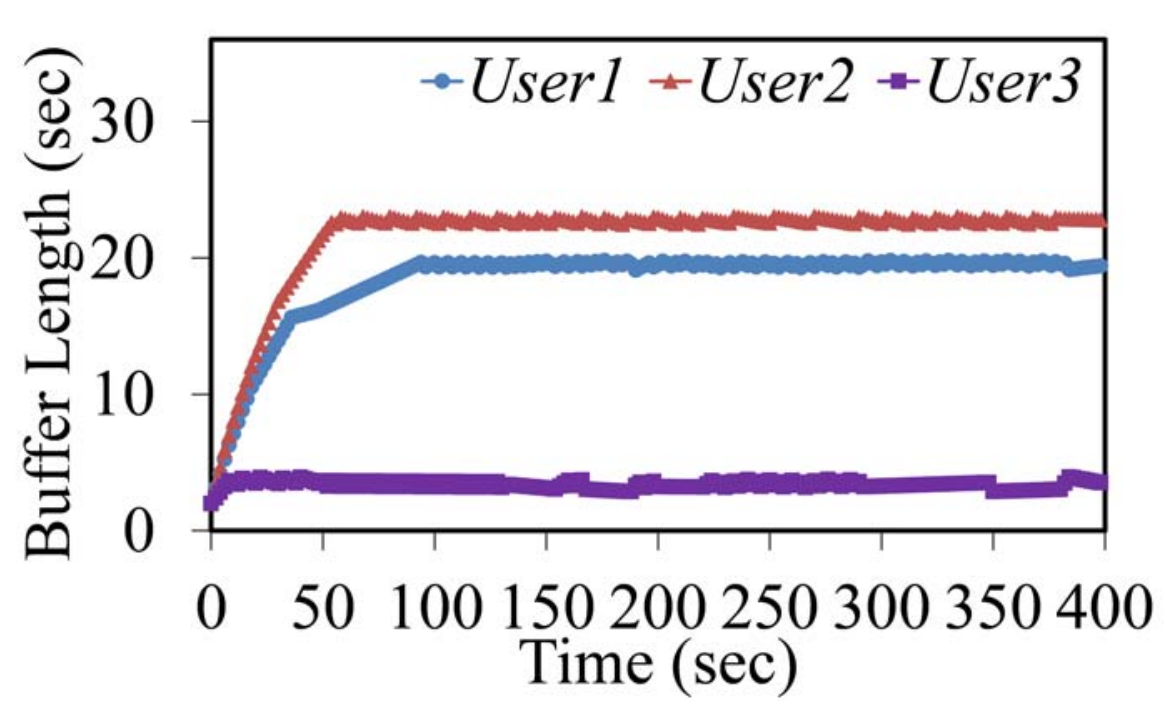}}
\subfigure[]{
\label{fig12:subfig:g} 
\includegraphics[width=4.25cm]{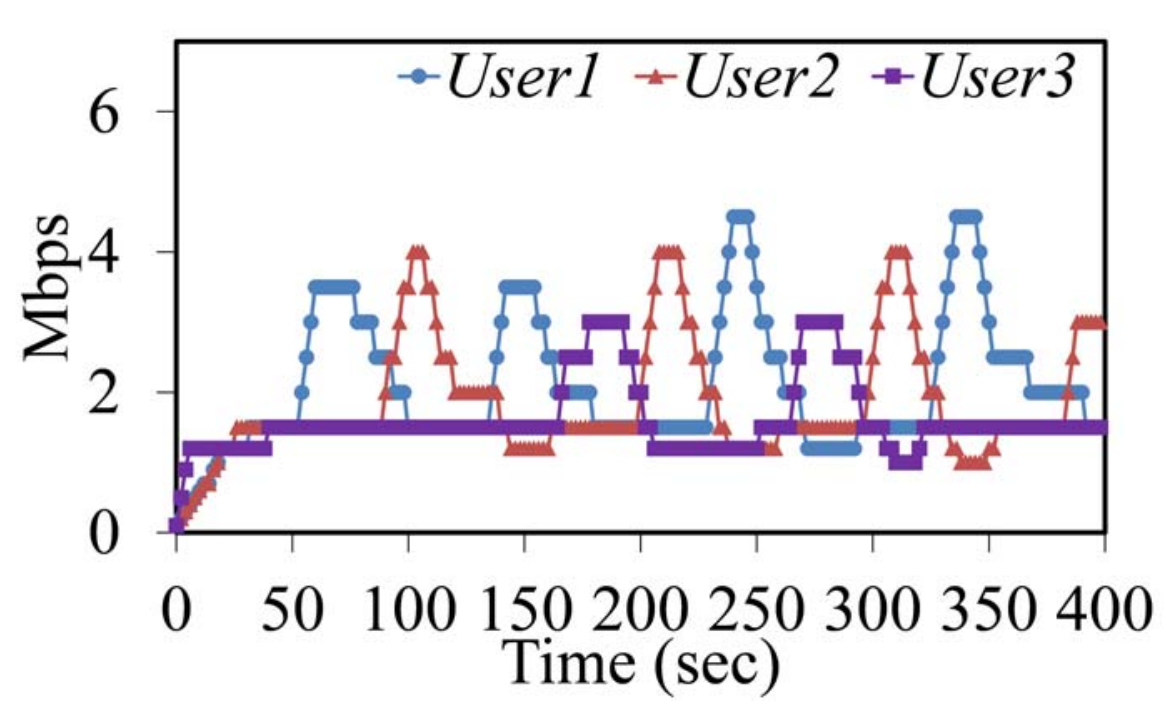}}
\subfigure[]{
\label{fig12:subfig:h} 
\includegraphics[width=4.25cm]{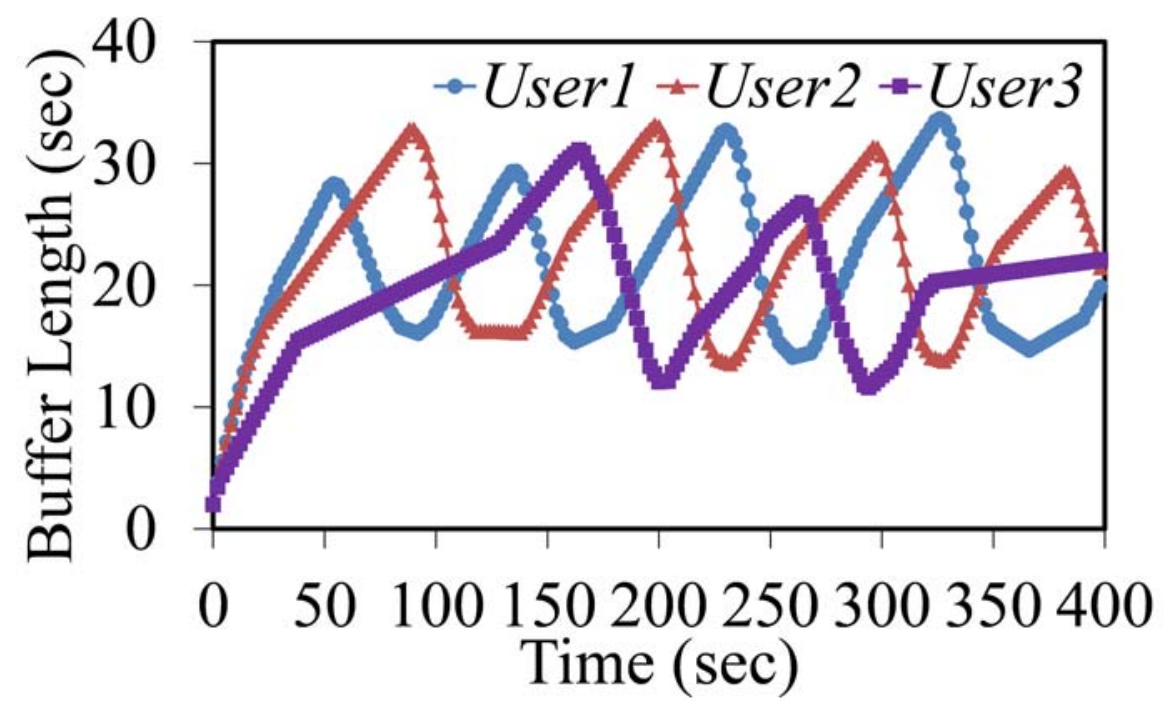}}
\caption{Results of case 4, in which three users (with
\textit{\textbf{random and limited}} throughputs) compete the server
export bandwidth (\textbf{\emph{fixed}} 6 Mbps), and request
\emph{BigBuckBunny}, \textit{ElephantsDream}, and
\textit{SitaSingstheBlues}, respectively, with learning rate
$\theta=50$. Here, $\mu=0.003$, $\nu=0.0041$. (a), (c), (e), and (g)
show the requested bitrates of \textit{\textbf{BF}},
\textit{\textbf{QF}}, \textit{\textbf{QBA}}, and the
\textit{\textbf{Proposed}} methods; while (b), (d), (f), and (h) show
the corresponding buffer length of each user.}
\label{fig12} 
\end{figure}

Fig. \ref{fig12} compares the requested bitrate and the buffer length
of each user when the server export bandwidth is fixed to 6 Mbps. We
can observe that the \textit{\textbf{BF}} method switches the video
bitrate frequently (see Fig. \ref{fig12:subfig:a}) in order to ensure
the actual buffer length is close to the reference buffer length (see
Fig. \ref{fig12:subfig:b}). As shown in Figs. \ref{fig12:subfig:c}
and (d), the \textit{\textbf{QF}} method first accumulates a certain
buffer length, and then struggles to request the highest bitrates,
resulting in frequent re-buffering and bitrate switching. For the
\textit{\textbf{QBA}} method, its requested bitrates are more stable
than those of \textit{\textbf{BF}} and \textit{\textbf{QF}} methods,
but the buffer lengths of the three users are not fair, i.e., the
buffer length of \emph{User3} is much smaller than that of
\emph{User1} and \emph{User2}, as shown in Figs. \ref{fig12:subfig:e}
and (f). The reason is that the \textit{\textbf{QBA}} method does not
take the fairness of users into consideration too much. From Figs.
\ref{fig12:subfig:g} and (h), it can be observed that the requested
video bitrates by the \textit{\textbf{Proposed}} method are stable
for all of the three users, and their buffer lengths of the three
users vary around the reference buffer length (set as 15s). Besides,
there is no re-buffering (playback interruption) for the
\textit{\textbf{Proposed}} method.

\begin{figure}
\setlength{\abovecaptionskip}{0.cm}
\setlength{\belowcaptionskip}{-0.cm} \centering \subfigure[]{
\label{fig13:subfig:a} 
\includegraphics[width=4.25cm]{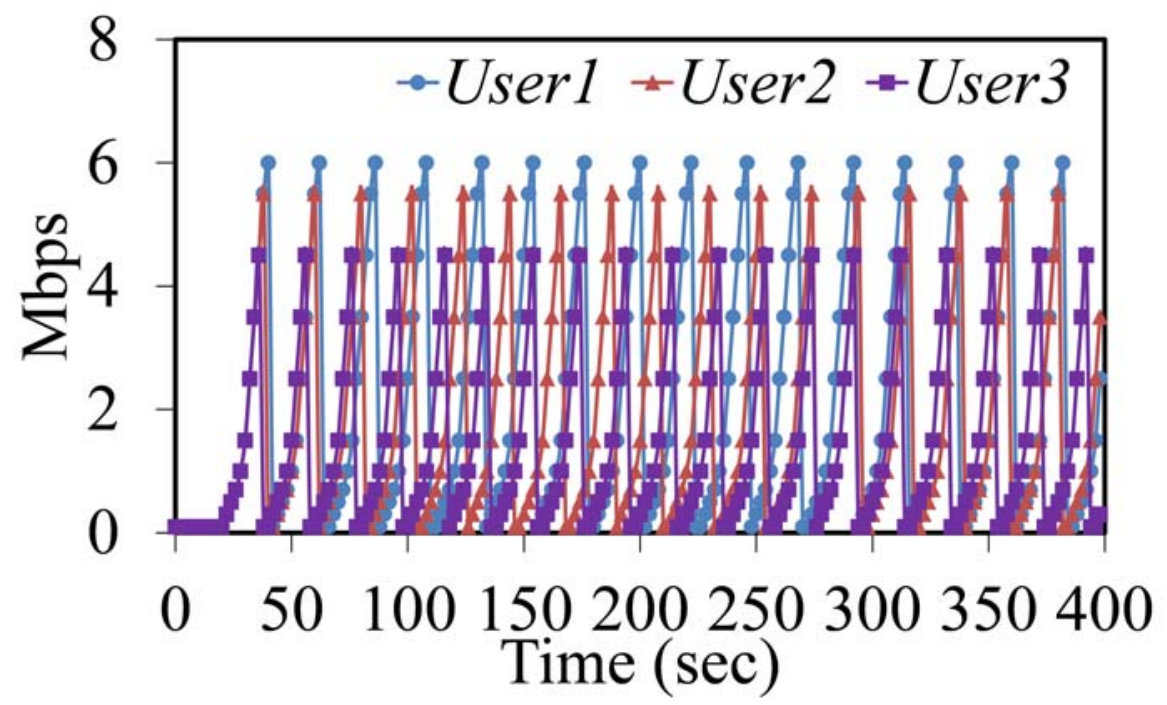}}
\subfigure[]{
\label{fig13:subfig:b} 
\includegraphics[width=4.25cm]{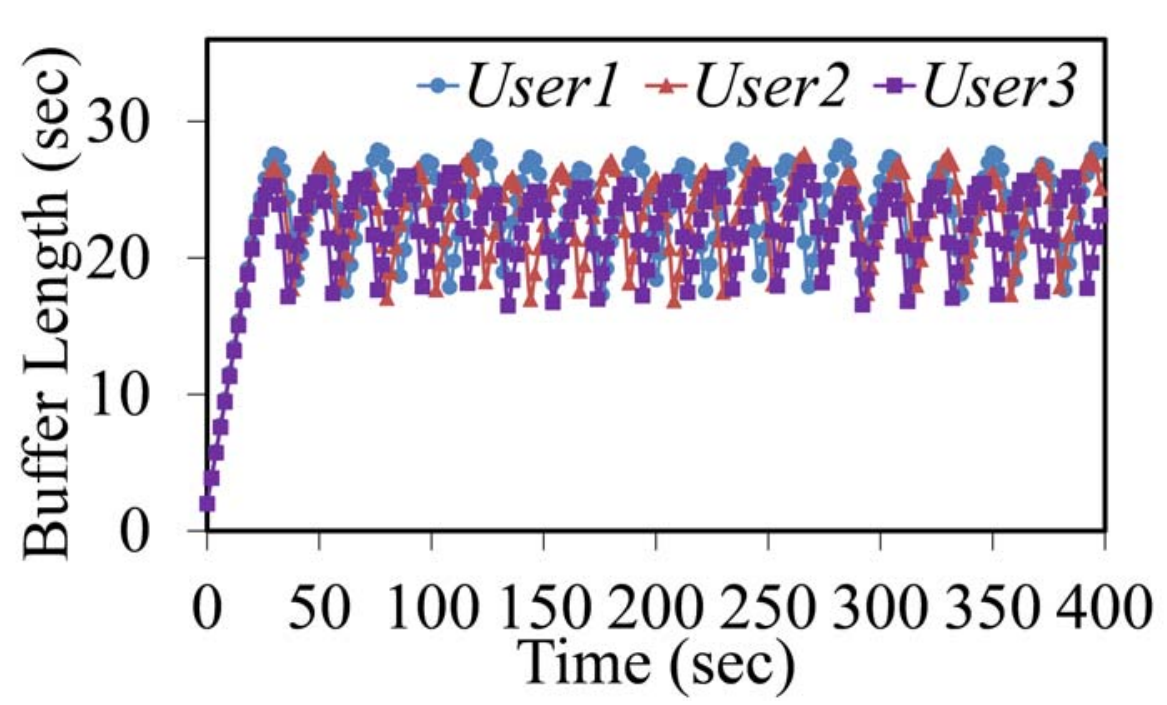}}
\subfigure[]{
\label{fig13:subfig:c} 
\includegraphics[width=4.25cm]{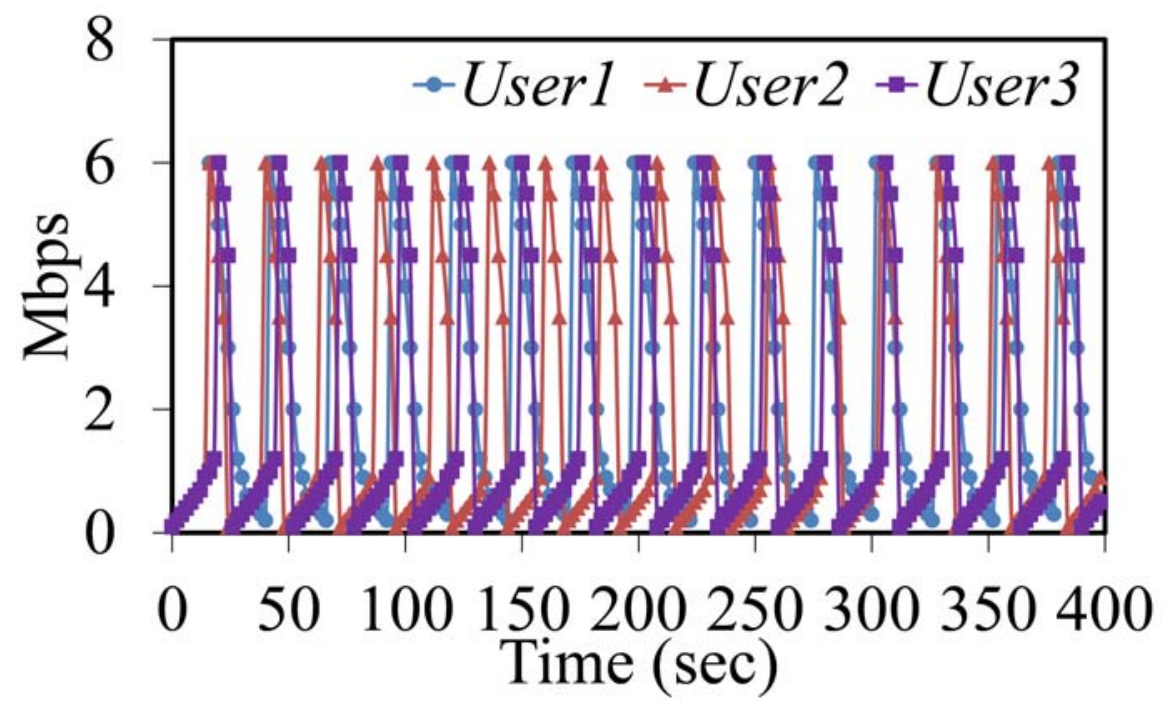}}
\subfigure[]{
\label{fig13:subfig:d} 
\includegraphics[width=4.25cm]{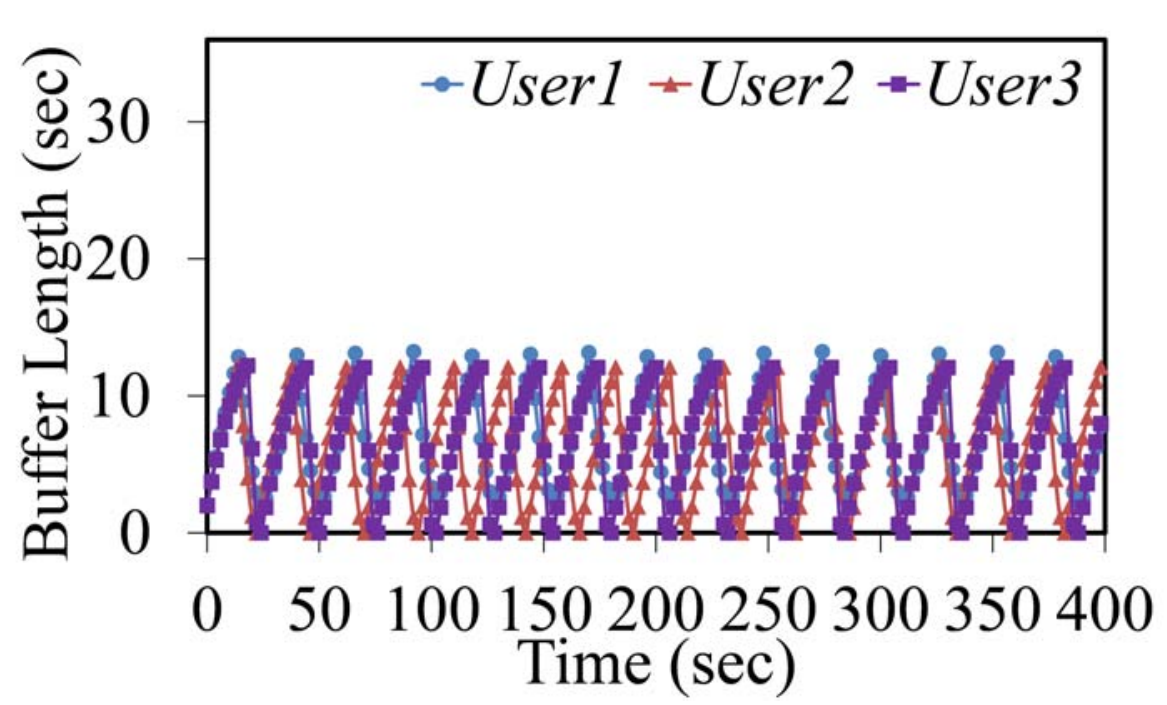}}
\subfigure[]{
\label{fig13:subfig:e} 
\includegraphics[width=4.25cm]{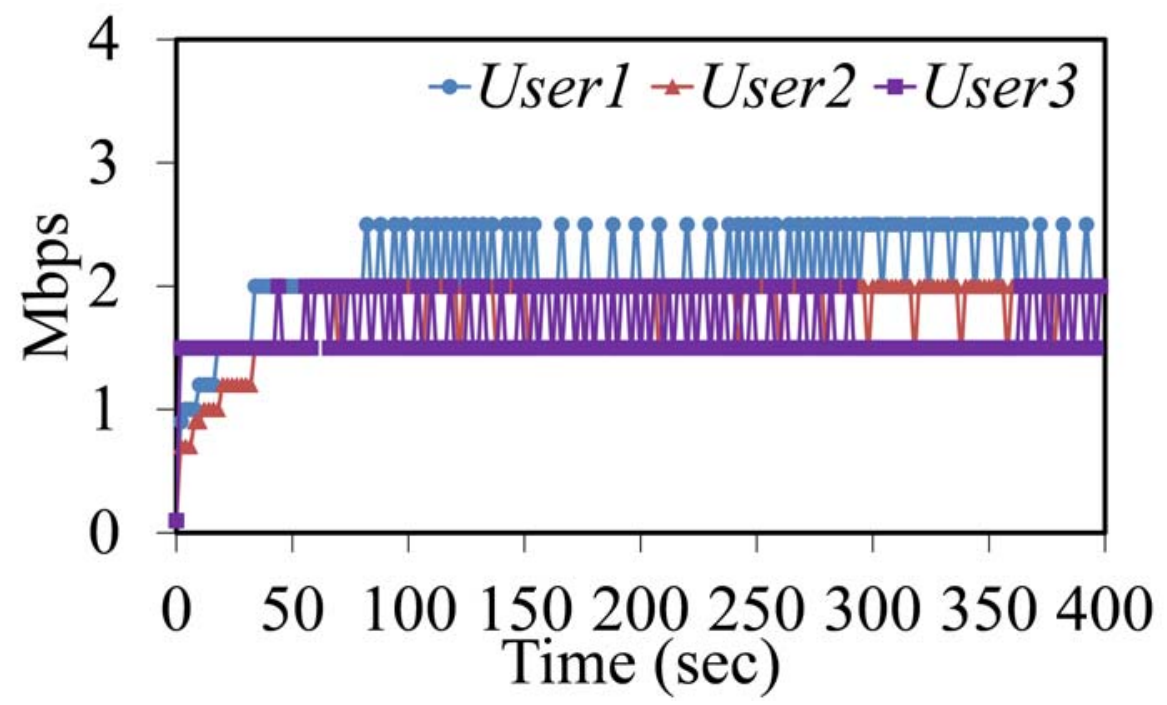}}
\subfigure[]{
\label{fig13:subfig:f} 
\includegraphics[width=4.25cm]{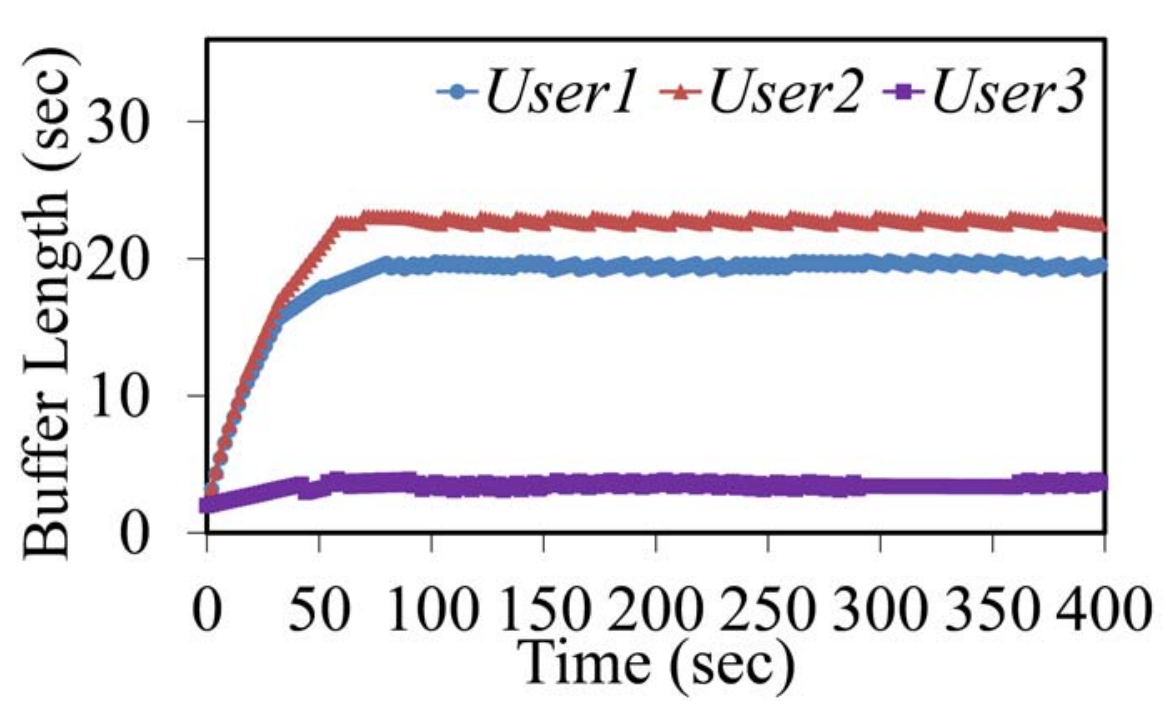}}
\subfigure[]{
\label{fig13:subfig:g} 
\includegraphics[width=4.25cm]{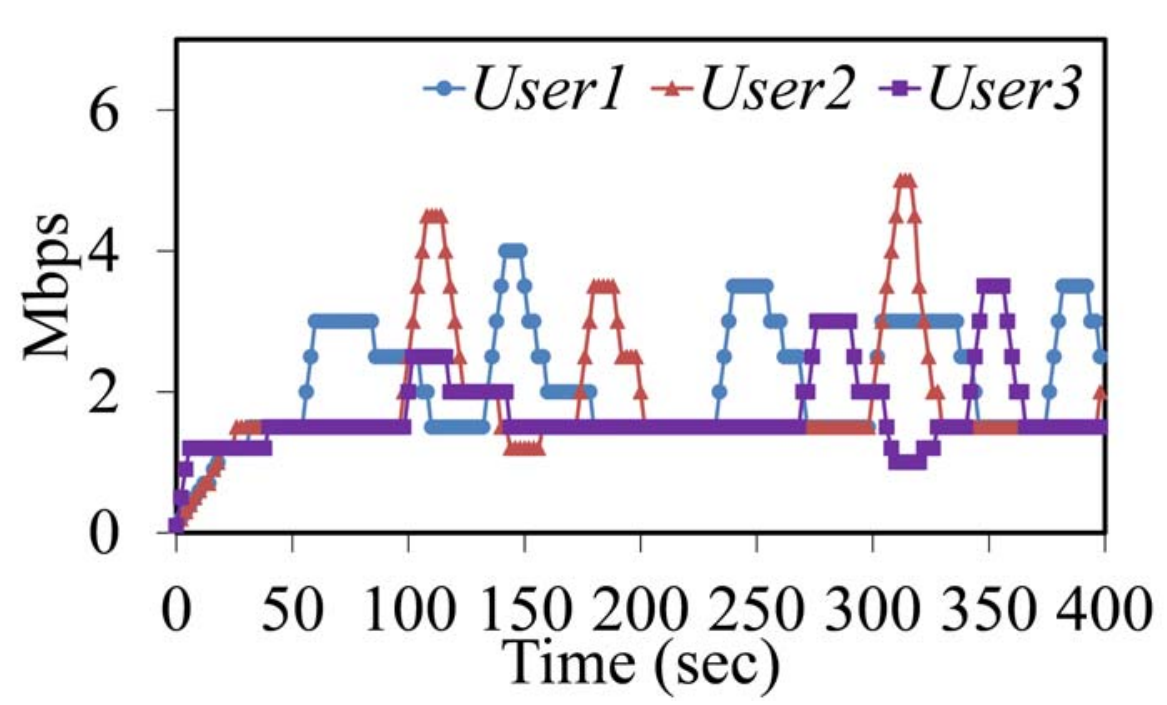}}
\subfigure[]{
\label{fig13:subfig:h} 
\includegraphics[width=4.25cm]{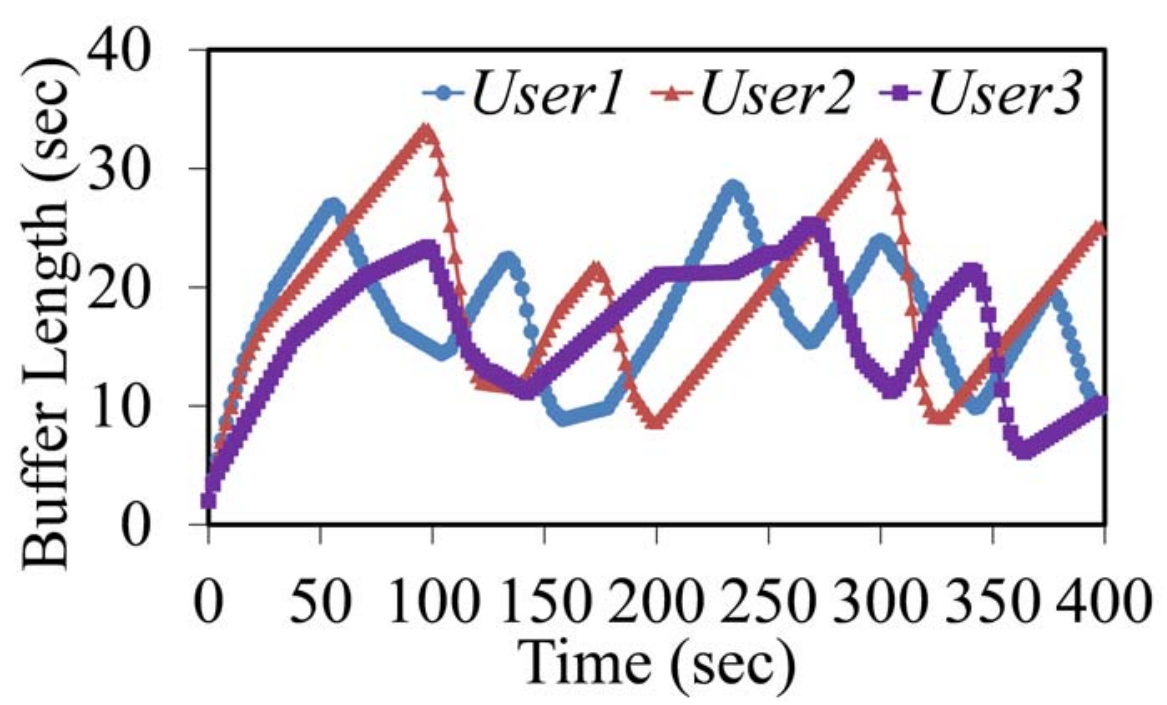}}
\caption{Results of case 4, in which three users (with
\textit{\textbf{random and limited}} throughputs) compete the server
export bandwidth (\textbf{\emph{persistent variation}}), and request
\emph{BigBuckBunny}, \textit{ElephantsDream}, and
\textit{SitaSingstheBlues}, respectively, with learning rate
$\theta=50$. Here, $\mu=0.003$, $\nu=0.0041$. (a), (c), (e), and (g)
show the requested bitrates of \textit{\textbf{BF}},
\textit{\textbf{QF}}, \textit{\textbf{QBA}}, and the
\textit{\textbf{Proposed}} methods; while (b), (d), (f), and (h) show
the corresponding buffer length of each user.}
\label{fig13} 
\end{figure}

\begin{figure}
\setlength{\abovecaptionskip}{0.cm}
\setlength{\belowcaptionskip}{-0.cm} \centering \subfigure[]{
\label{fig14:subfig:a} 
\includegraphics[width=4.25cm]{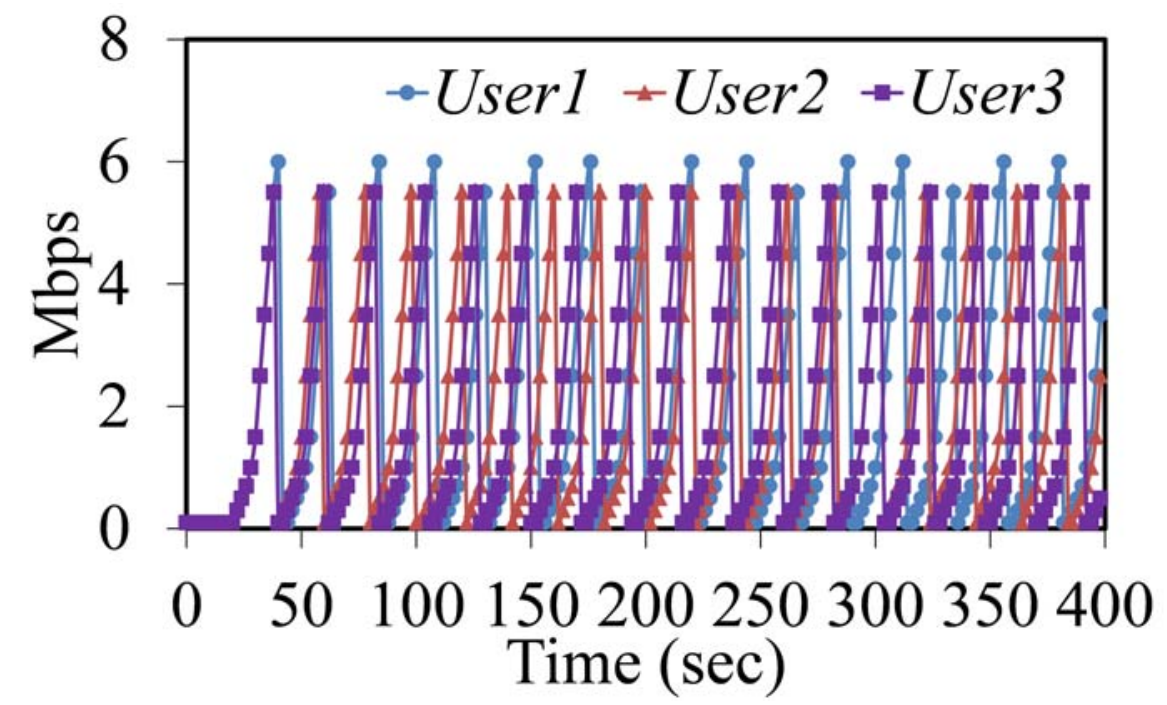}}
\subfigure[]{
\label{fig14:subfig:b} 
\includegraphics[width=4.25cm]{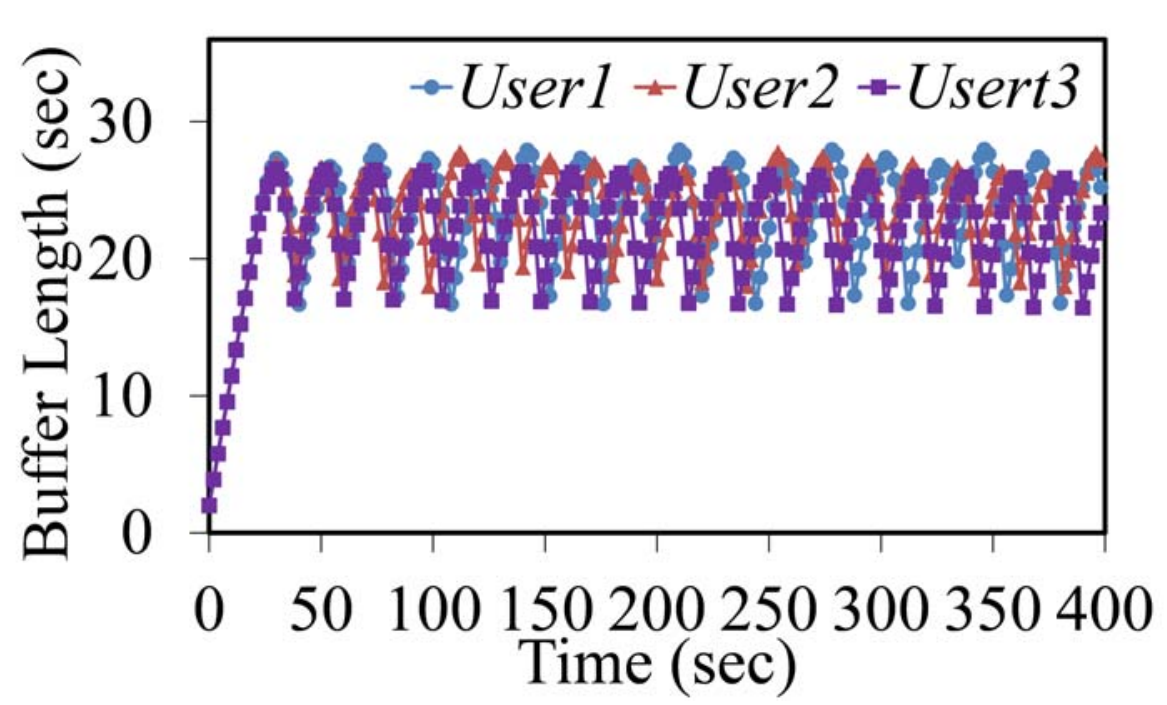}}
\subfigure[]{
\label{fig14:subfig:c} 
\includegraphics[width=4.25cm]{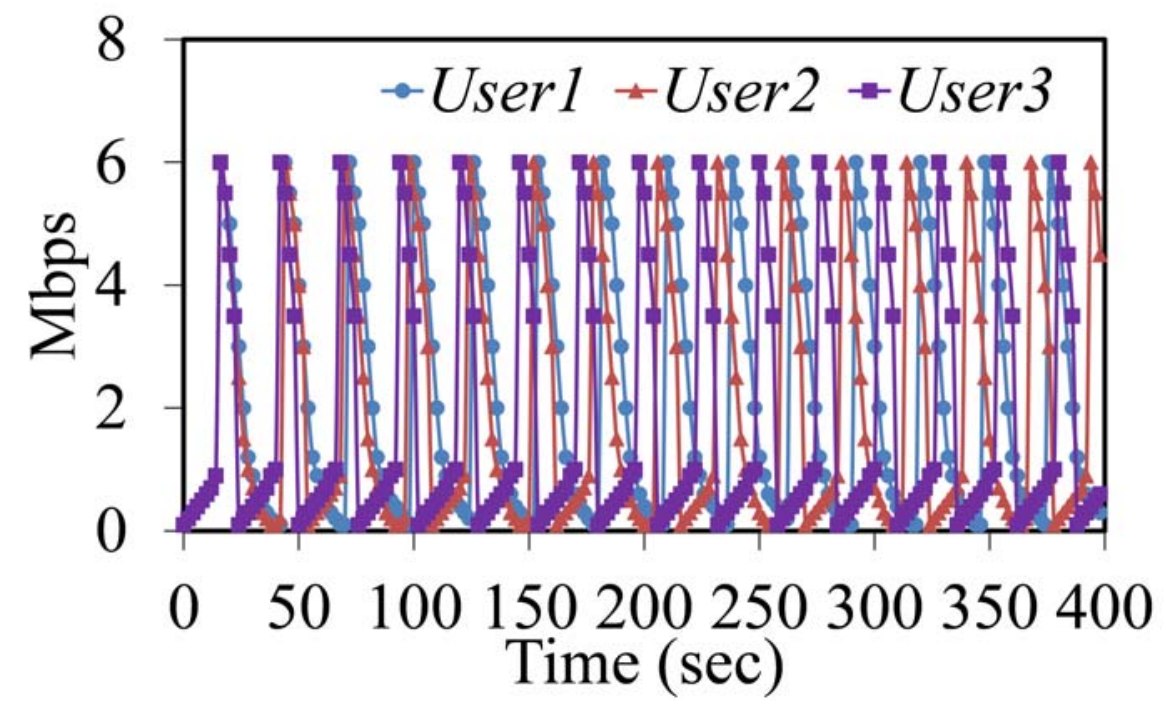}}
\subfigure[]{
\label{fig14:subfig:d} 
\includegraphics[width=4.25cm]{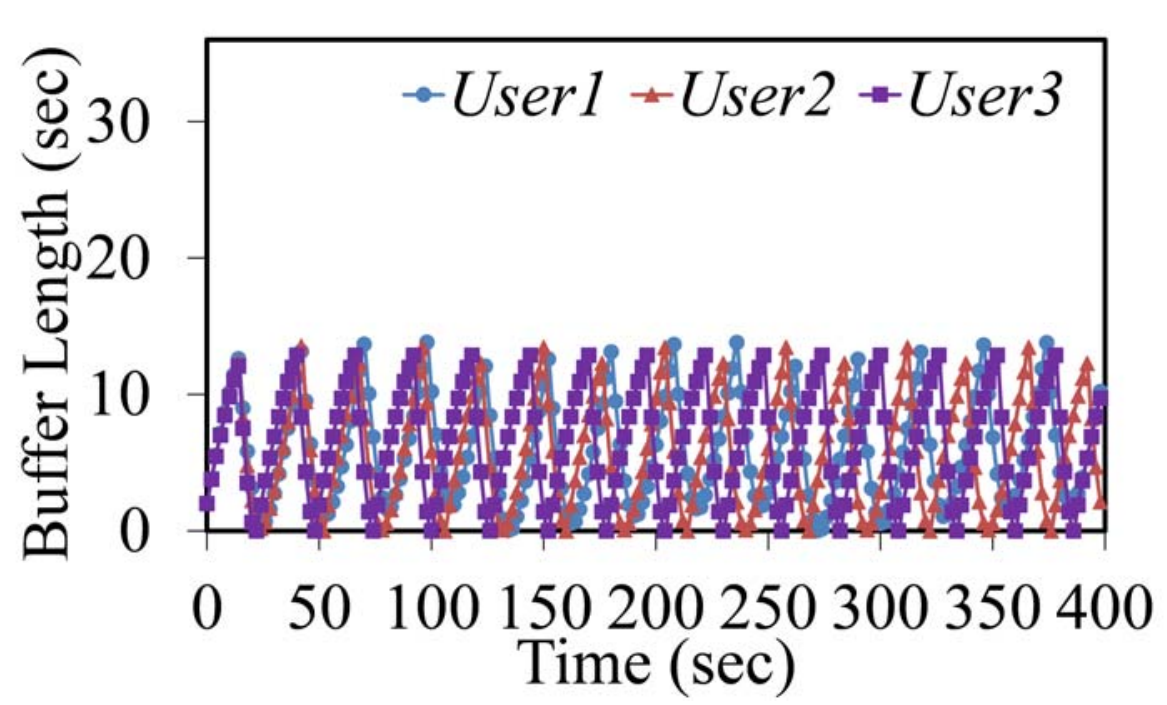}}
\subfigure[]{
\label{fig14:subfig:e} 
\includegraphics[width=4.25cm]{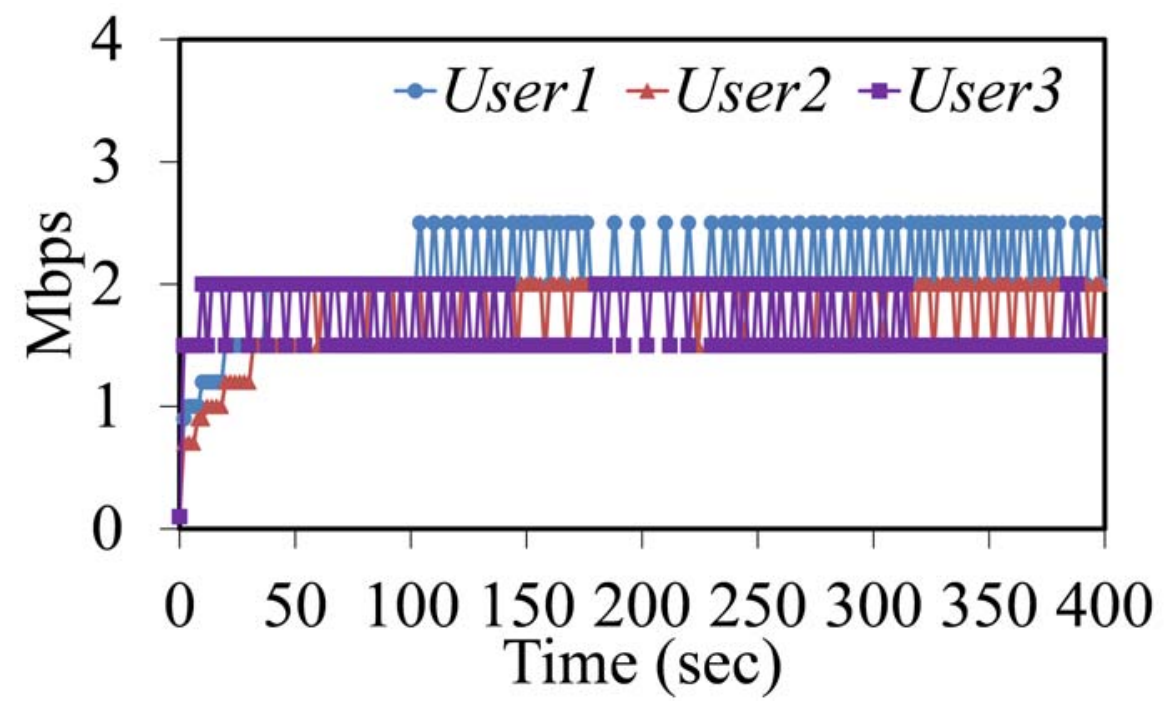}}
\subfigure[]{
\label{fig14:subfig:f} 
\includegraphics[width=4.25cm]{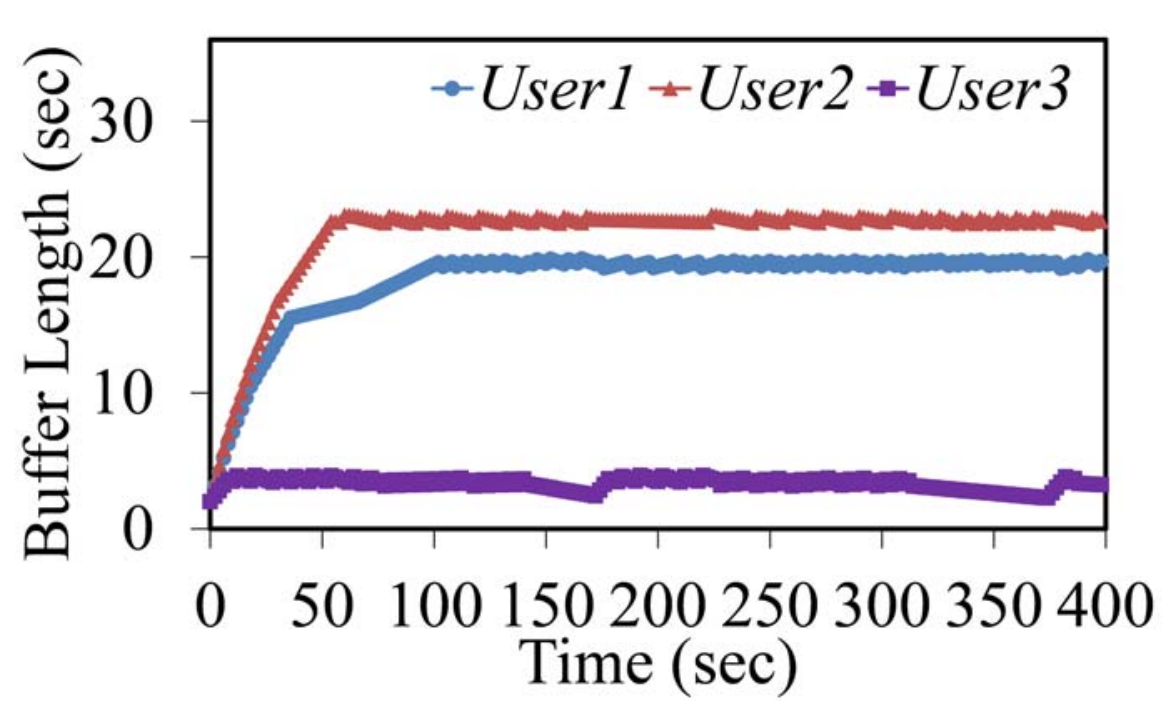}}
\subfigure[]{
\label{fig14:subfig:g} 
\includegraphics[width=4.25cm]{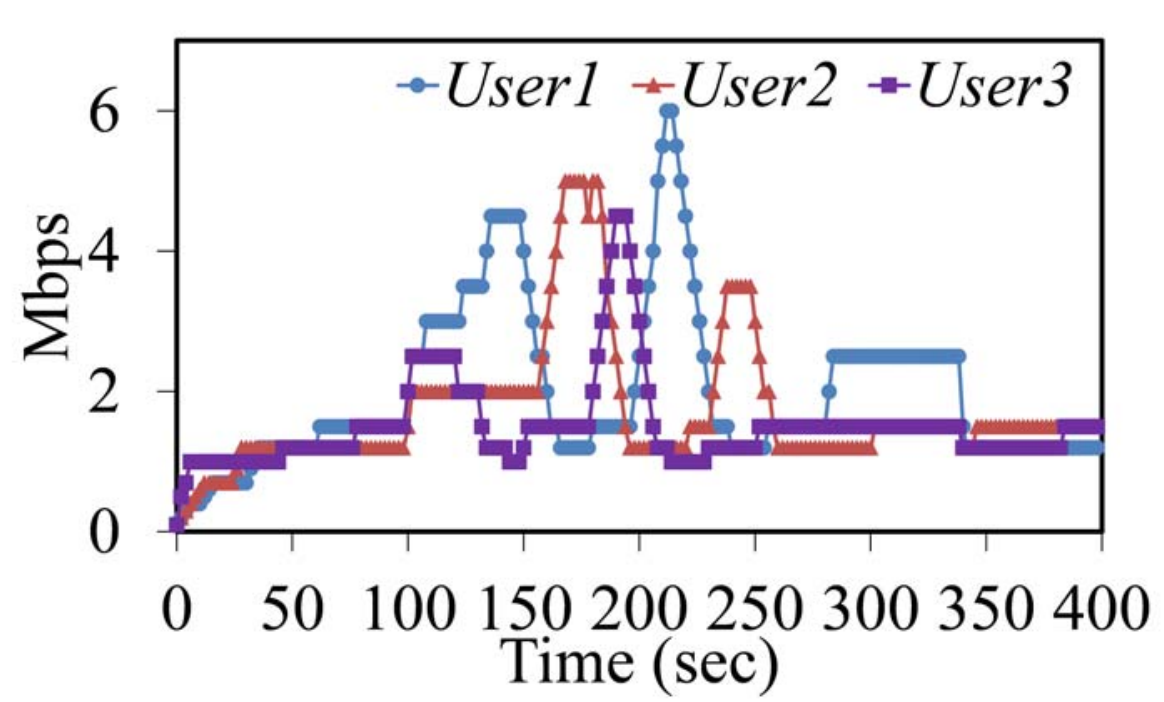}}
\subfigure[]{
\label{fig14:subfig:h} 
\includegraphics[width=4.25cm]{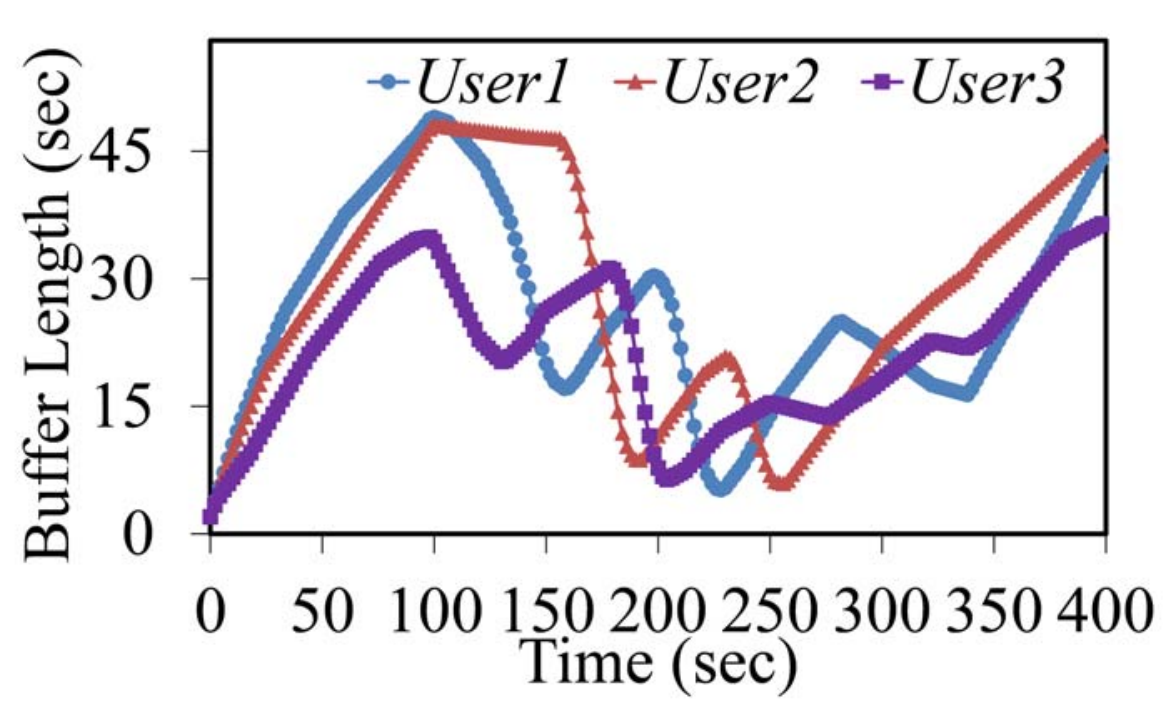}}
\caption{Results of case 4, in which three users (with
\textit{\textbf{random and limited}} throughputs) compete the server
export bandwidth (\textbf{\emph{staged variation}}), and request
\emph{BigBuckBunny}, \textit{ElephantsDream}, and
\textit{SitaSingstheBlues}, respectively, with learning rate
$\theta=50$. Here, $\mu=0.003$, $\nu=0.0041$. (a), (c), (e), and (g)
show the requested bitrates of \textit{\textbf{BF}},
\textit{\textbf{QF}}, \textit{\textbf{QBA}}, and the
\textit{\textbf{Proposed}} methods; while (b), (d), (f), and (h) show
the corresponding buffer length of each user.}
\label{fig14} 
\end{figure}

\begin{figure}
\setlength{\abovecaptionskip}{0.cm}
\setlength{\belowcaptionskip}{-0.cm} \centering \subfigure[]{
\label{fig15:subfig:a} 
\includegraphics[width=4.25cm]{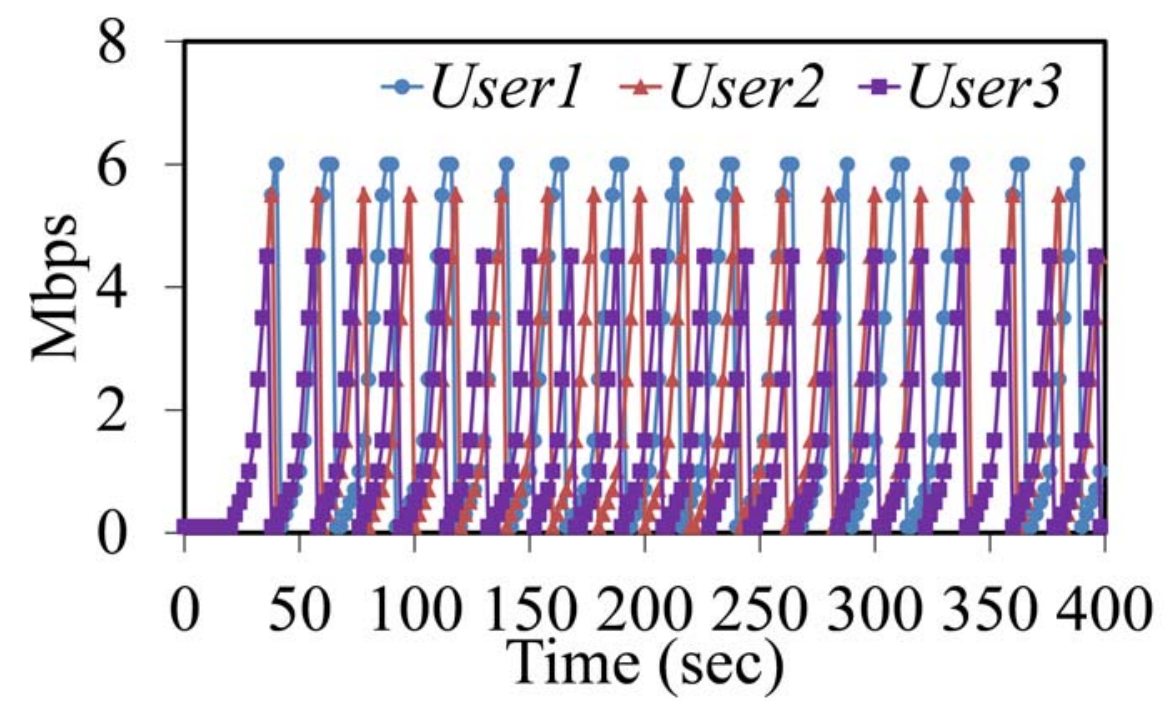}}
\subfigure[]{
\label{fig15:subfig:b} 
\includegraphics[width=4.25cm]{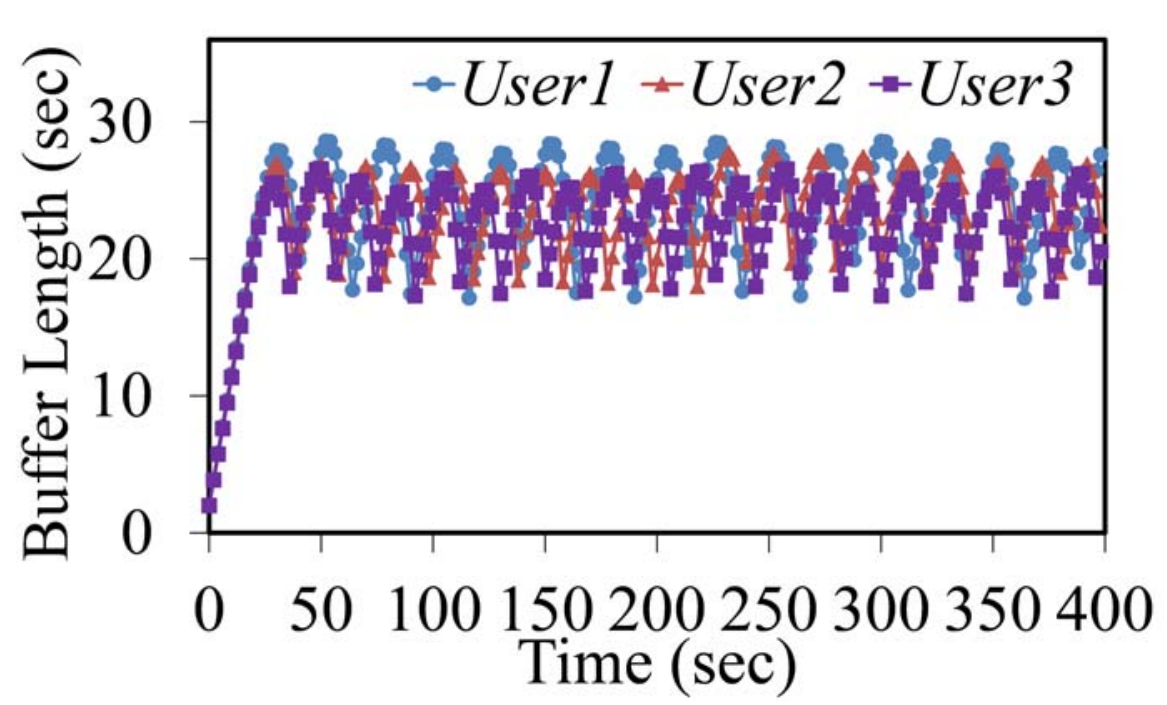}}
\subfigure[]{
\label{fig15:subfig:c} 
\includegraphics[width=4.25cm]{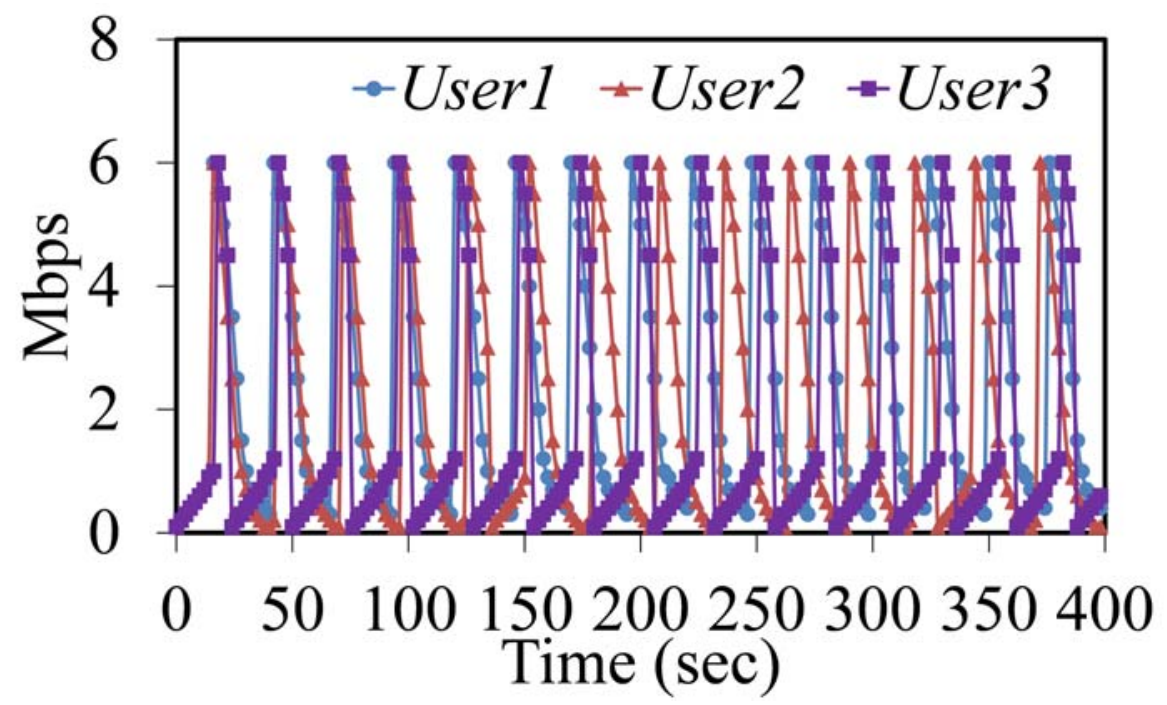}}
\subfigure[]{
\label{fig15:subfig:d} 
\includegraphics[width=4.25cm]{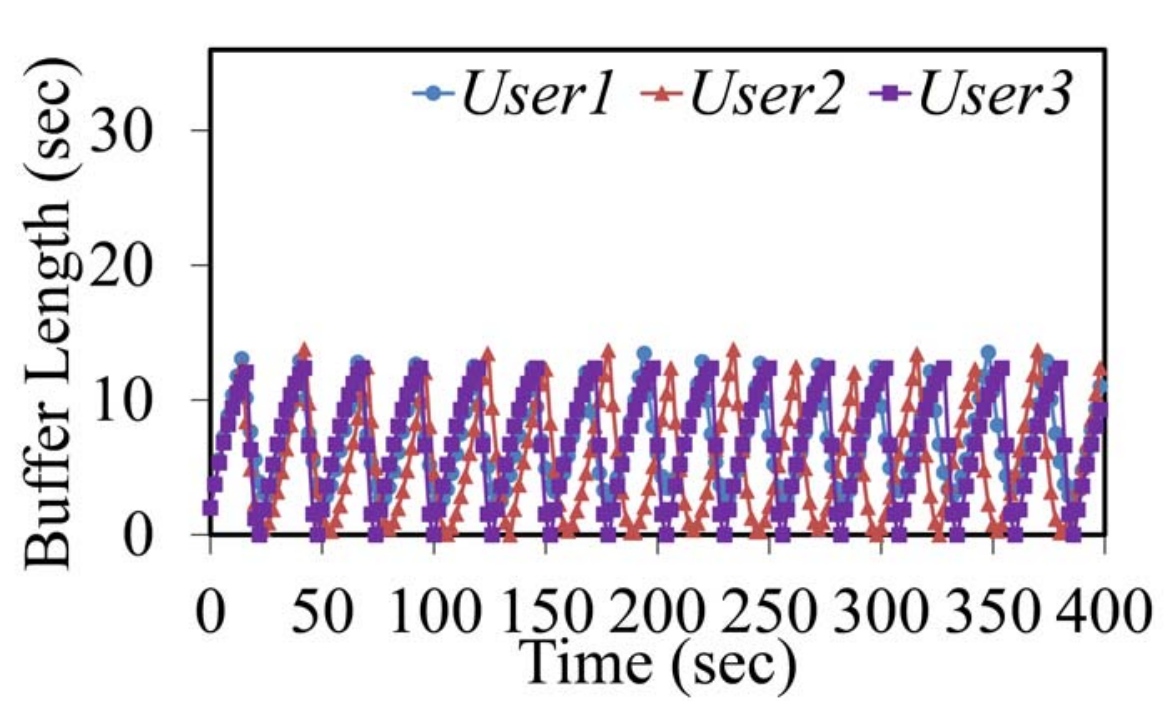}}
\subfigure[]{
\label{fig15:subfig:e} 
\includegraphics[width=4.25cm]{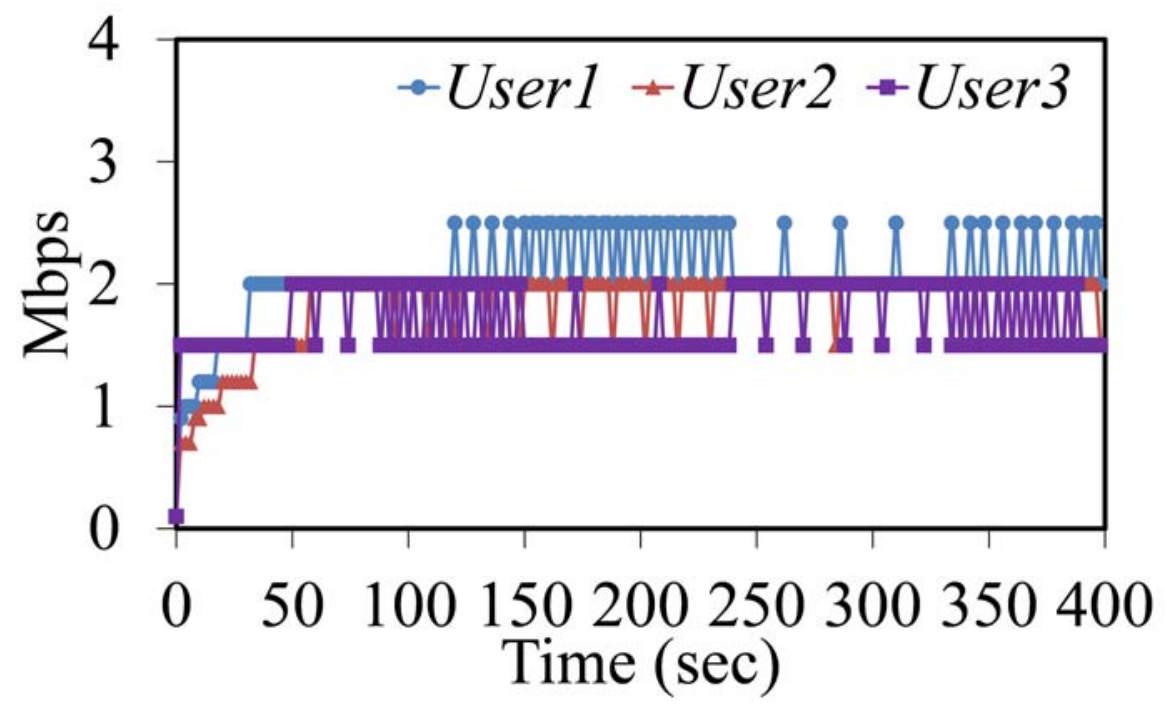}}
\subfigure[]{
\label{fig15:subfig:f} 
\includegraphics[width=4.25cm]{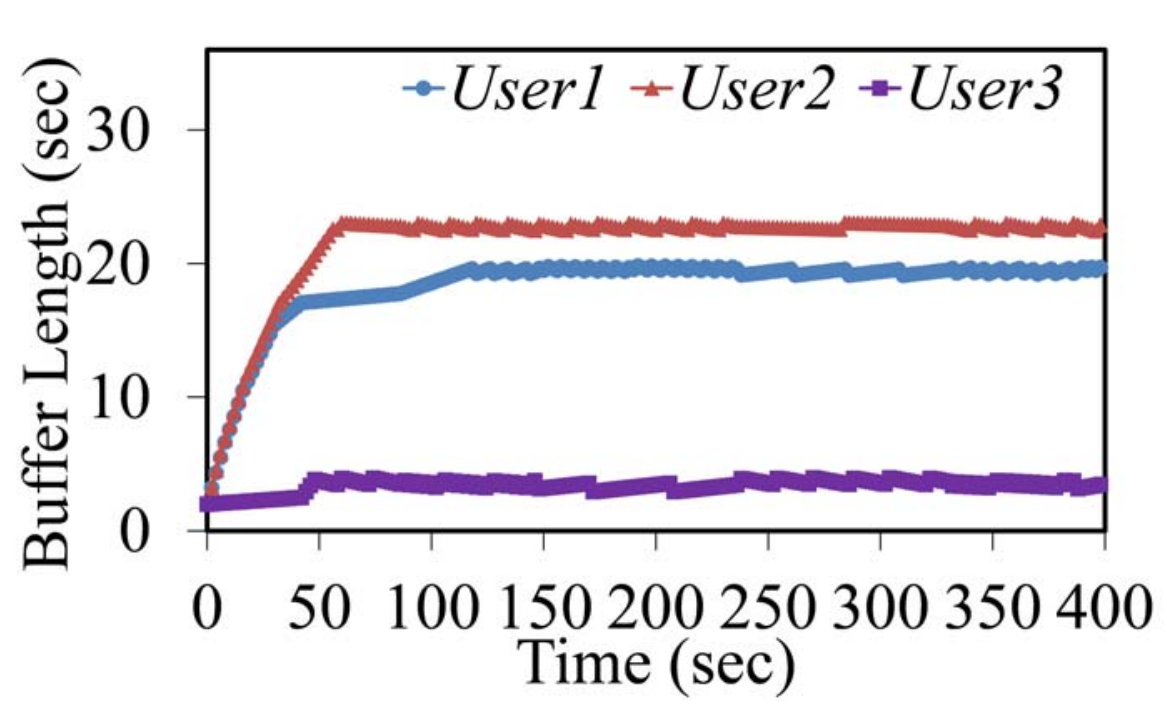}}
\subfigure[]{
\label{fig15:subfig:g} 
\includegraphics[width=4.25cm]{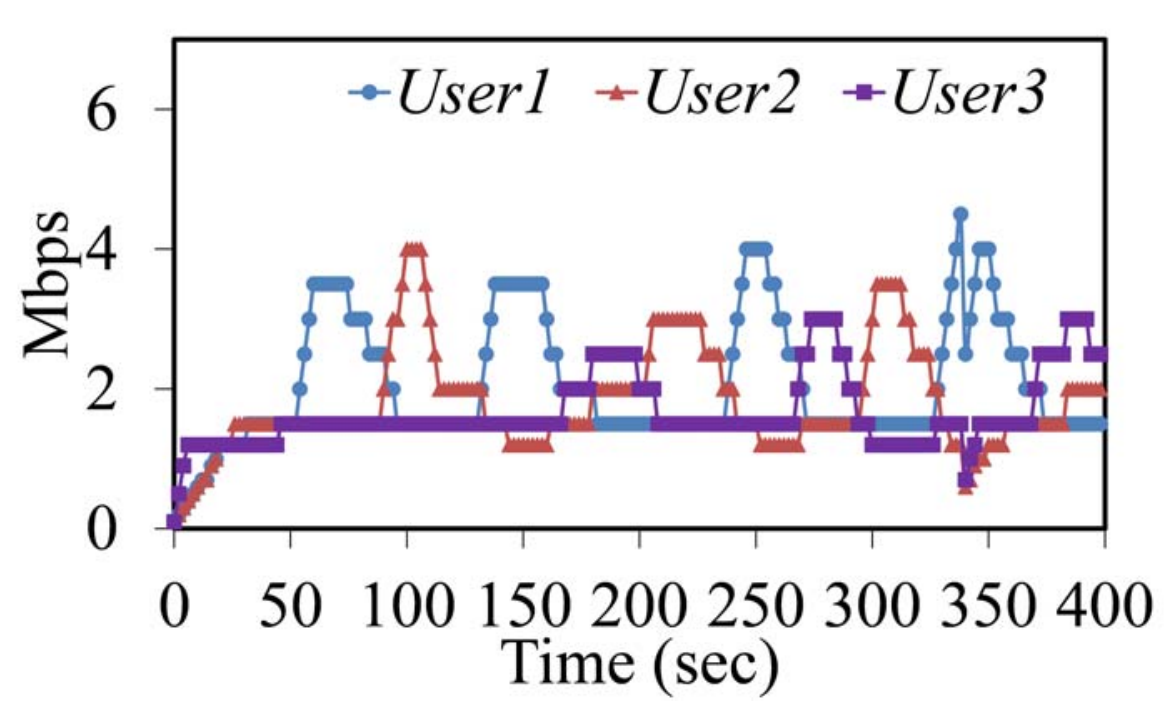}}
\subfigure[]{
\label{fig15:subfig:h} 
\includegraphics[width=4.25cm]{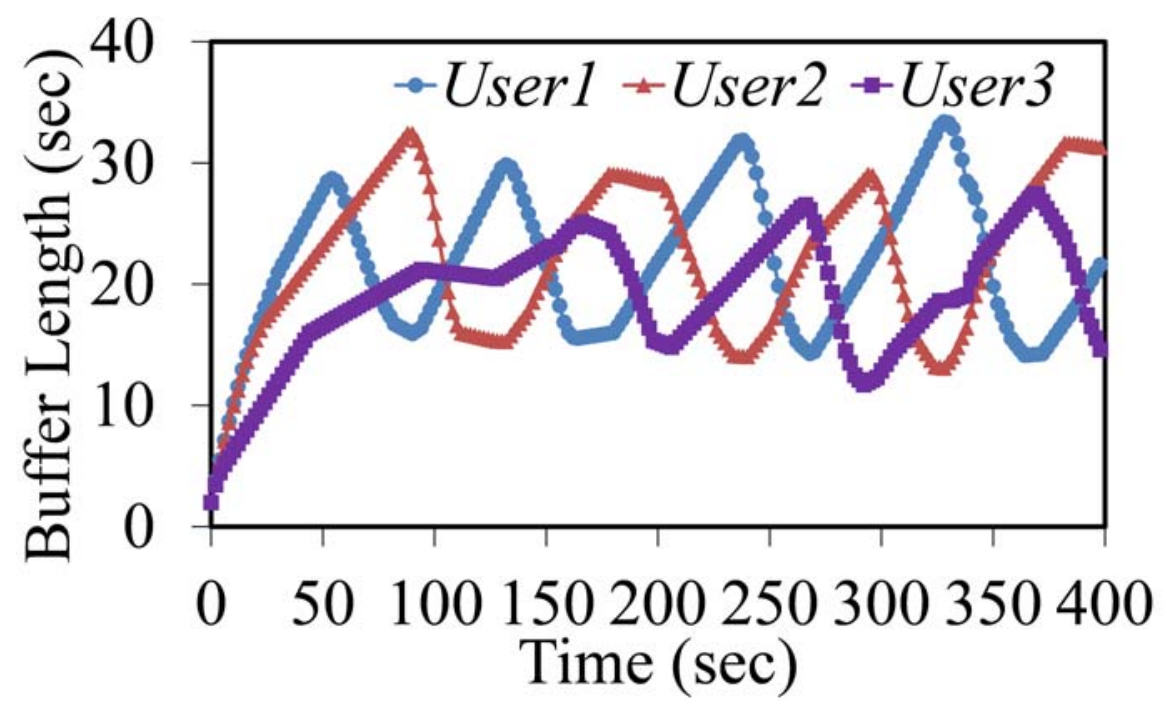}}
\caption{Results of case 4, in which three users (with
\textit{\textbf{random and limited}} throughputs) compete the server
export bandwidth (\textbf{\emph{short-term variation}}), and request
\emph{BigBuckBunny}, \textit{ElephantsDream}, and
\textit{SitaSingstheBlues}, respectively, with learning rate
$\theta=50$. Here, $\mu=0.003$, $\nu=0.0041$. (a), (c), (e), and (g)
show the requested bitrates of \textit{\textbf{BF}},
\textit{\textbf{QF}}, \textit{\textbf{QBA}}, and the
\textit{\textbf{Proposed}} methods; while (b), (d), (f), and (h) show
the corresponding buffer length of each user.}
\label{fig15} 
\end{figure}

Similar to Fig. \ref{fig12}, the corresponding comparison results of
the four methods with respect to the other three types of bandwidth
variations, i.e., \textit{\textbf{persistent variation}},
\textit{\textbf{staged variation}}, and \textit{\textbf{short-term
variation}}, are given in Figs. \ref{fig13}, \ref{fig14}, and
\ref{fig15}, respectively. Similar conclusions can be drawn, which
consistently demonstrates the superiority of the proposed algorithm.

\begin{table*}[htbp]
\setlength{\abovecaptionskip}{0.cm}
\setlength{\belowcaptionskip}{-0.cm}
  \centering
  \caption{Detailed Comparisons of the Four Methods under Case 4 (The best results are bolded).}
\includegraphics[width=15cm]{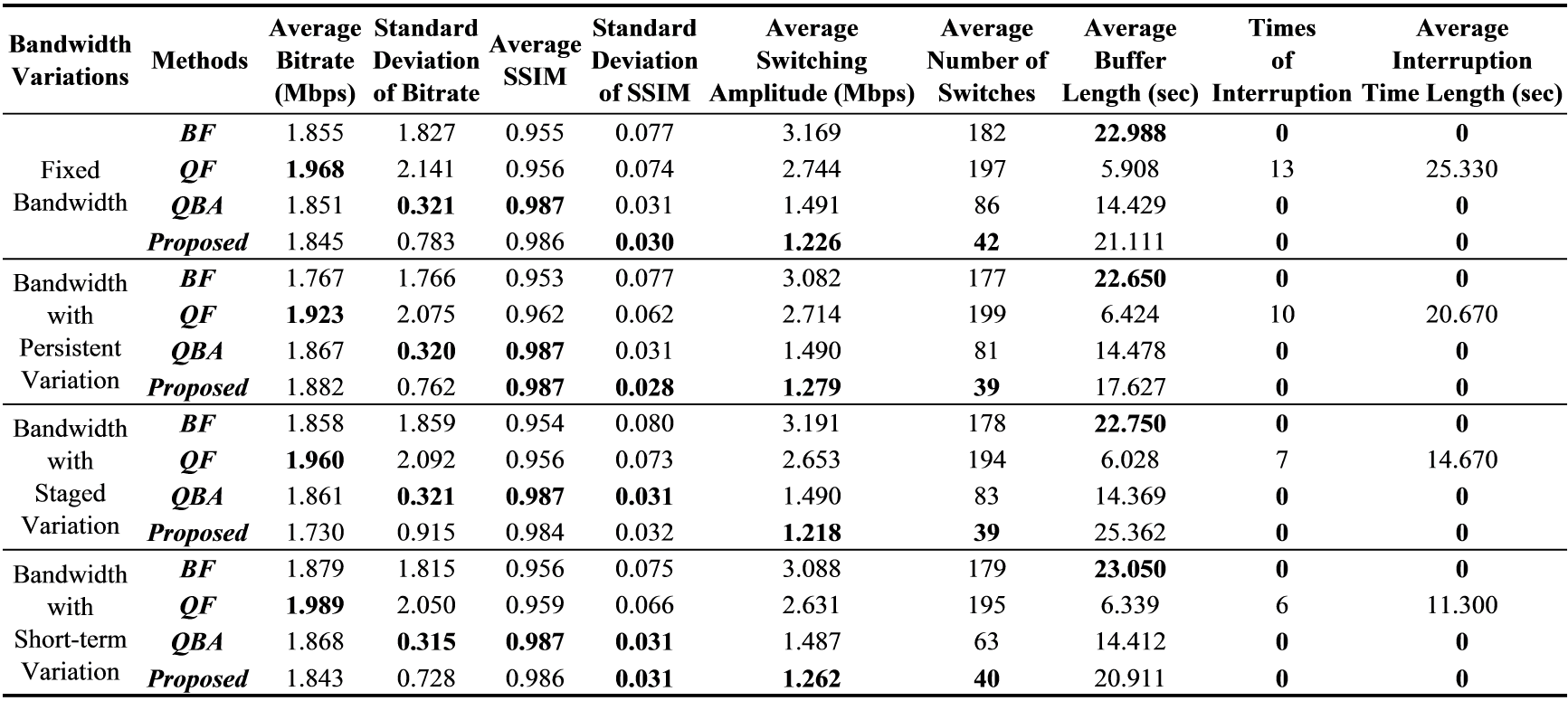}
\label{table3}
\end{table*}

Detailed numerical comparisons of the four methods are given in TABLE
\ref{table3}, from which we can observe that the average received
bitrate of the \textit{\textbf{QF}} method is largest, but the
bitrate fluctuations (see the standard deviation, average number of
switches and average switching amplitude of received bitrates) are
also the maximum, and there exist playback interruptions, while the
bitrate fluctuation of the \textit{\textbf{BF}} method is a little
small but still large. It can also be observed that the average video
quality (by observing the average SSIMs) of the \textit{\textbf{QF}}
and \textit{\textbf{BF}} methods is obviously lower than that of the
\textit{\textbf{QBA}} and the \textit{\textbf{Proposed}} methods, and
the quality variations (by observing the standard deviation of SSIM
values of received video segments) of the \textit{\textbf{QF}} and
the \textit{\textbf{BF}} methods are larger than those of the
\textit{\textbf{QBA}} and the \textit{\textbf{Proposed}} methods.
Moreover, although the average bitrates and SSIM values of the
\textit{\textbf{Proposed}} method are similar to those of the
\textit{\textbf{QBA}} method, it is obvious that the amplitudes of
bitrate switching and the numbers of switches of
\textit{\textbf{Proposed}} method are smaller, which means that the
performance of the \textit{\textbf{Proposed}} method is the best.

\begin{table}[htbp]
\setlength{\abovecaptionskip}{0.cm}
\setlength{\belowcaptionskip}{-0.cm}
  \centering
  \caption{Comparisons of the Four Methods under Case 4 in terms of the \textit{QoE} metric in \cite{R50} (The best results are bolded)}
\includegraphics[width=8cm]{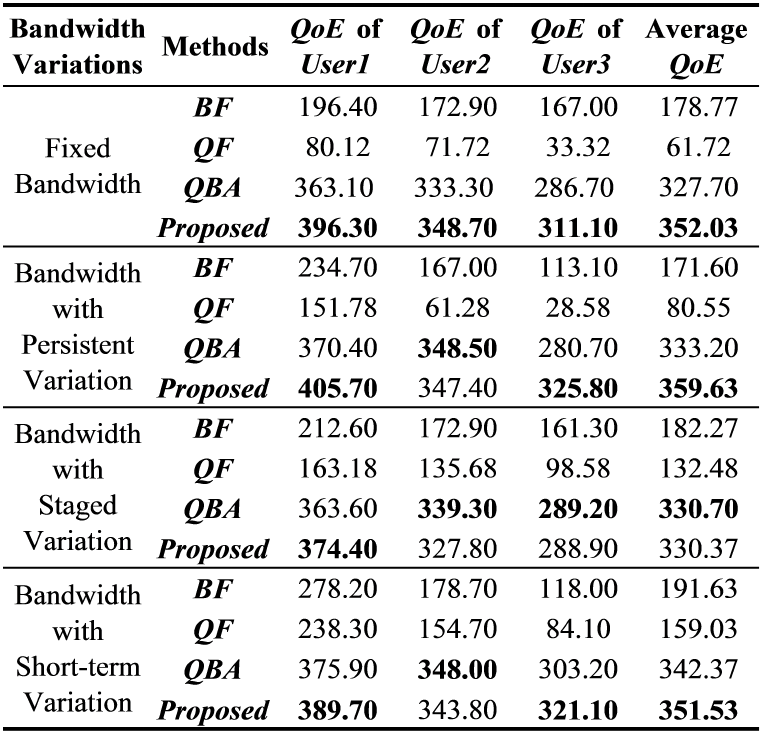}
\label{table4}
\end{table}

\begin{table}[htbp]
\setlength{\abovecaptionskip}{0.cm}
\setlength{\belowcaptionskip}{-0.cm}
  \centering
  \caption{Comparisons of the Four Methods under Case 4 in terms of the \textit{QoE} metric in \cite{R51} (The best results are bolded)}
\includegraphics[width=8cm]{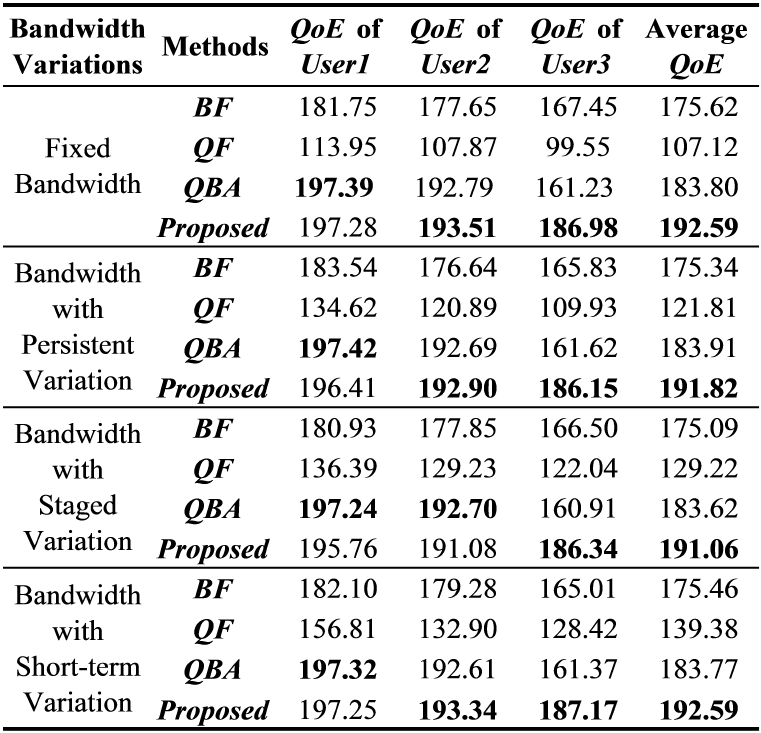}
\label{table5}
\end{table}

Last, we evaluate the performance of the four algorithms, i.e.,
\emph{\textbf{BF}}, \emph{\textbf{QF}}, \textbf{\emph{QBA}}, and
\emph{\textbf{Proposed}}, by comparing their produced \emph{QoE}
values that are measured via two extensively used \emph{QoE} models
\cite{R50}\cite{R51}:
\begin{equation}\label{E36}
\begin{aligned}
    QoE_{1}=&\sum^{M}_{k=1}r[k]-\xi\sum^{M-1}_{k=1}|r[k+1]-r[k]|\\
    &-\psi\sum^{M}_{k=1}\max\{0,T_{down}[k]-b[k]\},
\end{aligned}
\end{equation}
\begin{equation}\label{E37}
\begin{aligned}
    QoE_{2}=&\sum^{M}_{k=1}q[k]-\varphi\sum^{M-1}_{k=1}|q[k+1]-q[k]|\\
    &-\sigma\sum^{M-1}_{k=1}\left(\max\left(0,b_{ref}-b[k+1]\right)\right)^{2}\\
    &-\eta\sum^{M}_{k=1}\max\{0,T_{down}[k]-b[k]\},
\end{aligned}
\end{equation}
where $\xi=1$, $\psi=6$, $\varphi=2$, $\sigma=0.001$ and $\eta=2$ are
model parameters that are empirically defined in \cite{R50} and
\cite{R51}, \emph{M} is the number of received segments, $r[k]$ is
the bitrate of the $k^{th}$ requested video segment, $q[k]$ is the
corresponding SSIM value, $T_{down}[k]$ is the download time of the
$k^{th}$ segment, and $b[k]$ is the buffer length at the end time of
the $k^{th}$ segment and $b_{ref}=15$s. Note that we set $\eta$ to 2
instead of 50 in \cite{R51} to ensure that the \textit{QoE} values
are positive. From TABLE \ref{table4} and \ref{table5}, it can be
observed that, compared with the other three methods, the proposed
algorithm always produces the highest average \textit{QoE} with
respect to different types of bandwidth variations. Besides, the
proposed method provides optimal \textit{QoE} for most users, i.e.,
at least 2 out of 3 users.


\section{Experimental Result for Realistically Modeled Networking Scenarios}

We also established realistically modeled wireless and wired
networking scenarios, in which 6 users requested different videos
from a single server. In the wireless network, they were connected by
a movable router (2.4GHz, automatic frequency channel bandwidth,
802.11b/g/n mixed wireless mode, 1480 Byte maximum transmission unit
configuration); yet in the wired network, the users connected to the
server via the campus network of Shandong University which includes
many switches and routers. To ensure a large server export bandwidth
(which can be constrained to 6Mbps by \emph{DummyNet}), the
experiments were conducted at night. The initial buffer of each user
was set as 20s in the experiments in order to calculate the initial
playout delay. The performance of the proposed algorithm was verified
under 2 cases:

\emph{\textbf{Case 1}}, 6 users requested different video contents
all the time;

\emph{\textbf{Case 2}}, 6 users requested different video contents at
the beginning, and then some users leaved or joined the network.

\subsection{Results of Case 1}
\begin{figure}
\setlength{\abovecaptionskip}{0.cm}
\setlength{\belowcaptionskip}{-0.cm} \centering \subfigure[]{
\label{fig16:subfig:a} 
\includegraphics[width=4.25cm]{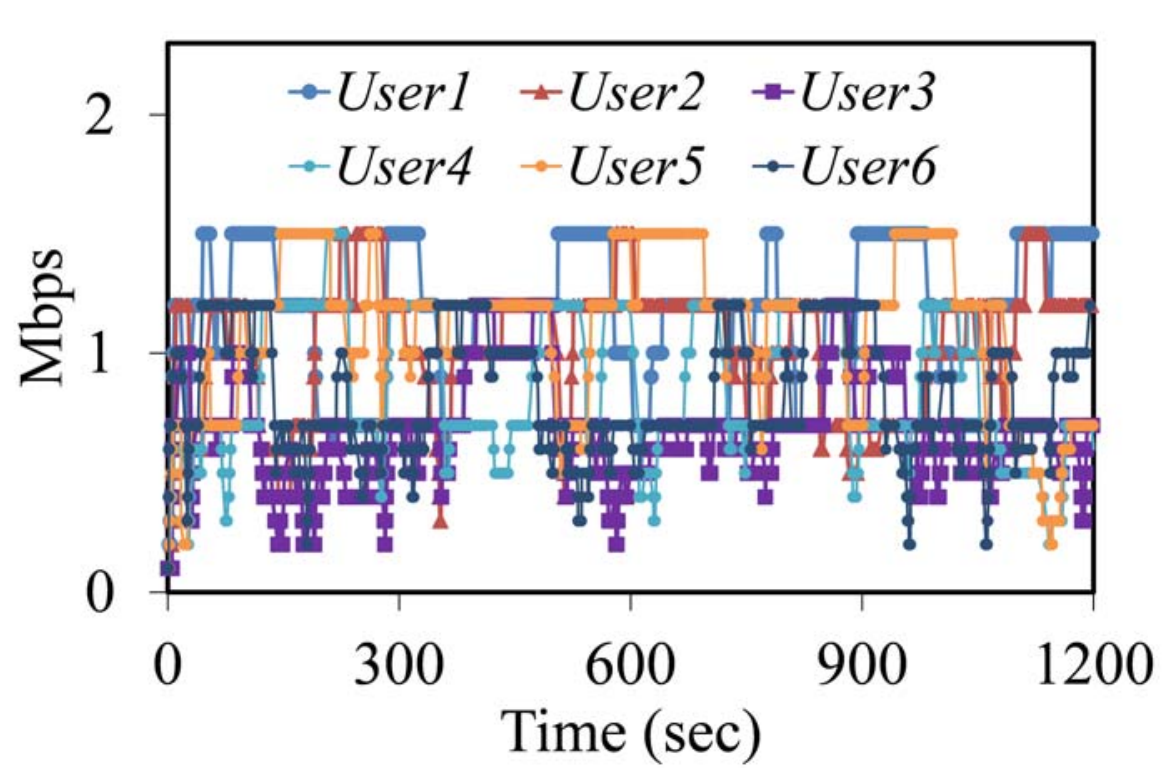}}
\subfigure[]{
\label{fig16:subfig:b} 
\includegraphics[width=4.25cm]{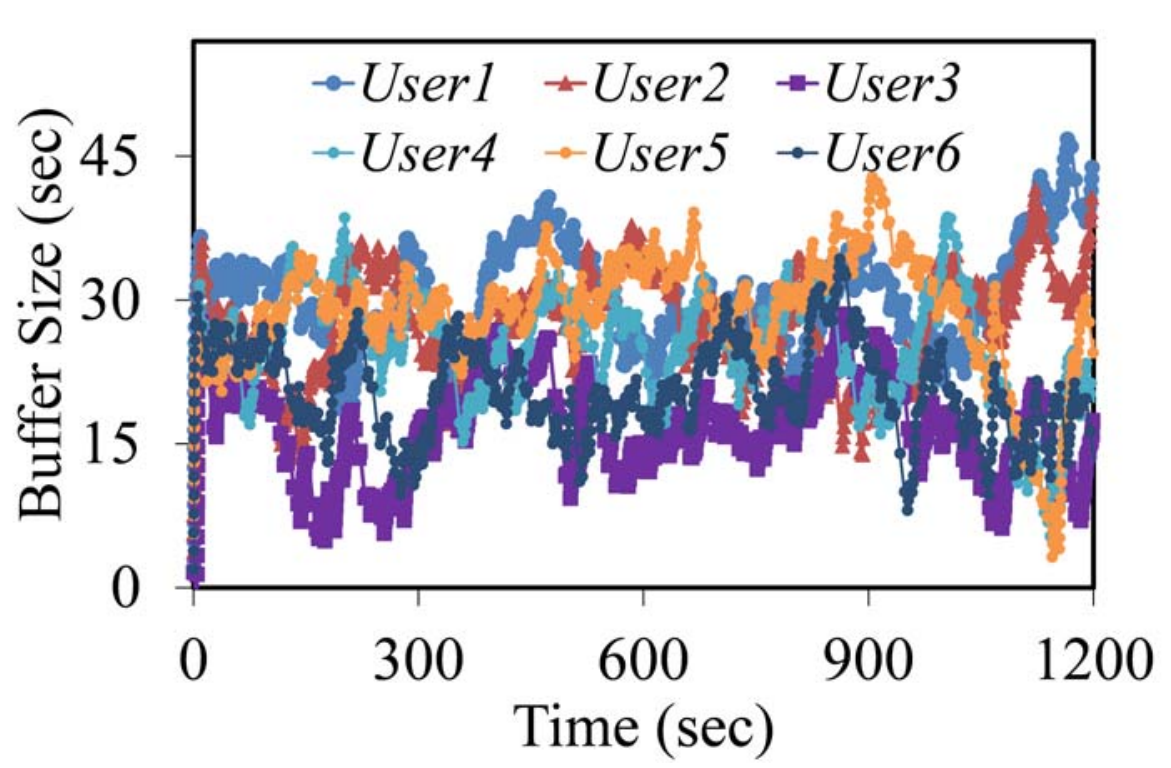}}
\caption{Experimental Results of realistically modeled wireless
network, in which 6 users compete the server export bandwidth (6
Mbps) with $\theta=40$, $\mu=0.003$, and $\nu=0.0041$, (a) dynamic
behavior of requested bitrates, (b) the actual buffer lengths of the
6 users with the reference buffer length of 20s.}
\label{fig16} 
\end{figure}

\begin{figure}
\setlength{\abovecaptionskip}{0.cm}
\setlength{\belowcaptionskip}{-0.cm} \centering \subfigure[]{
\label{fig17:subfig:a} 
\includegraphics[width=4.25cm]{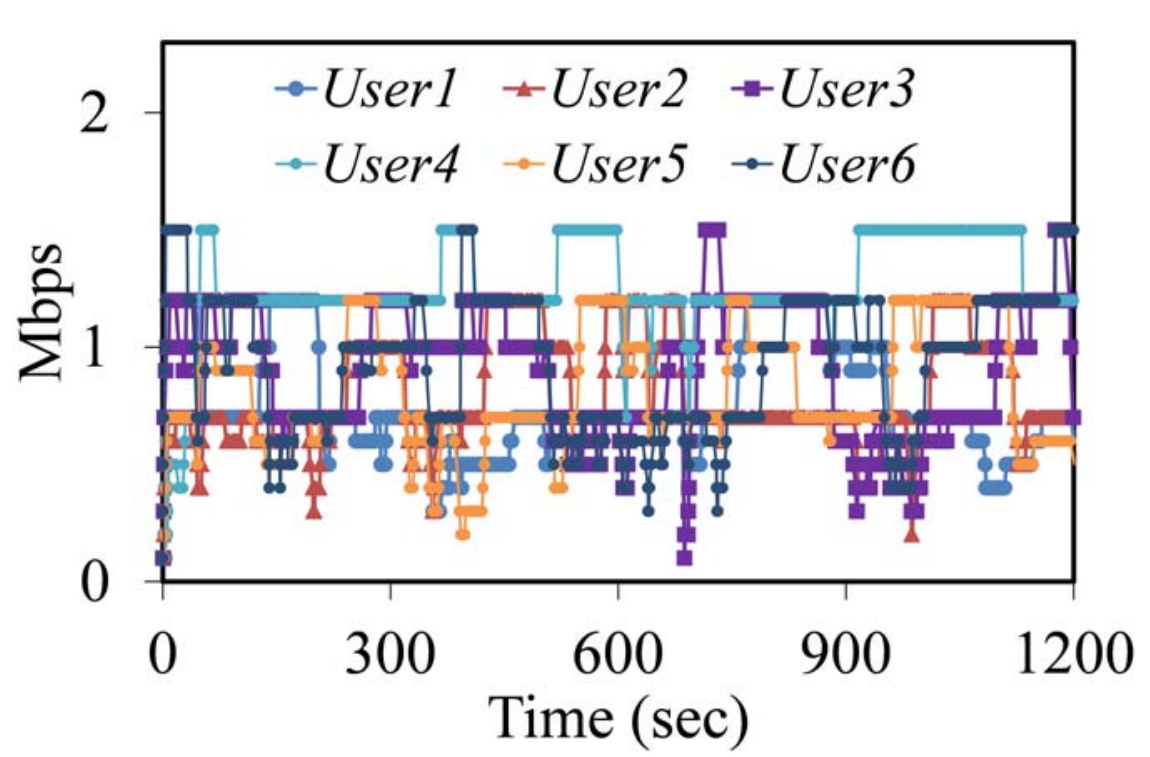}}
\subfigure[]{
\label{fig17:subfig:b} 
\includegraphics[width=4.25cm]{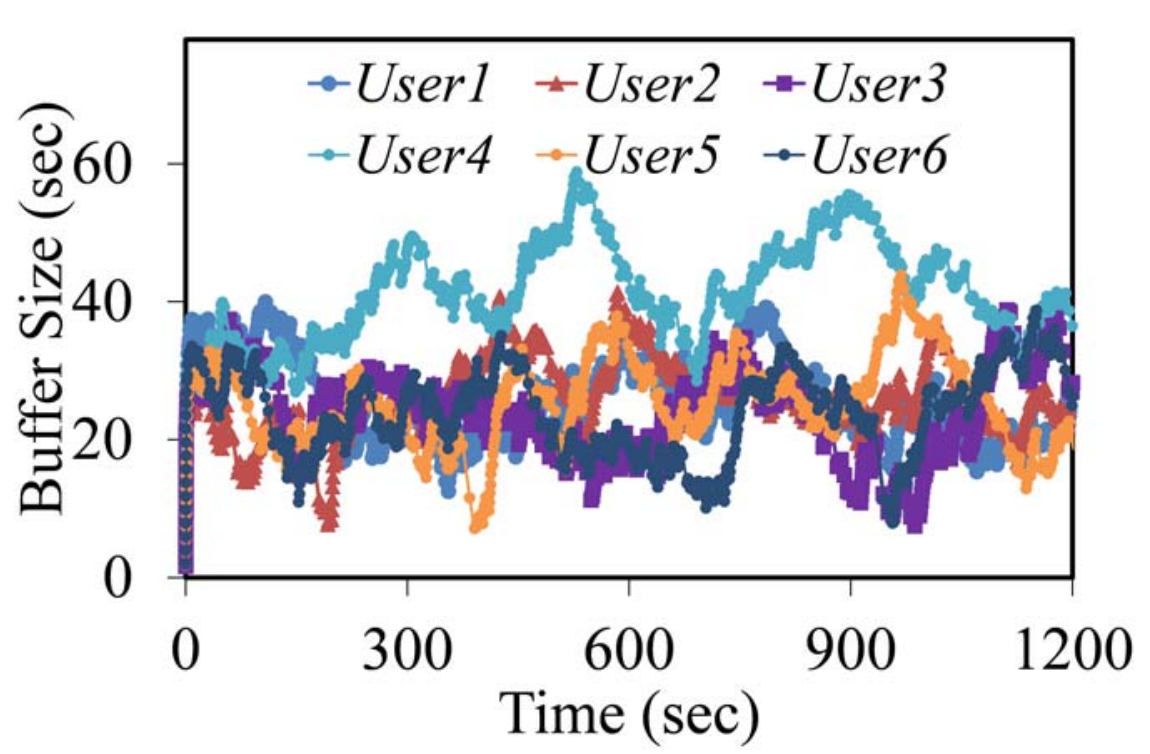}}
\caption{Experimental Results of realistically modeled wired network,
in which 6 users compete the server export bandwidth (6 Mbps) with
$\theta=30$, $\mu=0.003$, and $\nu=0.0041$, (a) dynamic behavior of
requested bitrates, (b) the actual buffer lengths of the 6 users with
the reference buffer length of 20s.}
\label{fig17} 
\end{figure}

The results of the wireless and wired networks are given in Figs.
\ref{fig16} and \ref{fig17}, respectively. We can observe that the
fluctuations of the requested bitrates and buffer lengths are larger
than those of the simulated networking scenario because of the
complexity of the realistic networking environment. From TABLE
\ref{table6}, we can observe that the average initial playout delays
of the \emph{\textbf{QF}} and the \emph{\textbf{BF}} methods are
similar, i.e., 3.61s and 2.70s, for the wireless networking
environment and 0.88s and 1.47s for the wired networking environment.
Similar to the results of the simulation, the average received
bitrates of the \emph{\textbf{Proposed}} method are not the largest
but are comparable with the other methods. More importantly, we can
see that there is no playback interruption under both the wireless
and wired networking environment for the proposed algorithm, which is
also demonstrated in Figs. \ref{fig16:subfig:b} and
\ref{fig17:subfig:b}. As shown in TABLE \ref{table6}, although there
is also no playback interruption for the \emph{\textbf{BF}} method,
the received bitrate fluctuations (average standard deviation) of all
the 6 users are much larger than the \emph{\textbf{Proposed}} method.
From Figs. \ref{fig16:subfig:a} and \ref{fig17:subfig:a}, we can see
that the requested bitrates of the \emph{\textbf{Proposed}} method
fluctuate around 1Mbps (the total server export bandwidth is set as
6Mbps). Moreover, the average buffer length of the
\emph{\textbf{Proposed}} method is the largest, which means that the
performance of the \emph{\textbf{Proposed}} method can cope with
network variations better.

Because DASH achieves lossless transmission to improve the \emph{QoE}
of users at the cost of transmitting additional signaling information
of TCP for HTTP sessions between the server and user, the overhead
problem is common and inevitable in DASH. The signaling overhead is
essentially determined by the number of HTTP sessions. Therefore, we
also investigated the signaling overhead of the information exchanges
between servers and users in the proposed algorithm, as shown in
TABLE \ref{table7}. We have to clarify that the reported signaling
overhead in our manuscript is calculated as the \emph{\textbf{time
ratio of the download time of overhead information to that of the
video segments}}. It can be observed that the average proportion of
the signaling overhead induced by the information exchanges is about
30\% of the whole download time. \textbf{It is worth pointing out
that by using additional HTTP sessions, the performance of a DASH
system can be improved}.

\begin{table}[htbp]
\setlength{\abovecaptionskip}{0.cm}
\setlength{\belowcaptionskip}{-0.cm}
  \centering
  \caption{Performance Comparisons of Different Methods}
\includegraphics[width=8cm]{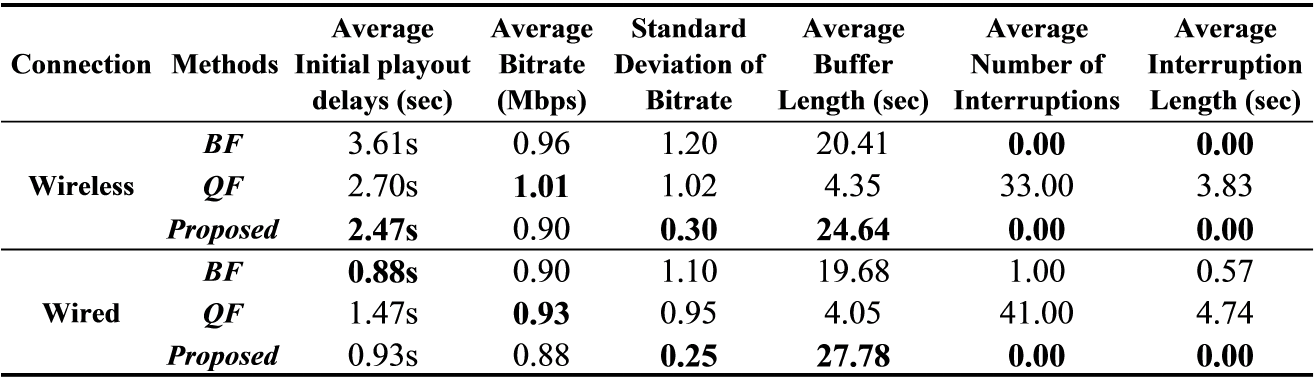}
\label{table6}
\end{table}

\begin{table}[htbp]
\setlength{\abovecaptionskip}{0.cm}
\setlength{\belowcaptionskip}{-0.cm}
  \centering
  \caption{The Proportion of Signaling Overhead Introduced by the Information Exchanges between Server and Users}
\includegraphics[width=8cm]{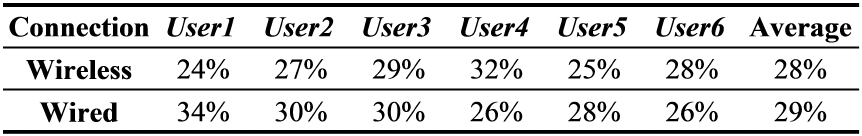}
\label{table7}
\end{table}

\subsection{Results of Case 2}
\begin{figure}
\setlength{\abovecaptionskip}{0.cm}
\setlength{\belowcaptionskip}{-0.cm} \centering \subfigure[]{
\label{fig18:subfig:a} 
\includegraphics[width=4.3cm]{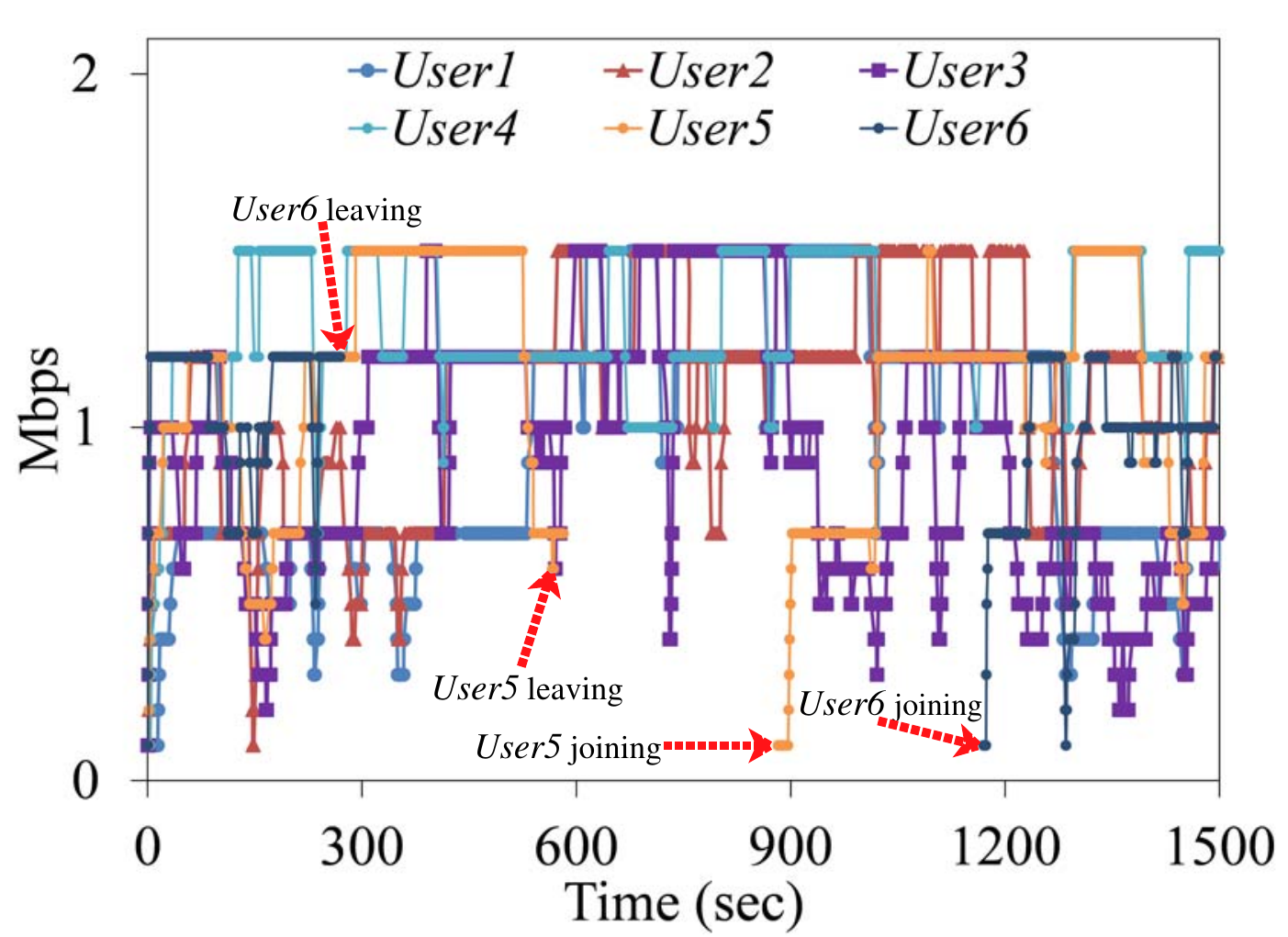}}
\subfigure[]{
\label{fig18:subfig:b} 
\includegraphics[width=4.3cm]{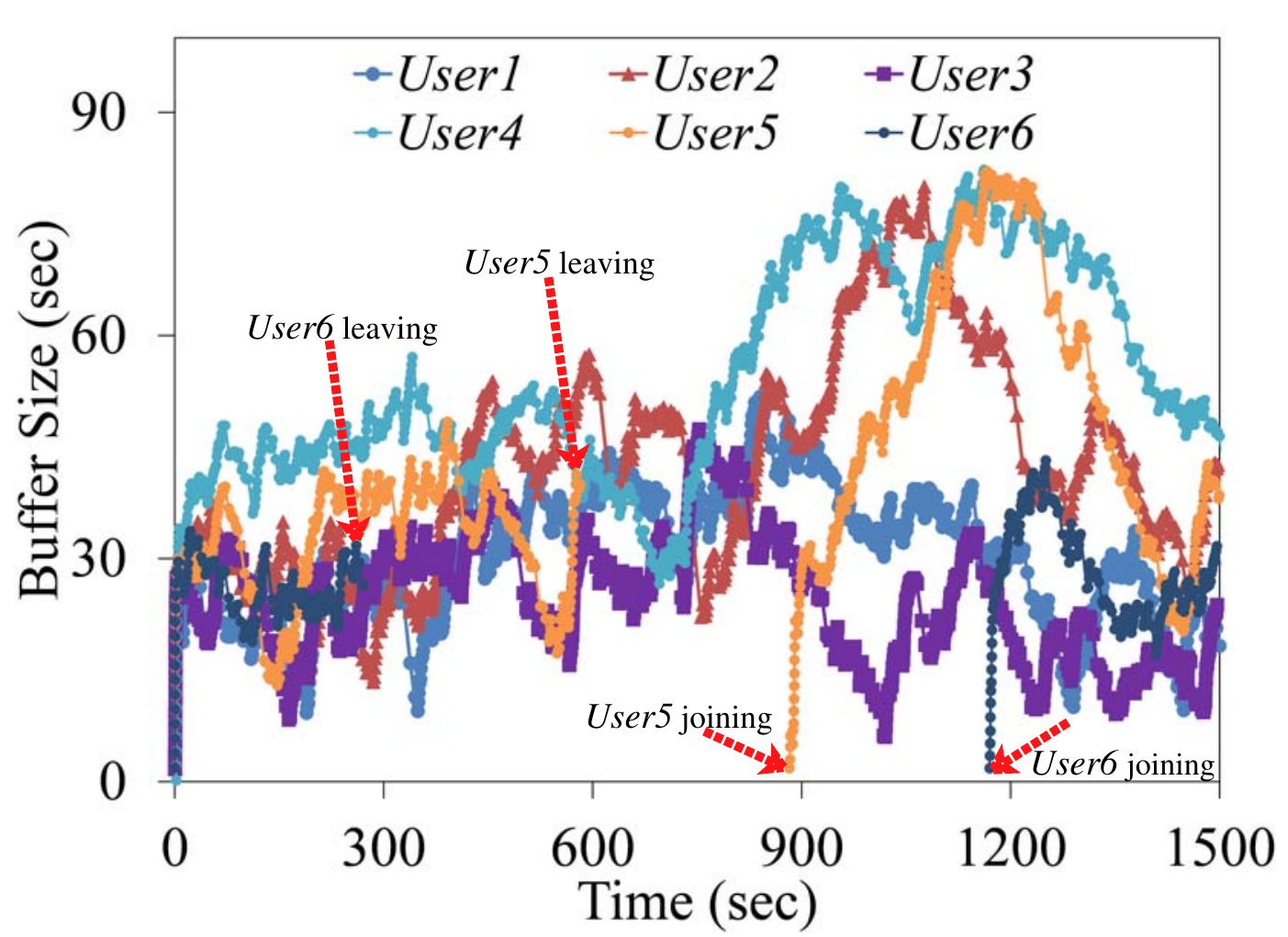}}
\caption{Experimental Results of realistically modeled wireless
network, in which users' requests come and leave at different time
(\emph{User5} leaving at about 600s and joining at 900s, \emph{User6}
leaving at about 300s and joining at 1200s), with $\theta=40$,
$\mu=0.003$, and $\nu=0.0041$, and the server export bandwidth is
6Mbps, (a) dynamic behavior of requested bitrates, (b) the actual
buffer lengths of the 6 users with the reference buffer length of
20s.}
\label{fig18} 
\end{figure}

\begin{figure}
\setlength{\abovecaptionskip}{0.cm}
\setlength{\belowcaptionskip}{-0.cm} \centering \subfigure[]{
\label{fig19:subfig:a} 
\includegraphics[width=4.3cm]{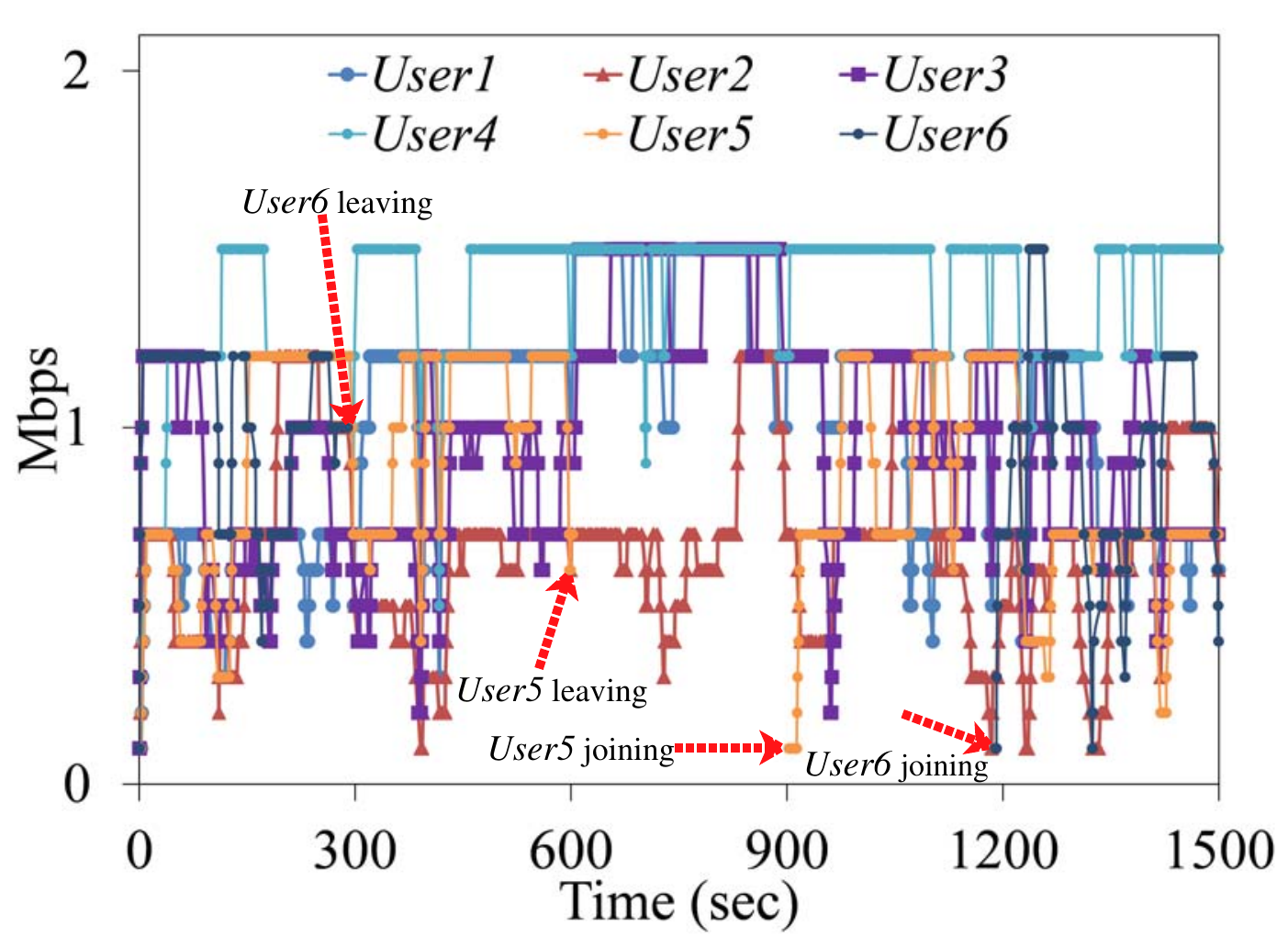}}
\subfigure[]{
\label{fig19:subfig:b} 
\includegraphics[width=4.3cm]{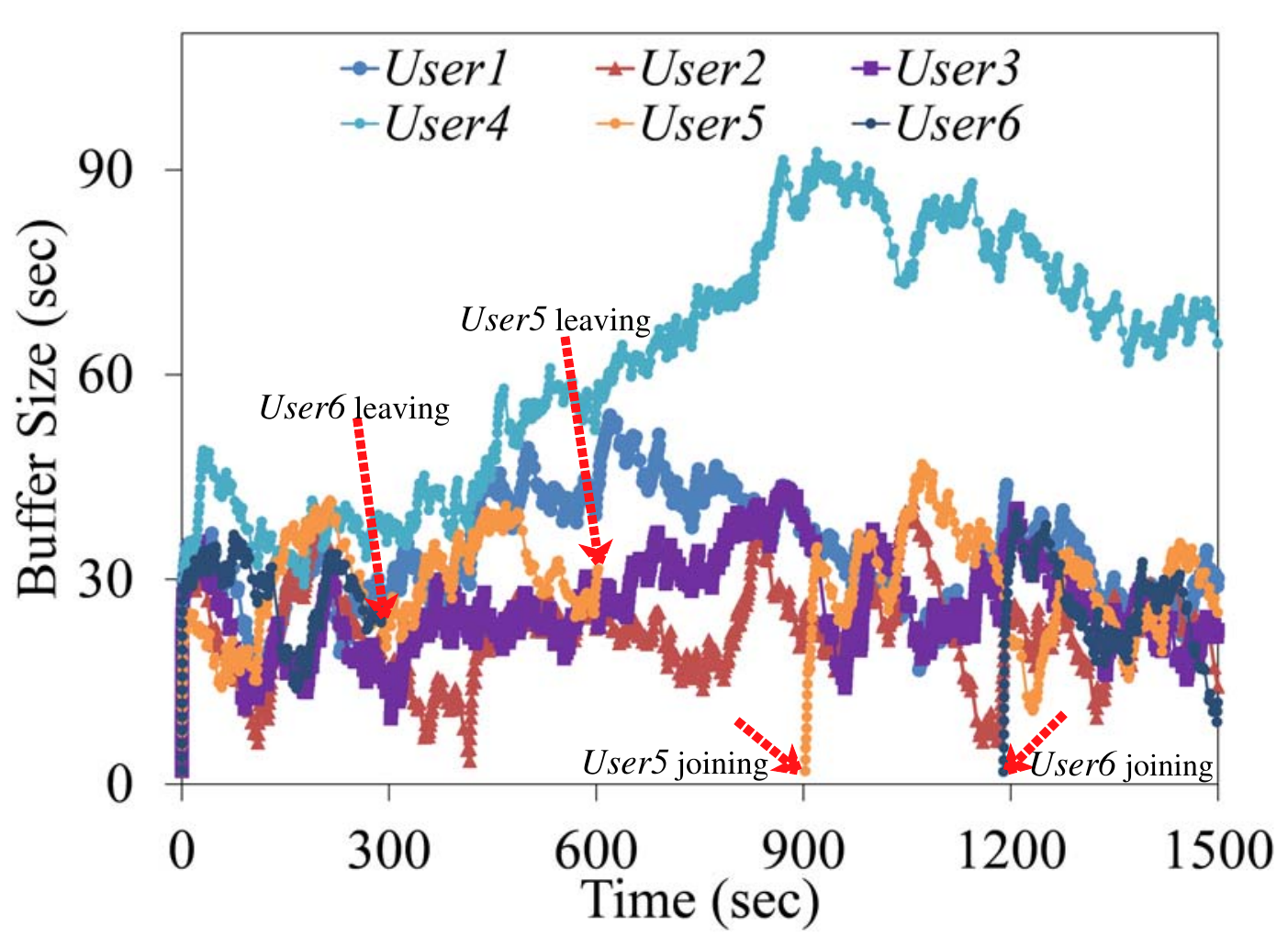}}
\caption{Experimental Results of realistically modeled wired network,
in which users' requests come and leave at different time
(\emph{User5} leaving at about 600s and joining at 900s, \emph{User6}
leaving at about 300s and joining at 1200s), with $\theta=40$,
$\mu=0.003$, and $\nu=0.0041$, and the server export bandwidth is
6Mbps, (a) dynamic behavior of requested bitrates, (b) the actual
buffer lengths of the 6 users with the reference buffer length of
20s.}
\label{fig19} 
\end{figure}

We verified the performance of the proposed algorithm when users'
requests dynamically come and leave at different times (i.e.,
\emph{User5} leaves at about 600s and joins at 900s, \emph{User6}
leaves at about 300s and joins at 1200s). Experimental results are
shown in Figs. \ref{fig18} (wireless) and \ref{fig19} (wired). From
Figs. \ref{fig18:subfig:a} and \ref{fig19:subfig:a}, we can observe
that when \emph{User} 5 and 6 leave, the requested bitrates of the
remaining users increase to 1.5Mbps gradually, while the requested
video bitrates of all the users gradually converge to 1Mbps when
\emph{User} 5 and 6 join again. Accordingly, the effectiveness of the
proposed algorithm is demonstrated for the realistic network scenario
with dynamic user leaving or joining.

\section{Conclusion And Discussion}
In this paper, we have presented a novel non-cooperative game theory
based algorithm to address the rate adaptation issue posed in a DASH
system with single-server and multi-users. The proposed algorithm can
not only guarantee user fairness but also improve user \emph{QoE}.
Moreover, no proxy is required with our algorithm. We have formulated
the rate adaptation problem as a non-cooperative game by building a
novel user \emph{QoE} model that considers the received video
quality, reference buffer length, and accumulated buffer lengths of
users. We have theoretically proven the existence of the Nash
Equilibrium of our specific game, which can be found by our
distributed iterative algorithm with stability analysis. Simulation
and experimental results show that the quality and bitrates of
received videos by the proposed algorithm are more stable than the
state-of-the-art methods, while the actual buffer length of each user
moves around the reference buffer all the time. Besides, there is no
playback interruption for the proposed algorithm.

Although the proposed rate adaptation algorithm can achieve
impressive performance compared with existing algorithm, we believe
it can be further improved by addressing the following limitations:

(1) Restriction on the \emph{QoE} model. In order to guarantee the
existence of the Nash Equilibrium, the \emph{QoE} model must be
designed as a continuous and quasi-concave function with respect to
the bitrates of all the users.

(2) Stability analysis of the scenarios with
multi-users or users joining and leaving dynamically. In the proposed
method, we used a distributed iterative algorithm to obtain the Nash
Equilibrium. However, the stability of the distributed iterative
algorithm is analyzed in detail for the scenario with only 2 users.
When there are more than 2 users, the stability analysis will be very
complex, and we only provide a sketch. Besides, for the scenario with
new users joining/leaving dynamically, the stability analysis of the
proposed method is not analyzed well.

(3) Additional HTTP sessions. For the proposed method, additional
HTTP sessions between users and servers are needed to achieve the
Nash Equilibrium, which will introduce additional signaling overhead.
Such signaling overhead may result in latency. Although this is a
common drawback of DASH, it is also highly desirable to develop new
algorithms to reduce the additional HTTP sessions as much as
possible.

\section*{Acknowledgment}

The authors would like to thank Institute of Information Technology
(ITEC) at Klagenfurt University for the valuable and basis work of
DASH. They would also like to thank the editors and anonymous
reviewers for their valuable comments.

\ifCLASSOPTIONcaptionsoff
  \newpage
\fi



%

\end{document}